\documentclass{article}
\usepackage{iclr2027_conference,times}
\iclrfinalcopy
\usepackage[T1]{fontenc}
\usepackage{amsmath,amssymb,booktabs,array,longtable,graphicx,xcolor,tikz}
\usepackage{hyperref,url}
\usepackage{placeins,colortbl,needspace}
\usepackage[normalem]{ulem}
\definecolor{revisionred}{HTML}{BC3038}
\definecolor{revisionpurple}{HTML}{7839A5}
\newif\ifpaperrevisions
\paperrevisionstrue

\hypersetup{colorlinks=true,citecolor=blue!50!black,linkcolor=blue!50!black,urlcolor=blue!50!black}
\definecolor{tbblue}{HTML}{286080}
\definecolor{sweorange}{HTML}{BA702A}
\definecolor{paperink}{HTML}{172B4D}
\definecolor{layerzero}{HTML}{7655CF}
\definecolor{layerone}{HTML}{2384D1}
\definecolor{layertwo}{HTML}{DB9226}
\definecolor{layerthree}{HTML}{168C82}
\newcommand{\modelhead}[3]{\begingroup\fontsize{6.8}{7.8}\selectfont\shortstack[c]{\raisebox{-0.45mm}{\resizebox{!}{2.2mm}{\includegraphics{figures/models/#1.pdf}}}\hspace{1.4pt}#2\\[1.0pt]#3}\endgroup}

\newcommand{\takeawaybox}[3]{%
  \par\medskip\Needspace{8\baselineskip}\noindent
  \begin{tikzpicture}
    \node[draw=#1, line width=0.8pt, rounded corners=5pt,
      fill=#1!11, inner xsep=10pt, inner ysep=10pt,
      text width=\dimexpr\linewidth-21pt\relax, align=left,
      font=\small] (takeaways) {%
      \setlength{\parindent}{0pt}%
      \vspace{2pt}%
      \noindent\ignorespaces#3\unskip\par
      };
    \node[anchor=west, fill=#1!70!black, text=white, rounded corners=2pt,
      inner xsep=7pt, inner ysep=3pt, font=\small\bfseries]
      at ([xshift=10pt]takeaways.north west) {#2 Takeaways};
  \end{tikzpicture}%
  \par\medskip
}

\newcommand{\promptheading}[2]{\par\smallskip\Needspace{4\baselineskip}\noindent\textcolor{#1}{\textbf{#2}}\par\nobreak}
\newenvironment{promptblock}{%
  \begin{list}{}{\setlength{\leftmargin}{8pt}\setlength{\rightmargin}{8pt}\setlength{\topsep}{3pt}\setlength{\parsep}{4pt}}%
  \item\relax\fontsize{8.5}{10}\selectfont\sffamily\raggedright
}{\end{list}}

\title{Rethinking Verification in Agent Harnesses}
\author{%
Zhengxin Zhang$^{1}$\thanks{This work was done during an internship at Alibaba Group.}\hspace{0.4em},
Erkang Zhu$^{2}$,
Yin Lin$^{2}$, Claire Cardie$^{1}$, Bolin Ding$^{2}$, Jingren Zhou$^{2}$ \\[3pt]
$^{1}$Cornell University \qquad $^{2}$Alibaba Group
}
\hypersetup{pdfauthor={Zhengxin Zhang, Erkang Zhu, Yin Lin, Claire Cardie, Bolin Ding, Jingren Zhou},pdftitle={Rethinking Verification in Agent Harnesses}}
\begin{document}
\maketitle
\lhead{}
\begin{abstract}
Agent harnesses enable large language models to act through tools, observe the results, and iteratively work toward a task goal. Verification checks and revises agents' work to improve output quality. Yet there is limited guidance on how harnesses should organize this effort. We study verification as an axis of test-time scaling by examining when to intervene, who checks the work, and how checking leads to revision. Our framework spans four layers of the agent lifecycle: initial instructions, mid-turn reminders, completion gates, and independent review. We compare 17 designs across seven frontier coding models on Terminal-Bench (TB) and SWE-bench Verified (SWE), and introduce VisualBench for expert evaluation of visual and interactive artifacts.
Initial instructions and execution-time reminders yield modest, model-dependent average gains on the coding benchmarks, without substantially increasing average output token use. Completion gates can yield larger gains but may substantially increase cost. Independent review by a critic offers a more favorable accuracy--cost tradeoff, improving average accuracy by $5.9$--$9.5$ percentage points on TB and $3.9$--$6.6$ on SWE at $1.13$--$1.25\times$ and $1.00$--$1.04\times$ baseline cost, respectively. On VisualBench, independent review achieves net preferences of $13.2$--$23.8$ points over the baseline across its four designs, exceeding the evaluated L0--L2 designs. Additional review rounds or independent critics further improve average TB accuracy by up to $3.5$ percentage points. Together, these results provide empirical guidance for verification design in agent harnesses and support independent review as a practical axis of test-time scaling.
\end{abstract}
\section{Introduction}

Coding agents built on large language models (LLMs) tackle long-horizon software engineering tasks, including resolving repository-level issues across large codebases \citep{yang2024sweagent,anthropic2026claudecode,openai2026codexcli,google2026geminicli}. These agents run inside harnesses that let the model act through tools, observe the results, and iterate toward a task goal \citep{yao2022react,yang2024sweagent}. Beyond exposing tools, the harness specifies the agent's instructions, controls intervention timing, and coordinates verification before accepting its work \citep{bolin2026codexloop,lopopolo2026harness}. These choices allocate computation to checking and revising the agent's work, making verification an important part of harness design \citep{lopopolo2026harness}.

Prior work develops learned verifiers for selecting solutions \citep{cobbe2021verifiers,lightman2023verify}, self-correction through critiques and reflections \citep{madaan2023selfrefine,shinn2023reflexion}, and tool-assisted checks that return execution evidence for repair \citep{gou2023critic,chen2023selfdebug,huang2023agentcoder}. In practice, verification in agent harnesses is often shaped by engineering heuristics and iterative experience: Claude Code provides hooks for completion checks \citep{anthropic2026hooks}, while Codex uses self-review and iteration \citep{lopopolo2026harness}. Yet systematic evidence remains limited on how to design and scale verification, when a design improves performance, and whether its gains justify the additional computation. Without this guidance, current designs may be suboptimal for a given model or task.

In this paper, we study verification as an axis of test-time scaling in agent harnesses. We organize the design space around three questions: when to intervene, who should check the work, and how findings should lead to revision. Following the agent lifecycle, \textbf{L0: Initial instructions} and \textbf{L1: Execution-time reminders} shape what the agent is told before and during execution to guide its own checking; \textbf{L2: Completion gates} reopen its work for self-checking; and \textbf{L3: Independent review} assigns a separate critic to check the work and return findings for repair before it is accepted. This framework lets us compare prompting the agent to verify, extending its self-checking, and introducing an independent checking perspective. Within these layers, we vary message role, intervention timing, critic context, and review strategy, then scale independent review through additional rounds or critics.

We compare 17 verification designs on Terminal-Bench (TB) \citep{merrill2026terminal} and SWE-bench Verified (SWE) \citep{jimenez2023swebench,openai2024verified} tasks using seven frontier coding models: Qwen3.7-Max \citep{qwen2026qwen37}, Qwen3.8-Max \citep{qwen2026qwen38}, GLM-5.2 \citep{zai2026glm52}, DeepSeek V4-Pro \citep{deepseek2026v4}, DeepSeek V4-Flash \citep{deepseek2026v4}, Kimi-K2.7-Code \citep{moonshot2026kimi27code}, and MiniMax-M2.5 \citep{minimax2026m25}. We also introduce an internal VisualBench of 18 visual and interactive tasks, evaluated through blinded expert comparisons of functionality and visual quality.
L0 initial instructions and L1 execution-time reminders yield modest, model-dependent average gains on the coding benchmarks, using $0.90$--$1.00\times$ and $0.75$--$0.93\times$ baseline output tokens, respectively. L2 completion gates can yield larger gains but may substantially increase cost: rejection-based designs cost $1.60$--$1.76\times$ baseline on TB and $1.48$--$1.56\times$ on SWE. L3 independent review by a separate critic offers a more favorable accuracy--cost tradeoff, improving average accuracy by $5.9$--$9.5$ percentage points on TB and $3.9$--$6.6$ on SWE at $1.13$--$1.25\times$ and $1.00$--$1.04\times$ baseline cost, respectively, including critic review and agent revision. On VisualBench, all four independent-review designs outperform the evaluated L0--L2 designs, with net preferences of $+13.2$--$+23.8$ points. Changes-based review outperforms trajectory-based review both with and without required visual inspection. Additional review rounds or independent critics further improve average TB accuracy by up to $3.5$ percentage points. Together, these results provide empirical guidance for verification design in agent harnesses and support independent review as a practical axis of test-time scaling, enabling users to configure verification effort for different tasks to improve agent performance.

\section{Related work}
\noindent\textbf{Verification.}
There is a line of works exploring verification for assessing candidates, guiding self-correction, and checking outputs against external evidence. Learned verifiers score complete solutions or intermediate reasoning steps to guide candidate selection \citep{cobbe2021verifiers,lightman2023verify,wang2024mathshepherd}. Code-focused assessment ranges from trained critics that highlight bugs \citep{mcaleese2024critics} to review protocols that combine agent judgments or introduce structured disagreement \citep{tang2024codeagent,qiu2026adversarial}. Self-correction methods use natural-language critiques and reflections to revise outputs or improve subsequent attempts \citep{madaan2023selfrefine,shinn2023reflexion}. Training-based approaches further teach models to improve across successive attempts through fine-tuning or multi-turn reinforcement learning \citep{qu2024rise,kumar2024score}. Studies of reasoning correction and code repair examine the limits of self-generated feedback and show that reported gains depend on feedback quality and the cost of additional attempts \citep{huang2023selfcorrect,olausson2023selfrepair,kamoi2024survey}. Tool-assisted verification grounds error detection in evidence from code execution and external tools \citep{chen2023selfdebug,gou2023critic}, while dedicated test-generation and execution agents return test feedback to guide code revision \citep{huang2023agentcoder}. Together, these strands provide the assessment and feedback mechanisms from which harness-level verification designs can be built.

\noindent\textbf{Agent harnesses.}
Agent harnesses determine how models interact with tools and environments by organizing instructions, tool interfaces, execution flow, and context. Prior work studies interleaved reasoning and action \citep{yao2022react}, agent--computer interfaces \citep{yang2024sweagent}, execution platforms \citep{wang2024openhands}, and structured repair workflows \citep{xia2024agentless}. These studies examine how the surrounding execution design shapes agent behavior. Harness design also governs how computation and information are used across attempts, including agent sampling and selection \citep{zhu2025agentscaling} and reuse of coding rollouts through structured summaries \citep{kim2026codingscaling}. Agent harnesses have become an essential part of modern coding agents, including Claude Code \citep{anthropic2026claudecode}, Codex CLI \citep{openai2026codexcli}, and Gemini CLI \citep{google2026geminicli}.

These works establish useful assessment mechanisms and agent workflows, but offer limited comparative guidance on how a harness should organize checking and revision. We systematically study verification design in agent harnesses across multiple models and benchmarks, examining when to intervene, who performs the checks, how feedback leads to revision, and how additional verification effort affects performance and cost.

\section{Verification Designs}
\label{sec:methods}
\begin{figure}[!t]
\centering
\vspace{-12pt}
\includegraphics[width=\linewidth,pagebox=cropbox]{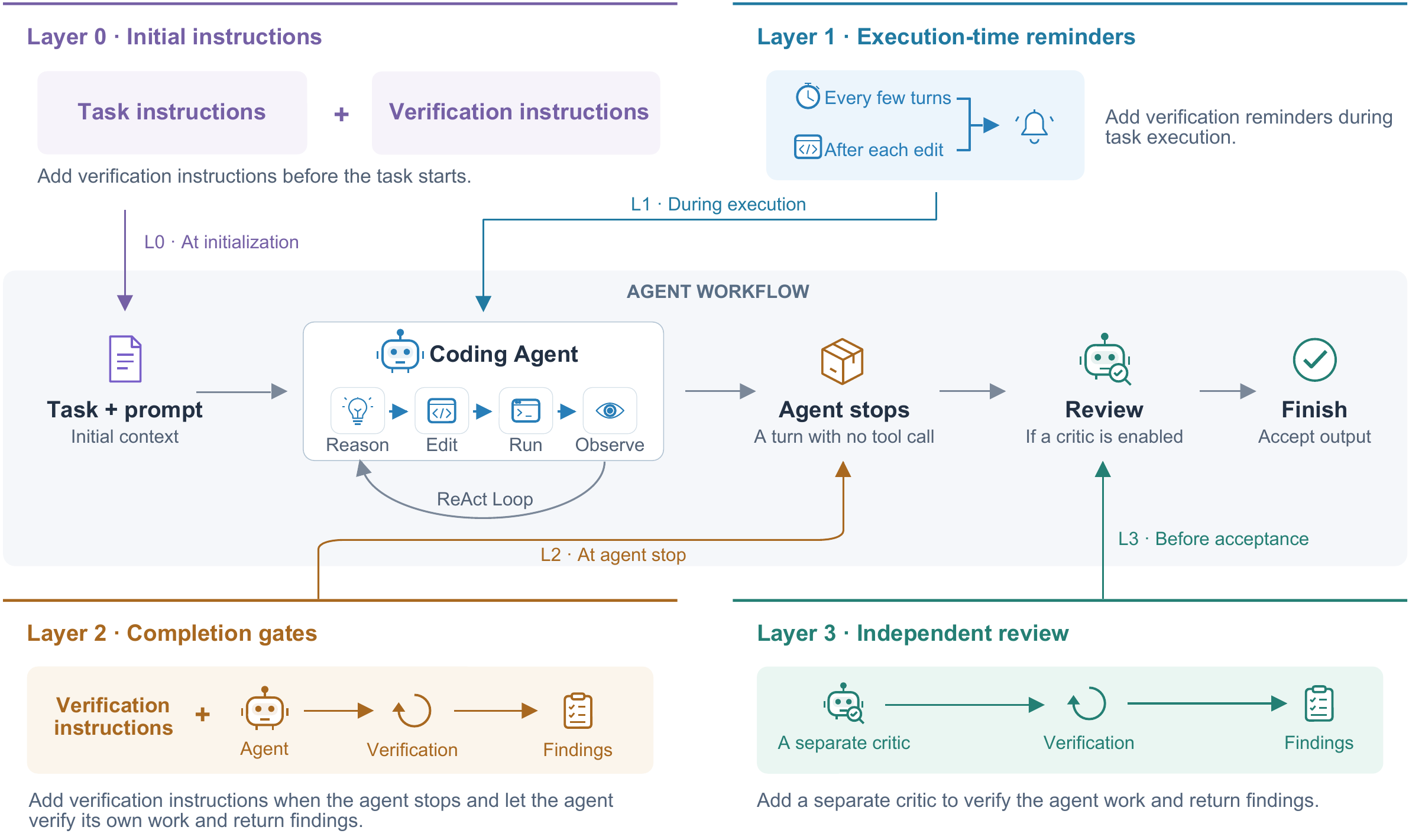}
\vspace{-6pt}
\caption{Framework overview.}
\label{fig:framework}
\vspace{-12pt}
\end{figure}

In this section, we introduce verification designs by following the agent lifecycle, from receiving a task to delivering a result as shown in Figure~\ref{fig:framework}. Table~\ref{tab:layers} summarizes when each layer intervenes, who checks the work, and how execution continues.

\begin{table}[!t]
\centering\small
\caption{Verification layers and their roles in the execution and revision process.}
\label{tab:layers}
\renewcommand{\arraystretch}{1.12}
\begin{tabular}{p{1.1cm}p{2.35cm}p{2.45cm}p{6.2cm}}
\toprule
\textbf{Layer} & \textbf{When} & \textbf{Who checks} & \textbf{How execution continues} \\
\midrule
\rowcolor{layerzero!9}\textcolor{layerzero}{\textbf{L0}} & At initialization & Agent & Use initial verification instructions to guide checking and revision. \\
\rowcolor{layerone!9}\textcolor{layerone}{\textbf{L1}} & During execution & Agent & Check recent changes and revise ongoing work after a reminder. \\
\rowcolor{layertwo!9}\textcolor{layertwo}{\textbf{L2}} & At completion  & Agent & Resume work for self-checking and revision after a check request or rejected completion. \\
\rowcolor{layerthree!9}\textcolor{layerthree}{\textbf{L3}} & Before acceptance & Separate critic & Return critic findings to guide agent revision. \\
\bottomrule
\end{tabular}
\end{table}

Before execution begins, the agent receives the task and the instructions that guide its work. At this stage, \textbf{L0: Initial instructions} introduces verification guidance into the initial prompt. We vary the message role of the verification instructions: \emph{System} or \emph{User}. In the \emph{System} variant, we append it to the system prompt as standing guidance. Because users do not always include verification instructions in their requests, we introduce a \emph{User} variant that appends the same prompt to the initial user request. Both variants ask the agent to run the deliverable through its intended interface, inspect the output, and fix observed defects. The L0 verification instructions are in Appendix~\ref{app:instructions-l0}. For visual or interactive artifacts, the instruction also asks the agent to render the result and exercise its core interactions. The L0 verification instructions for visual tasks in our VisualBench are in Appendix~\ref{app:instructions-l0}.

As execution proceeds, changes to the codebase accumulate across turns. In our experiment, we find that only $6.2$--$13.6\%$ of detected editing turns are followed by a recognized checking command before the next editing turn or the end of the run, as reported in Appendix Table~\ref{tab:spontaneous-checks}. We therefore introduce \textbf{L1: Execution-time reminders} to encourage the agent to check recent changes as it works, creating opportunities to identify and correct errors before further changes accumulate. These reminders arrive after tool results and before the next model call, so the agent can check its recent changes before continuing. We evaluate three designs. \emph{Periodic reminder (System)} sends a reminder a few turns after the edit. \emph{After-edit reminder (System)} sends the same reminder after a turn that changes a file. These two designs use a marker identifying the message as an automated self-verification checkpoint. \emph{After-edit reminder (User)} uses the same edit-based trigger but delivers the reminder as a user message without that marker. Thus, the first comparison varies the trigger, while the second varies the message role and the presence of the automated marker. The L1 verification instructions are in Appendix~\ref{app:instructions-l1}.

When the agent attempts to finish, \textbf{L2: Completion gates} reopen the agent's work for checking and revision. To examine how the requested action guides subsequent checking and revision, we vary whether the gate asks the agent to inspect changes or run checks, or rejects completion and asks it to continue working. We vary the message role (\emph{System} or \emph{User}) and whether user messages carry an automated marker to study how these choices affect the agent's response. We also vary intervention timing to test whether reserving time for checking and repair improves outcomes. The L2 prompts and triggering rules are in Appendix~\ref{app:instructions-l2}.

Before final acceptance, the harness can also assign verification to a separate model, which we call the \emph{critic}. This introduces \textbf{L3: Independent review}: the critic starts in a fresh context and assesses the agent's deliverable. The critic receives the task specification and access to the live sandbox, where it can run its own checks. The critic returns its findings to the agent for revision. We vary the critic's context and whether its checks are guided by task requirements alone (\emph{Task-first}) or also by the agent's assumptions and claims (\emph{Claim-guided}). The \emph{Changes} condition provides the critic with the task and the agent's changes. The \emph{Trajectory} condition provides the task, changes, and the agent's execution trajectory. The critic instructions are in Appendix~\ref{app:instructions-l3}.

\section{Experimental setup}
\label{sec:setup}
\noindent\textbf{Models.}
We evaluate seven coding models through Alibaba Cloud Model Studio (DashScope): Qwen3.7-Max \citep{qwen2026qwen37}, Qwen3.8-Max \citep{qwen2026qwen38}, GLM-5.2 \citep{zai2026glm52}, DeepSeek V4-Pro \citep{deepseek2026v4}, DeepSeek V4-Flash \citep{deepseek2026v4}, Kimi-K2.7-Code \citep{moonshot2026kimi27code}, and MiniMax-M2.5 \citep{minimax2026m25}. Qwen3.7-Plus serves as the independent critic for every coding model.

\noindent\textbf{Benchmarks.}
We evaluate on task subsets from Terminal-Bench 2.1 (TB) \citep{merrill2026terminal} and SWE-bench Verified (SWE) \citep{jimenez2023swebench,openai2024verified}, both with automatic graders. Task selection aims to cover a range of difficulties, using baseline success rates for TB and human time-to-fix tiers followed by baseline evaluation for SWE. Appendix~\ref{app:taskselection} provides task counts and the selection procedure. Each coding condition is configured for three trials per model and task.

VisualBench contains 18 visual and interactive tasks. Each provides a task description and any supporting files, and asks the agent to produce an artifact, such as a playable game, an interactive 3D configurator, or a sales data explorer. For models without native visual capabilities, a \texttt{look\_at} tool renders the artifact and returns a description from Qwen3.7-Plus.
Raters from an internal company data team compare artifacts from a verification condition and the baseline for the same task and model, with conditions hidden and sides randomized. They receive the task brief and short interaction instructions, and judge functionality and visual quality against the task requirements. Outcomes are recorded as wins, losses, or ties; cases where both artifacts fail are recorded separately.
We evaluate one representative design from each of L0--L2 and four L3 designs, with two trials configured per task, model, and condition. Appendix~\ref{app:human-eval} provides the full descriptions of all 18 tasks, expert instructions, and screenshots of the evaluation interface.

\noindent\textbf{Environment.}
All experiments run in E2B cloud sandboxes \citep{e2b2026docs}, orchestrated with Harbor \citep{harbor2026framework}. Each trial starts in a fresh sandbox containing the task files, dependencies, and execution tools. The agent interacts with this environment by reading and editing files, running commands, and inspecting outputs. For independent review, the critic accesses the same workspace to examine the deliverable and execute its own checks. The critic does not have permission to modify the deliverable. The agent can then revise the artifact using the returned feedback.

\noindent\textbf{Metrics.}
For TB and SWE, we report accuracy (the percentage of tasks solved) and cost. We average accuracy and cost across trials for each model--task pair. Cost includes the coding agent, the critic, and subsequent revisions, priced separately by model. Figure~\ref{fig:main} reports each design's mean cost relative to its baseline, with per-model plots in Appendix~\ref{app:model-cost}. 
Appendix~\ref{app:cost} lists the API cost prices, and Appendix~\ref{app:token-use} separates agent and critic output tokens.

\section{Results}
\label{sec:results}
In this section, we analyze L0 initial instructions in Section~\mbox{\ref{sec:results-l0}}, L1 execution-time reminders in Section~\mbox{\ref{sec:results-l1}}, L2 completion gates in Section~\mbox{\ref{sec:results-l2}}, and L3 independent review in Section~\mbox{\ref{sec:results-l3}}.

\begin{table}[!t]
\centering
\caption{Accuracy changes (pp) on Terminal-Bench across verification methods and models.}
\label{tab:main-tb}
\begingroup
\definecolor{tableink}{HTML}{172B4D}
\definecolor{gainfill}{HTML}{168C82}
\definecolor{lossfill}{HTML}{DB6A62}
\definecolor{lossink}{HTML}{A4423B}
\definecolor{layerzero}{HTML}{7655CF}
\definecolor{layerone}{HTML}{2384D1}
\definecolor{layertwo}{HTML}{DB9226}
\definecolor{layerthree}{HTML}{168C82}
\fontsize{8.3}{9.7}\selectfont
\setlength{\fboxsep}{1.3pt}
\setlength{\tabcolsep}{1.5pt}
\renewcommand{\arraystretch}{1.02}
\begin{tabular*}{\linewidth}{@{\extracolsep{\fill}}llrrrrrrrr@{}}
\toprule
\textbf{Method} & \textbf{Variant} & \modelhead{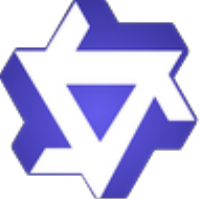}{Qwen}{3.7-Max} & \modelhead{qwen}{Qwen}{3.8-Max} & \modelhead{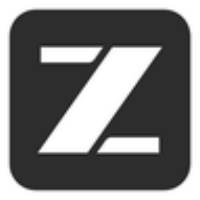}{GLM}{5.2} & \modelhead{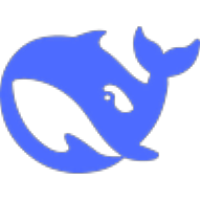}{DeepSeek}{V4-Pro} & \modelhead{deepseek}{DeepSeek}{V4-Flash} & \modelhead{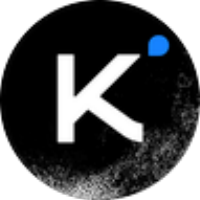}{Kimi}{K2.7 Code} & \modelhead{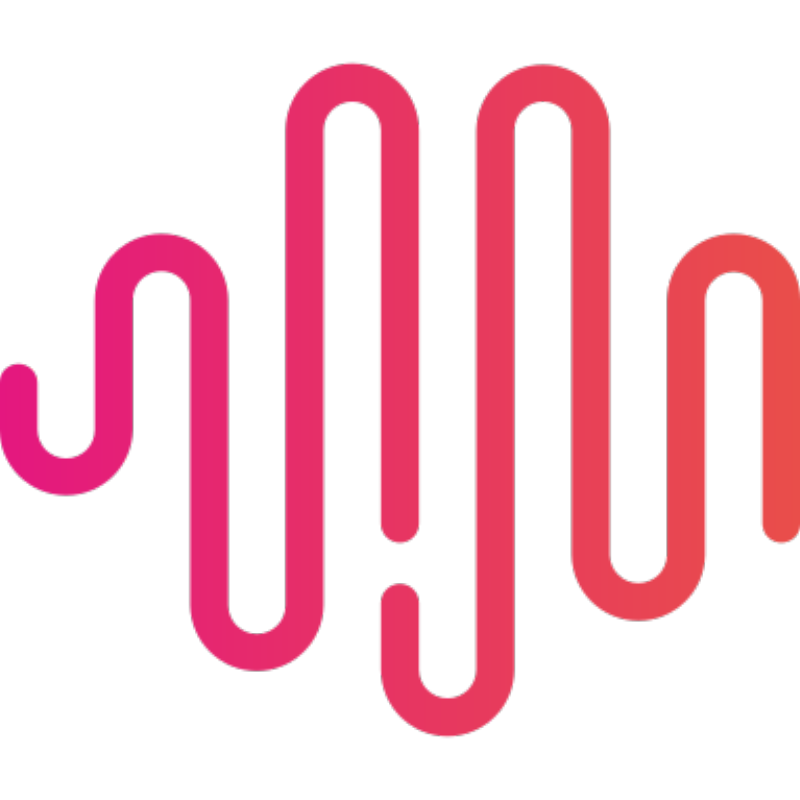}{MiniMax}{M2.5} & \textbf{Average} \\
\midrule
\addlinespace[2pt]
\rowcolor{layerzero!11}\multicolumn{10}{l}{\colorbox{layerzero}{\textcolor{white}{\textbf{L0}}}\hspace{4pt}\textcolor{tableink}{\textbf{Initial instructions}}} \\
\makebox[1.35em][r]{1}\hspace{0.4em}Verification instruction & System & \cellcolor{gainfill!15}$+7.1$ & \cellcolor{gainfill!17}$+8.0$ & \cellcolor{gainfill!14}$+6.0$ & $0.0$ & \cellcolor{gainfill!12}$+4.8$ & \cellcolor{lossfill!8}\textcolor{lossink}{$-2.4$} & $0.0$ & \cellcolor{gainfill!10}$+3.3$ \\
\makebox[1.35em][r]{2}\hspace{0.4em}Verification instruction & User & \cellcolor{gainfill!10}$+3.6$ & \cellcolor{gainfill!11}$+4.0$ & \cellcolor{gainfill!14}$+6.0$ & \cellcolor{gainfill!7}$+1.2$ & \cellcolor{lossfill!7}\textcolor{lossink}{$-1.2$} & \cellcolor{lossfill!17}\textcolor{lossink}{$-8.3$} & \cellcolor{gainfill!10}$+3.6$ & \cellcolor{gainfill!7}$+1.2$ \\
\addlinespace[2pt]
\rowcolor{layerone!11}\multicolumn{10}{l}{\colorbox{layerone}{\textcolor{white}{\textbf{L1}}}\hspace{4pt}\textcolor{tableink}{\textbf{Execution-time reminders}}} \\
\makebox[1.35em][r]{3}\hspace{0.4em}Periodic reminder & System & \cellcolor{gainfill!21}$+10.7$ & \cellcolor{gainfill!22}$+12.0$ & \cellcolor{gainfill!7}$+1.2$ & \cellcolor{gainfill!14}$+6.0$ & \cellcolor{lossfill!7}\textcolor{lossink}{$-1.2$} & \cellcolor{lossfill!7}\textcolor{lossink}{$-1.2$} & $0.0$ & \cellcolor{gainfill!11}$+3.8$ \\
\makebox[1.35em][r]{4}\hspace{0.4em}After-edit reminder & System & \cellcolor{gainfill!15}$+7.1$ & \cellcolor{gainfill!22}$+12.0$ & \cellcolor{gainfill!8}$+2.4$ & $0.0$ & \cellcolor{gainfill!7}$+1.2$ & \cellcolor{lossfill!8}\textcolor{lossink}{$-2.4$} & \cellcolor{gainfill!12}$+4.8$ & \cellcolor{gainfill!10}$+3.5$ \\
\makebox[1.35em][r]{5}\hspace{0.4em}After-edit reminder & User & \cellcolor{gainfill!26}$+14.3$ & \cellcolor{gainfill!24}$\mathbf{+13.3}$ & \cellcolor{gainfill!12}$+4.8$ & \cellcolor{gainfill!10}$+3.6$ & \cellcolor{gainfill!12}$+4.8$ & \cellcolor{gainfill!7}$+1.2$ & \cellcolor{lossfill!10}\textcolor{lossink}{$-3.6$} & \cellcolor{gainfill!13}$+5.4$ \\
\addlinespace[2pt]
\rowcolor{layertwo!11}\multicolumn{10}{l}{\colorbox{layertwo}{\textcolor{white}{\textbf{L2}}}\hspace{4pt}\textcolor{tableink}{\textbf{Completion gates}}} \\
\makebox[1.35em][r]{6}\hspace{0.4em}Inspect changes & Automated & \cellcolor{gainfill!10}$+3.6$ & \cellcolor{gainfill!21}$+10.7$ & \cellcolor{lossfill!10}\textcolor{lossink}{$-3.6$} & \cellcolor{gainfill!12}$+4.8$ & \cellcolor{lossfill!17}\textcolor{lossink}{$-8.3$} & \cellcolor{lossfill!12}\textcolor{lossink}{$-4.8$} & \cellcolor{gainfill!10}$+3.6$ & \cellcolor{gainfill!6}$+0.7$ \\
\makebox[1.35em][r]{7}\hspace{0.4em}Run and check & Automated & \cellcolor{gainfill!21}$+10.7$ & \cellcolor{gainfill!22}$+12.0$ & \cellcolor{gainfill!7}$+1.2$ & \cellcolor{gainfill!17}$+8.3$ & \cellcolor{lossfill!21}\textcolor{lossink}{$-10.7$} & \cellcolor{lossfill!10}\textcolor{lossink}{$-3.6$} & \cellcolor{gainfill!7}$+1.2$ & \cellcolor{gainfill!9}$+2.6$ \\
\makebox[1.35em][r]{8}\hspace{0.4em}Run and check & System & \cellcolor{gainfill!21}$+10.7$ & \cellcolor{gainfill!17}$+8.0$ & $0.0$ & \cellcolor{gainfill!19}$+9.5$ & \cellcolor{gainfill!7}$+1.2$ & \cellcolor{lossfill!8}\textcolor{lossink}{$-2.4$} & \cellcolor{gainfill!12}$+4.8$ & \cellcolor{gainfill!12}$+4.5$ \\
\makebox[1.35em][r]{9}\hspace{0.4em}Reject completion & Automated & \cellcolor{gainfill!12}$+4.8$ & \cellcolor{gainfill!21}$+10.7$ & \cellcolor{lossfill!8}\textcolor{lossink}{$-2.4$} & \cellcolor{gainfill!17}$+8.3$ & \cellcolor{gainfill!8}$+2.4$ & \cellcolor{gainfill!7}$+1.2$ & \cellcolor{gainfill!10}$+3.6$ & \cellcolor{gainfill!11}$+4.0$ \\
\makebox[1.35em][r]{10}\hspace{0.4em}Reject completion & User & \cellcolor{gainfill!19}$+9.5$ & \cellcolor{gainfill!11}$+4.0$ & \cellcolor{lossfill!7}\textcolor{lossink}{$-1.2$} & \cellcolor{gainfill!10}$+3.6$ & \cellcolor{gainfill!12}$+4.8$ & \cellcolor{lossfill!12}\textcolor{lossink}{$-4.8$} & \cellcolor{lossfill!7}\textcolor{lossink}{$-1.2$} & \cellcolor{gainfill!8}$+2.1$ \\
\makebox[1.35em][r]{11}\hspace{0.4em}Recheck before finishing & User & \cellcolor{gainfill!27}$+15.5$ & \cellcolor{gainfill!22}$+12.0$ & \cellcolor{gainfill!8}$+2.4$ & \cellcolor{gainfill!14}$+6.0$ & \cellcolor{gainfill!7}$+1.2$ & \cellcolor{lossfill!12}\textcolor{lossink}{$-4.8$} & \cellcolor{lossfill!8}\textcolor{lossink}{$-2.4$} & \cellcolor{gainfill!11}$+4.1$ \\
\makebox[1.35em][r]{12}\hspace{0.4em}Recheck before finishing & User / early & \cellcolor{gainfill!19}$+9.5$ & \cellcolor{gainfill!24}$\mathbf{+13.3}$ & \cellcolor{gainfill!10}$+3.6$ & \cellcolor{gainfill!19}$+9.5$ & \cellcolor{gainfill!14}$+6.0$ & \cellcolor{lossfill!15}\textcolor{lossink}{$-7.1$} & $0.0$ & \cellcolor{gainfill!12}$+4.8$ \\
\makebox[1.35em][r]{13}\hspace{0.4em}Reject completion & User / early & \cellcolor{gainfill!17}$+8.3$ & \cellcolor{gainfill!17}$+8.0$ & $0.0$ & \cellcolor{gainfill!31}$\mathbf{+17.9}$ & \cellcolor{gainfill!7}$+1.2$ & $0.0$ & \cellcolor{gainfill!15}$+7.1$ & \cellcolor{gainfill!14}$+6.0$ \\
\addlinespace[2pt]
\rowcolor{layerthree!11}\multicolumn{10}{l}{\colorbox{layerthree}{\textcolor{white}{\textbf{L3}}}\hspace{4pt}\textcolor{tableink}{\textbf{Independent review}}} \\
\makebox[1.35em][r]{14}\hspace{0.4em}Claim-guided review & Changes & \cellcolor{gainfill!21}$+10.7$ & \cellcolor{gainfill!15}$+6.7$ & \cellcolor{gainfill!10}$+3.6$ & $0.0$ & $0.0$ & \cellcolor{gainfill!17}$\mathbf{+8.3}$ & \cellcolor{gainfill!22}$\mathbf{+11.9}$ & \cellcolor{gainfill!14}$+5.9$ \\
\makebox[1.35em][r]{15}\hspace{0.4em}Task-first review & Changes & \cellcolor{gainfill!33}$\mathbf{+19.0}$ & \cellcolor{gainfill!21}$+10.7$ & $0.0$ & \cellcolor{gainfill!19}$+9.5$ & \cellcolor{gainfill!10}$+3.6$ & \cellcolor{lossfill!7}\textcolor{lossink}{$-1.2$} & \cellcolor{gainfill!15}$+7.1$ & \cellcolor{gainfill!15}$+6.9$ \\
\makebox[1.35em][r]{16}\hspace{0.4em}Claim-guided review & Trajectory & \cellcolor{gainfill!26}$+14.3$ & \cellcolor{gainfill!24}$\mathbf{+13.3}$ & \cellcolor{gainfill!19}$\mathbf{+9.5}$ & \cellcolor{gainfill!24}$+13.1$ & \cellcolor{gainfill!7}$+1.2$ & \cellcolor{gainfill!8}$+2.4$ & \cellcolor{gainfill!13}$+5.4$ & \cellcolor{gainfill!17}$+8.4$ \\
\makebox[1.35em][r]{17}\hspace{0.4em}Task-first review & Trajectory & \cellcolor{gainfill!27}$+15.5$ & \cellcolor{gainfill!21}$+10.7$ & \cellcolor{gainfill!12}$+4.8$ & \cellcolor{gainfill!26}$+14.3$ & \cellcolor{gainfill!15}$\mathbf{+7.1}$ & \cellcolor{gainfill!15}$+7.1$ & \cellcolor{gainfill!15}$+7.1$ & \cellcolor{gainfill!19}$\mathbf{+9.5}$ \\
\bottomrule
\end{tabular*}
\endgroup

\par\smallskip
\end{table}

\subsection{L0: Initial instructions}
\label{sec:results-l0}
L0 verification instructions yield modest average coding gains, with substantial differences across models and benchmarks. Instructions with the system role improve accuracy by $+3.3$ points on TB and $+3.9$ points on SWE, compared with $+1.2$ and $-1.1$ points for instructions with the user role, as shown in Tables~\ref{tab:main-tb} and~\ref{tab:main-swe}. On SWE, MiniMax-M2.5 benefits from both variants ($+12.6/+8.0$ points for system/user), whereas Qwen3.8-Max declines under both ($-2.3/-2.3$).
Qwen3.8-Max instead gains $+8.0/+4.0$ points on TB with the system/user roles. On VisualBench, L0 yields an average net preference of $-3.2$ points over the baseline, as shown in Figure~\ref{fig:visual-results}. DeepSeek V4-Flash receives a net preference of $+5.6$ points, while GLM-5.2 receives $-8.3$ points despite its positive accuracy gains on both coding benchmarks.

Cost changes are modest: system/user instructions incur $1.08/1.03\times$ baseline cost on TB and $0.94/0.93\times$ on SWE, as shown in Figure~\ref{fig:main}. Average output tokens and agent turns remain close to or below baseline levels across both benchmarks and message roles, at $0.90$--$1.00\times$ and $0.94$--$0.99\times$, respectively, as reported in Appendix Table~\ref{tab:l0-activity}.

These average gains are not accompanied by consistently increased generation or longer interactions. Overall, we recommend concise verification instructions in the system prompt, as system instructions outperform user instructions on average in the coding benchmarks, while their uneven benefits favor leaving room for other harness instructions.

\begin{table}[!t]
\centering
\caption{Accuracy changes (pp) on SWE-bench Verified across verification methods and models.}
\label{tab:main-swe}
\begingroup
\definecolor{tableink}{HTML}{172B4D}
\definecolor{gainfill}{HTML}{168C82}
\definecolor{lossfill}{HTML}{DB6A62}
\definecolor{lossink}{HTML}{A4423B}
\definecolor{layerzero}{HTML}{7655CF}
\definecolor{layerone}{HTML}{2384D1}
\definecolor{layertwo}{HTML}{DB9226}
\definecolor{layerthree}{HTML}{168C82}
\fontsize{8.3}{9.7}\selectfont
\setlength{\fboxsep}{1.3pt}
\setlength{\tabcolsep}{1.5pt}
\renewcommand{\arraystretch}{1.02}
\begin{tabular*}{\linewidth}{@{\extracolsep{\fill}}llrrrrrrrr@{}}
\toprule
\textbf{Method} & \textbf{Variant} & \modelhead{qwen}{Qwen}{3.7-Max} & \modelhead{qwen}{Qwen}{3.8-Max} & \modelhead{glm}{GLM}{5.2} & \modelhead{deepseek}{DeepSeek}{V4-Pro} & \modelhead{deepseek}{DeepSeek}{V4-Flash} & \modelhead{kimi}{Kimi}{K2.7 Code} & \modelhead{minimax}{MiniMax}{M2.5} & \textbf{Average} \\
\midrule
\addlinespace[2pt]
\rowcolor{layerzero!11}\multicolumn{10}{l}{\colorbox{layerzero}{\textcolor{white}{\textbf{L0}}}\hspace{4pt}\textcolor{tableink}{\textbf{Initial instructions}}} \\
\makebox[1.35em][r]{1}\hspace{0.4em}Verification instruction & System & \cellcolor{gainfill!7}$+1.1$ & \cellcolor{lossfill!8}\textcolor{lossink}{$-2.3$} & \cellcolor{gainfill!15}$+6.9$ & \cellcolor{gainfill!8}$+2.3$ & \cellcolor{gainfill!12}$+4.6$ & \cellcolor{gainfill!8}$+2.3$ & \cellcolor{gainfill!23}$+12.6$ & \cellcolor{gainfill!11}$+3.9$ \\
\makebox[1.35em][r]{2}\hspace{0.4em}Verification instruction & User & $0.0$ & \cellcolor{lossfill!8}\textcolor{lossink}{$-2.3$} & \cellcolor{lossfill!8}\textcolor{lossink}{$-2.3$} & \cellcolor{lossfill!10}\textcolor{lossink}{$-3.4$} & \cellcolor{lossfill!13}\textcolor{lossink}{$-5.7$} & \cellcolor{lossfill!8}\textcolor{lossink}{$-2.3$} & \cellcolor{gainfill!17}$+8.0$ & \cellcolor{lossfill!7}\textcolor{lossink}{$-1.1$} \\
\addlinespace[2pt]
\rowcolor{layerone!11}\multicolumn{10}{l}{\colorbox{layerone}{\textcolor{white}{\textbf{L1}}}\hspace{4pt}\textcolor{tableink}{\textbf{Execution-time reminders}}} \\
\makebox[1.35em][r]{3}\hspace{0.4em}Periodic reminder & System & \cellcolor{gainfill!7}$+1.1$ & $0.0$ & \cellcolor{gainfill!10}$+3.4$ & \cellcolor{lossfill!7}\textcolor{lossink}{$-1.1$} & $0.0$ & \cellcolor{gainfill!13}$+5.7$ & \cellcolor{gainfill!23}$+12.6$ & \cellcolor{gainfill!9}$+3.1$ \\
\makebox[1.35em][r]{4}\hspace{0.4em}After-edit reminder & System & \cellcolor{gainfill!12}$\mathbf{+4.6}$ & $0.0$ & \cellcolor{gainfill!13}$+5.7$ & \cellcolor{gainfill!10}$+3.4$ & \cellcolor{lossfill!8}\textcolor{lossink}{$-2.3$} & \cellcolor{lossfill!7}\textcolor{lossink}{$-1.1$} & \cellcolor{gainfill!27}$\mathbf{+14.9}$ & \cellcolor{gainfill!10}$+3.6$ \\
\makebox[1.35em][r]{5}\hspace{0.4em}After-edit reminder & User & \cellcolor{lossfill!12}\textcolor{lossink}{$-4.6$} & $0.0$ & \cellcolor{gainfill!7}$+1.1$ & \cellcolor{lossfill!8}\textcolor{lossink}{$-2.3$} & \cellcolor{lossfill!7}\textcolor{lossink}{$-1.1$} & \cellcolor{lossfill!7}\textcolor{lossink}{$-1.1$} & \cellcolor{gainfill!13}$+5.7$ & \cellcolor{lossfill!5}\textcolor{lossink}{$-0.3$} \\
\addlinespace[2pt]
\rowcolor{layertwo!11}\multicolumn{10}{l}{\colorbox{layertwo}{\textcolor{white}{\textbf{L2}}}\hspace{4pt}\textcolor{tableink}{\textbf{Completion gates}}} \\
\makebox[1.35em][r]{6}\hspace{0.4em}Inspect changes & Automated & \cellcolor{lossfill!8}\textcolor{lossink}{$-2.3$} & \cellcolor{gainfill!7}$+1.1$ & $0.0$ & \cellcolor{lossfill!18}\textcolor{lossink}{$-9.2$} & \cellcolor{gainfill!15}$+6.9$ & \cellcolor{lossfill!8}\textcolor{lossink}{$-2.3$} & \cellcolor{gainfill!18}$+9.2$ & \cellcolor{gainfill!6}$+0.5$ \\
\makebox[1.35em][r]{7}\hspace{0.4em}Run and check & Automated & \cellcolor{lossfill!13}\textcolor{lossink}{$-5.7$} & \cellcolor{gainfill!12}$+4.6$ & $0.0$ & \cellcolor{lossfill!7}\textcolor{lossink}{$-1.1$} & $0.0$ & \cellcolor{lossfill!10}\textcolor{lossink}{$-3.4$} & \cellcolor{gainfill!8}$+2.3$ & \cellcolor{lossfill!6}\textcolor{lossink}{$-0.5$} \\
\makebox[1.35em][r]{8}\hspace{0.4em}Run and check & System & \cellcolor{gainfill!10}$+3.4$ & \cellcolor{lossfill!8}\textcolor{lossink}{$-2.3$} & $0.0$ & \cellcolor{gainfill!7}$+1.1$ & \cellcolor{gainfill!17}$+8.0$ & \cellcolor{lossfill!15}\textcolor{lossink}{$-6.9$} & \cellcolor{gainfill!15}$+6.9$ & \cellcolor{gainfill!7}$+1.5$ \\
\makebox[1.35em][r]{9}\hspace{0.4em}Reject completion & Automated & $0.0$ & \cellcolor{gainfill!8}$+2.3$ & \cellcolor{gainfill!18}$+9.2$ & \cellcolor{gainfill!12}$+4.6$ & \cellcolor{gainfill!20}$\mathbf{+10.3}$ & \cellcolor{lossfill!8}\textcolor{lossink}{$-2.3$} & \cellcolor{gainfill!23}$+12.6$ & \cellcolor{gainfill!13}$+5.3$ \\
\makebox[1.35em][r]{10}\hspace{0.4em}Reject completion & User & $0.0$ & $0.0$ & \cellcolor{gainfill!25}$\mathbf{+13.8}$ & $0.0$ & \cellcolor{gainfill!10}$+3.4$ & $0.0$ & \cellcolor{gainfill!22}$+11.5$ & \cellcolor{gainfill!11}$+4.1$ \\
\makebox[1.35em][r]{11}\hspace{0.4em}Recheck before finishing & User & \cellcolor{gainfill!12}$\mathbf{+4.6}$ & \cellcolor{lossfill!15}\textcolor{lossink}{$-6.9$} & \cellcolor{lossfill!8}\textcolor{lossink}{$-2.3$} & \cellcolor{lossfill!10}\textcolor{lossink}{$-3.4$} & \cellcolor{gainfill!7}$+1.1$ & \cellcolor{lossfill!12}\textcolor{lossink}{$-4.6$} & \cellcolor{gainfill!18}$+9.2$ & \cellcolor{lossfill!5}\textcolor{lossink}{$-0.3$} \\
\makebox[1.35em][r]{12}\hspace{0.4em}Recheck before finishing & User / early & \cellcolor{lossfill!12}\textcolor{lossink}{$-4.6$} & \cellcolor{gainfill!8}$+2.3$ & \cellcolor{gainfill!7}$+1.1$ & \cellcolor{gainfill!13}$+5.7$ & \cellcolor{gainfill!20}$\mathbf{+10.3}$ & \cellcolor{gainfill!13}$+5.7$ & \cellcolor{gainfill!17}$+8.0$ & \cellcolor{gainfill!11}$+4.1$ \\
\makebox[1.35em][r]{13}\hspace{0.4em}Reject completion & User / early & \cellcolor{gainfill!8}$+2.3$ & \cellcolor{gainfill!7}$+1.1$ & \cellcolor{gainfill!22}$+11.5$ & \cellcolor{gainfill!12}$+4.6$ & \cellcolor{gainfill!18}$+9.2$ & \cellcolor{gainfill!10}$+3.4$ & \cellcolor{gainfill!22}$+11.5$ & \cellcolor{gainfill!14}$+6.2$ \\
\addlinespace[2pt]
\rowcolor{layerthree!11}\multicolumn{10}{l}{\colorbox{layerthree}{\textcolor{white}{\textbf{L3}}}\hspace{4pt}\textcolor{tableink}{\textbf{Independent review}}} \\
\makebox[1.35em][r]{14}\hspace{0.4em}Claim-guided review & Changes & \cellcolor{lossfill!7}\textcolor{lossink}{$-1.1$} & \cellcolor{gainfill!13}$\mathbf{+5.7}$ & \cellcolor{gainfill!13}$+5.7$ & \cellcolor{gainfill!15}$\mathbf{+6.9}$ & \cellcolor{lossfill!10}\textcolor{lossink}{$-3.4$} & \cellcolor{gainfill!25}$+13.8$ & \cellcolor{gainfill!17}$+8.0$ & \cellcolor{gainfill!12}$+5.1$ \\
\makebox[1.35em][r]{15}\hspace{0.4em}Task-first review & Changes & $0.0$ & \cellcolor{gainfill!12}$+4.6$ & $0.0$ & \cellcolor{gainfill!15}$\mathbf{+6.9}$ & \cellcolor{gainfill!10}$+3.4$ & \cellcolor{gainfill!30}$+17.2$ & \cellcolor{gainfill!25}$+13.8$ & \cellcolor{gainfill!15}$\mathbf{+6.6}$ \\
\makebox[1.35em][r]{16}\hspace{0.4em}Claim-guided review & Trajectory & \cellcolor{gainfill!12}$\mathbf{+4.6}$ & \cellcolor{gainfill!8}$+2.3$ & \cellcolor{gainfill!13}$+5.7$ & $0.0$ & \cellcolor{gainfill!8}$+2.3$ & \cellcolor{gainfill!34}$\mathbf{+20.7}$ & \cellcolor{gainfill!15}$+6.9$ & \cellcolor{gainfill!14}$+6.1$ \\
\makebox[1.35em][r]{17}\hspace{0.4em}Task-first review & Trajectory & \cellcolor{gainfill!7}$+1.1$ & \cellcolor{gainfill!10}$+3.4$ & \cellcolor{lossfill!10}\textcolor{lossink}{$-3.4$} & \cellcolor{gainfill!12}$+4.6$ & \cellcolor{lossfill!8}\textcolor{lossink}{$-2.3$} & \cellcolor{gainfill!28}$+16.1$ & \cellcolor{gainfill!17}$+8.0$ & \cellcolor{gainfill!11}$+3.9$ \\
\bottomrule
\end{tabular*}
\vspace{-\baselineskip}
\endgroup

\par\smallskip
\end{table}

The case studies illustrate how L0 instructions can alter verification behavior that is already present. On \texttt{bn-fit-modify}, MiniMax's baseline run inspects the generated files and their row counts but misses the required edge orientation. With the verification instruction, the agent revisits this requirement, corrects the CSV convention, and passes, as shown in Appendix Figure~\ref{fig:case-1}. On \texttt{git-multibranch}, Qwen3.7-Max checks deployment endpoints in both conditions. However, the instructed run deletes the repository to prepare for another test, removing the deployment hook, and fails despite subsequent reconstruction; the baseline passes, as shown in Appendix Figure~\ref{fig:case-2}.

\takeawaybox{layerzero}{L0}{
\textbf{Keep verification guidance in system prompt lightweight.}
Use concise instructions, leaving room for other harness guidance.
Use user message for task requirements and context.
}

\subsection{L1: Execution-time reminders}
\label{sec:results-l1}
\begin{figure}[!t]
\centering
\vspace{-12pt}
\includegraphics[width=\linewidth]{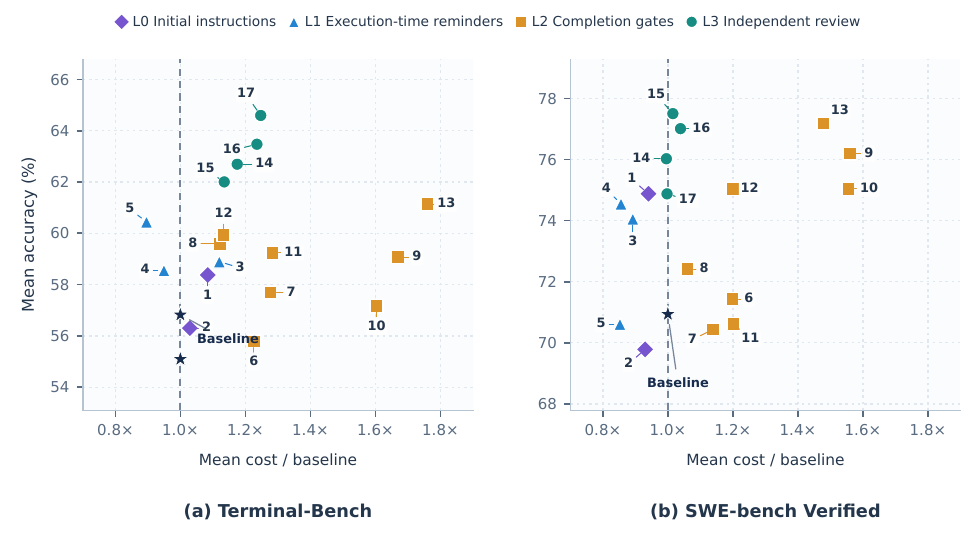}
\vspace{-6pt}
\caption{Mean accuracy versus mean cost relative to baseline (method IDs are in Tables~\ref{tab:main-tb} and~\ref{tab:main-swe}).}
\label{fig:main}
\vspace{-12pt}
\end{figure}

Periodic and after-edit L1 reminders have similar average effects with the system message role: $+3.8$ versus $+3.5$ points on TB and $+3.1$ versus $+3.6$ on SWE. Switching the message role of after-edit reminders to user raises the average gain to $+5.4$ points on TB but lowers it to $-0.3$ on SWE, as shown in Tables~\ref{tab:main-tb} and~\ref{tab:main-swe}.  On VisualBench, MiniMax-M2.5 receives $+12.5$ points, while GLM-5.2 and Qwen3.7-Max each receive $-12.5$ points. Across the three L1 designs, average output tokens decrease by $14$--$25\%$ on TB and $7$--$10\%$ on SWE, while cost ranges from $0.90$--$1.12\times$ and $0.85$--$0.89\times$ baseline, respectively, as shown in Figure~\ref{fig:main} and Appendix Figure~\ref{fig:token-components}. Higher accuracy can also coexist with less generated output. On TB, Qwen3.7-Max gains $+7.1/+14.3$ points with after-edit system/user reminders while using $33\%/28\%$ fewer output tokens, respectively. On VisualBench, L1 yields an average net preference of $-3.2$ points over the baseline, as shown in Figure~\ref{fig:visual-results}. Overall, we recommend verification reminders after code edits to focus the agent on task requirements and the resulting artifact alongside the code itself. The average accuracy gains with fewer output tokens on TB and SWE support using L1, while the VisualBench results motivate checks that also cover rendered output and interactive behavior.

Trajectory analysis distinguishes continued execution from task-relevant verification. On TB, $90$--$97\%$ of after-edit user reminders are followed by execution commands within two agent turns, as reported in Appendix Table~\ref{tab:reminder-diagnostics}. The checks themselves also differ in coverage. On \texttt{nginx-request-logging}, Qwen3.7-Max checks the Nginx configuration, HTTP responses, and access log after a reminder and passes, whereas the baseline stops with an unfinished plan, as shown in Appendix Figure~\ref{fig:case-3}. However, Kimi-K2.7-Code verifies compilation while overlooking invalid word substitutions on \texttt{overfull-hbox}, and DeepSeek V4-Flash verifies output equality yet misses the runtime criterion on \texttt{query-optimize}, as shown in Appendix Figure~\ref{fig:case-4} and Appendix Figure~\ref{fig:case-l1-query}.

\takeawaybox{layerone}{L1}{
\textbf{Use verification reminders after code edits to guide the agent in checking both the code and the resulting artifact against task requirements.}
Include checks of rendered output and interactive behavior when relevant.
}

\subsection{L2: Completion gates}
\label{sec:results-l2}
\begin{figure}[!t]
\centering
\includegraphics[width=\linewidth]{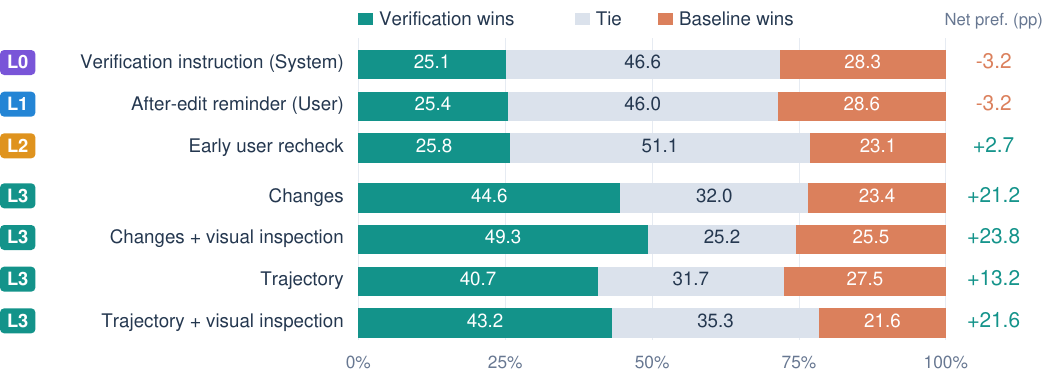}
\caption{VisualBench win, tie, and loss results.}
\label{fig:visual-results}
\vspace{-\baselineskip}
\end{figure}

L2 yields limited gains relative to its cost, even when rejection and earlier intervention improve average accuracy. Automated \emph{Reject completion} yields $+4.0/+5.3$ points on TB/SWE, compared with $+0.7/+0.5$ for \emph{Inspect changes} and $+2.6/-0.5$ for automated \emph{Run and check}. Using the system role for \emph{Run and check} raises the gains to $+4.5/+1.5$ points. Earlier triggering also improves both user rechecking and user rejection: rechecking rises from $+4.1/-0.3$ to $+4.8/+4.1$ points, while rejection rises from $+2.1/+4.1$ to $+6.0/+6.2$ points, the largest average gains among the tested L2 designs, as shown in Tables~\ref{tab:main-tb} and~\ref{tab:main-swe}. Across the rejection designs, cost reaches $1.60$--$1.76\times$ baseline on TB and $1.48$--$1.56\times$ on SWE. Early user rechecking achieves its gains at a lower cost of $1.13/1.20\times$ baseline, respectively, as shown in Figure~\ref{fig:main}. On VisualBench, early rechecking yields an average net preference of $+2.7$ points, as shown in Figure~\ref{fig:visual-results}. GLM-5.2 receives $+9.4$ points, while DeepSeek V4-Flash receives $-8.0$ points. Overall, we recommend avoiding completion-time self-checking, as its gains are limited relative to the additional cost and reopening work does not consistently improve the final artifact.

The L2 cases illustrate both useful repair and invalid final states. On TB, model-level medians range from $12$--$19$ additional steps after user rejection, compared with $4$--$6$ after \emph{Inspect changes}. File-changing actions follow user rejection in $86$--$96\%$ of L2 runs in which user rejection is triggered, as reported in Appendix Table~\ref{tab:gate-diagnostics}. On \texttt{regex-log}, MiniMax-M2.5 resumes after early user rejection and finds that its regex misses dates placed before the IP address. It removes this ordering assumption and passes grading, whereas the baseline retains it and fails, as shown in Appendix Figure~\ref{fig:case-l2-regex}. On \texttt{git-multibranch}, Kimi-K2.7-Code responds to an early recheck with deployment tests that commit and push test content to the served branches. It leaves that content in place and fails, whereas the passing baseline restores the required branch contents, as shown in Appendix Figure~\ref{fig:case-6}. These cases show why further execution alone is insufficient: a completion gate creates an opportunity for repair, but the agent must still identify the relevant defect and preserve a valid final artifact.

\takeawaybox{layertwo}{L2}{
\textbf{Avoid completion-time self-checking.}
Asking the main agent to recheck its own work can improve accuracy, but the gains are limited relative to the additional cost.
}

\subsection{L3: Independent review}
\label{sec:results-l3}

L3 delivers consistent improvement across all four designs on TB and SWE at modest additional cost. The gains range from $+5.9$--$9.5$ points on TB and $+3.9$--$6.6$ on SWE, as shown in Tables~\ref{tab:main-tb} and~\ref{tab:main-swe}. Cost, including critic review and agent revision, is $1.13$--$1.25\times$ baseline on TB and $1.00$--$1.04\times$ on SWE, as shown in Figure~\ref{fig:main}. Compared with the strongest L2 design, the strongest L3 design achieves a larger gain on TB ($+9.5$ versus $+6.0$ points) and a comparable gain on SWE ($+6.6$ versus $+6.2$), at lower cost on both benchmarks. On VisualBench, all four L3 designs receive positive net preference, ranging from $+13.2$ to $+23.8$ points and exceeding the evaluated L0--L2 designs, as shown in Appendix Table~\ref{tab:visual-models}. Changes-based review outperforms trajectory-based review both without required visual inspection ($+21.2$ versus $+13.2$ points) and with it ($+23.8$ versus $+21.6$ points). Required visual inspection improves net preference for both contexts, with a larger increase for trajectory-based review. Both changes-based designs receive positive net preference for all seven models, while trajectory-based designs retain some negative model-level results. Across the three benchmarks, the full trajectory does not consistently improve outcomes, and trajectory-based designs incur $29$--$43\%$ higher critic cost on TB and SWE. Overall, we recommend task requirements and changes as the default context for a separate, low-cost critic.

The trajectories illustrate how independent review turns additional checks into concrete repairs. On \texttt{regex-log}, MiniMax-M2.5 tests its regex and declares completion, but the critic tests a case with the date appearing before the IP address and exposes a missed match. The agent then revises the regex and passes, whereas the baseline performs its own tests but still fails, as shown in Appendix Figure~\ref{fig:case-7}. On \texttt{cobol-modernization}, Qwen3.7-Max's implementation passes its own checks for valid and invalid transactions. The critic tests an arithmetic underflow and finds that Python writes a negative sign that COBOL's unsigned field omits, breaking the required record format. The agent reproduces the discrepancy, repairs the numeric formatting, and passes the external grader; the baseline declares success after its own comparisons but fails grading, as shown in Appendix Figure~\ref{fig:case-9}. 
In both examples, the critic tests a case that the agent overlooked and supplies an executable failure that guides revision. This illustrates how an independent checking perspective can address gaps in the agent's self-verification. On \texttt{cancel-async-tasks}, GLM-5.2's critic confirms that asynchronous cleanup remains incomplete after cancellation, but the run reaches its 45-minute limit without a completed repair after review, as shown in Appendix Figure~\ref{fig:case-10}.

\begin{figure}[!t]
\centering
\includegraphics[width=\linewidth]{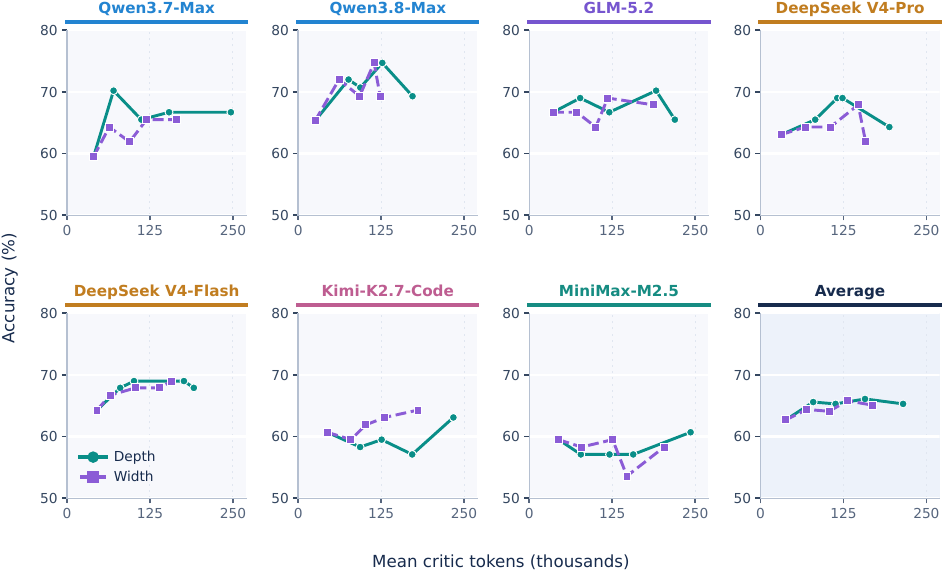}
\caption{Scaling independent review on TB through additional review rounds (depth) and critics (width), showing accuracy versus mean critic token use.}
\label{fig:scale}
\vspace{-\baselineskip}
\end{figure}

\noindent\textbf{Scaling review.}
Increasing review depth (successive review rounds) or width (independent critics) further improves average accuracy on TB. One critic and one round achieve $62.7\%$ accuracy at 38.2k mean critic tokens, compared with $65.6\%$ at 79.5k for two rounds and $64.4\%$ at 69.7k for two critics. All larger settings outperform a single review, reaching $66.1\%$ with four rounds and $65.8\%$ with four critics, as shown in Figure~\ref{fig:scale}. Four critics improve average accuracy by $3.1$ points over one critic at $1.33\times$ total cost, including revisions. Scaling verification through low-cost critics can limit repeated execution by the main agent. By consolidating independent feedback before revision, four critics approach four-round accuracy at $19\%$ lower total cost, as shown in Appendix Figure~\ref{fig:scale-cost}.

\takeawaybox{layerthree}{L3}{
\textbf{Use a separate, low-cost critic to check the deliverable and guide agent revision.} Scaling review through additional rounds or independent critics further improves average accuracy.
}

\FloatBarrier

\section{Conclusion}
In this paper, we systematically study verification in agent harnesses, comparing 17 designs across four layers of the agent lifecycle and seven frontier coding models on TB and SWE, with evaluation on VisualBench. We examine when to intervene, who checks the work, and how feedback leads to revision. Initial instructions and reminders yield modest, model-dependent coding gains without substantially increasing average output token use. Completion gates can improve accuracy but may substantially increase cost. Independent review offers a more favorable accuracy--cost tradeoff on TB and SWE. On VisualBench, its four designs achieve net preferences of $13.2$--$23.8$ points over the baseline, exceeding the evaluated L0--L2 designs. Additional rounds or critics further improve average TB accuracy by up to $3.5$ percentage points. Together, these results provide empirical guidance for verification design in agent harnesses and support independent review as a practical axis of test-time scaling.

\clearpage
\section*{Ethics statement}
VisualBench's human evaluation received ethics approval. Raters are members of an internal company data team; they provided informed consent and were compensated under their existing contracts with the company. The interface explains the study's purpose and assessment criteria, while hiding verification conditions and randomizing artifact placement. Appendix~\ref{app:human-eval} documents the evaluation protocol, task descriptions, rater instructions, and interface.

\section*{AI use statement}
The authors developed the research ideas and system design and led the manuscript writing. Generative AI tools assisted with implementation and manuscript polishing. The authors are responsible for the final content.
\bibliography{references}

\begin{thebibliography}{41}
\providecommand{\natexlab}[1]{#1}
\providecommand{\url}[1]{\texttt{#1}}
\expandafter\ifx\csname urlstyle\endcsname\relax
  \providecommand{\doi}[1]{doi: #1}\else
  \providecommand{\doi}{doi: \begingroup \urlstyle{rm}\Url}\fi

\bibitem[{Alibaba Cloud}(2026{\natexlab{a}})]{alibaba2026cache}
{Alibaba Cloud}.
\newblock {Context Cache}.
\newblock Alibaba Cloud Model Studio documentation, 2026{\natexlab{a}}.
\newblock URL \url{https://help.aliyun.com/zh/model-studio/context-cache}.
\newblock Accessed September 19, 2026.

\bibitem[{Alibaba Cloud}(2026{\natexlab{b}})]{alibaba2026pricing}
{Alibaba Cloud}.
\newblock {Model Studio Model Pricing}.
\newblock Alibaba Cloud Model Studio documentation, 2026{\natexlab{b}}.
\newblock URL \url{https://help.aliyun.com/zh/model-studio/model-pricing}.
\newblock Accessed September 19, 2026.

\bibitem[{Anthropic}(2026{\natexlab{a}})]{anthropic2026claudecode}
{Anthropic}.
\newblock {How Claude Code works}.
\newblock Claude Code documentation, 2026{\natexlab{a}}.
\newblock URL \url{https://code.claude.com/docs/en/how-claude-code-works}.
\newblock Documentation version accessed September 18, 2026.

\bibitem[{Anthropic}(2026{\natexlab{b}})]{anthropic2026hooks}
{Anthropic}.
\newblock {Hooks reference}.
\newblock Claude Code documentation, 2026{\natexlab{b}}.
\newblock URL \url{https://code.claude.com/docs/en/hooks}.
\newblock Documentation version accessed September 21, 2026.

\bibitem[Bolin(2026)]{bolin2026codexloop}
Michael Bolin.
\newblock {Unrolling the Codex agent loop}.
\newblock OpenAI Engineering, January 2026.
\newblock URL \url{https://openai.com/index/unrolling-the-codex-agent-loop/}.
\newblock Published January 23, 2026.

\bibitem[Chen et~al.(2023)Chen, Lin, Schärli, and Zhou]{chen2023selfdebug}
Xinyun Chen, Maxwell Lin, Nathanael Schärli, and Denny Zhou.
\newblock {Teaching Large Language Models to Self-Debug}, 2023.
\newblock URL \url{https://arxiv.org/abs/2304.05128}.

\bibitem[Cobbe et~al.(2021)Cobbe, Kosaraju, Bavarian, Chen, Jun, Kaiser,
  Plappert, Tworek, Hilton, Nakano, Hesse, and Schulman]{cobbe2021verifiers}
Karl Cobbe, Vineet Kosaraju, Mohammad Bavarian, Mark Chen, Heewoo Jun, Lukasz
  Kaiser, Matthias Plappert, Jerry Tworek, Jacob Hilton, Reiichiro Nakano,
  Christopher Hesse, and John Schulman.
\newblock {Training Verifiers to Solve Math Word Problems}, 2021.
\newblock URL \url{https://arxiv.org/abs/2110.14168}.

\bibitem[{DeepSeek-AI} et~al.(2026)]{deepseek2026v4}
{DeepSeek-AI} et~al.
\newblock {DeepSeek-V4: Towards Highly Efficient Million-Token Context
  Intelligence}, 2026.
\newblock URL \url{https://arxiv.org/abs/2606.19348}.

\bibitem[{E2B}(2026)]{e2b2026docs}
{E2B}.
\newblock {E2B} documentation, 2026.
\newblock URL \url{https://docs.e2b.dev/}.
\newblock Documentation version accessed September 19, 2026.

\bibitem[{Google}(2026)]{google2026geminicli}
{Google}.
\newblock {Gemini CLI core}.
\newblock Gemini CLI documentation, 2026.
\newblock URL \url{https://geminicli.com/docs/core/}.
\newblock Documentation version accessed September 18, 2026.

\bibitem[Gou et~al.(2023)Gou, Shao, Gong, Shen, Yang, Duan, and
  Chen]{gou2023critic}
Zhibin Gou, Zhihong Shao, Yeyun Gong, Yelong Shen, Yujiu Yang, Nan Duan, and
  Weizhu Chen.
\newblock {CRITIC: Large Language Models Can Self-Correct with Tool-Interactive
  Critiquing}, 2023.
\newblock URL \url{https://arxiv.org/abs/2305.11738}.

\bibitem[{Harbor Framework Team}(2026)]{harbor2026framework}
{Harbor Framework Team}.
\newblock {Harbor}: A framework for evaluating and optimizing agents and models
  in container environments, 2026.
\newblock URL \url{https://zenodo.org/doi/10.5281/zenodo.20953922}.

\bibitem[Huang et~al.(2023{\natexlab{a}})Huang, Zhang, Luck, Bu, Qing, and
  Cui]{huang2023agentcoder}
Dong Huang, Jie~M. Zhang, Michael Luck, Qingwen Bu, Yuhao Qing, and Heming Cui.
\newblock {AgentCoder: Multi-Agent-based Code Generation with Iterative Testing
  and Optimisation}, 2023{\natexlab{a}}.
\newblock URL \url{https://arxiv.org/abs/2312.13010}.

\bibitem[Huang et~al.(2023{\natexlab{b}})Huang, Chen, Mishra, Zheng, Yu, Song,
  and Zhou]{huang2023selfcorrect}
Jie Huang, Xinyun Chen, Swaroop Mishra, Huaixiu~Steven Zheng, Adams~Wei Yu,
  Xinying Song, and Denny Zhou.
\newblock {Large Language Models Cannot Self-Correct Reasoning Yet},
  2023{\natexlab{b}}.
\newblock URL \url{https://arxiv.org/abs/2310.01798}.

\bibitem[Jimenez et~al.(2023)Jimenez, Yang, Wettig, Yao, Pei, Press, and
  Narasimhan]{jimenez2023swebench}
Carlos~E. Jimenez, John Yang, Alexander Wettig, Shunyu Yao, Kexin Pei, Ofir
  Press, and Karthik Narasimhan.
\newblock {SWE-bench: Can Language Models Resolve Real-World GitHub Issues?},
  2023.
\newblock URL \url{https://arxiv.org/abs/2310.06770}.

\bibitem[Kamoi et~al.(2024)Kamoi, Zhang, Zhang, Han, and
  Zhang]{kamoi2024survey}
Ryo Kamoi, Yusen Zhang, Nan Zhang, Jiawei Han, and Rui Zhang.
\newblock {When Can LLMs Actually Correct Their Own Mistakes? A Critical Survey
  of Self-Correction of LLMs}, 2024.
\newblock URL \url{https://arxiv.org/abs/2406.01297}.

\bibitem[Kim et~al.(2026)Kim, Yang, Niu, Zhang, Zhu, Helenowski, Silva, Chen,
  Iyer, Zaheer, Fried, Hajishirzi, Arora, Synnaeve, Salakhutdinov, and
  Goyal]{kim2026codingscaling}
Joongwon Kim, Wannan Yang, Kelvin Niu, Hongming Zhang, Yun Zhu, Eryk
  Helenowski, Ruan Silva, Zhengxing Chen, Srinivasan Iyer, Manzil Zaheer,
  Daniel Fried, Hannaneh Hajishirzi, Sanjeev Arora, Gabriel Synnaeve, Ruslan
  Salakhutdinov, and Anirudh Goyal.
\newblock {Scaling Test-Time Compute for Agentic Coding}, 2026.
\newblock URL \url{https://arxiv.org/abs/2604.16529}.

\bibitem[Kumar et~al.(2024)Kumar, Zhuang, Agarwal, Su, Co-Reyes, Singh, Baumli,
  Iqbal, Bishop, Roelofs, Zhang, McKinney, Shrivastava, Paduraru, Tucker,
  Precup, Behbahani, and Faust]{kumar2024score}
Aviral Kumar, Vincent Zhuang, Rishabh Agarwal, Yi~Su, John~D Co-Reyes, Avi
  Singh, Kate Baumli, Shariq Iqbal, Colton Bishop, Rebecca Roelofs, Lei~M
  Zhang, Kay McKinney, Disha Shrivastava, Cosmin Paduraru, George Tucker, Doina
  Precup, Feryal Behbahani, and Aleksandra Faust.
\newblock {Training Language Models to Self-Correct via Reinforcement
  Learning}, 2024.
\newblock URL \url{https://arxiv.org/abs/2409.12917}.

\bibitem[Lightman et~al.(2023)Lightman, Kosaraju, Burda, Edwards, Baker, Lee,
  Leike, Schulman, Sutskever, and Cobbe]{lightman2023verify}
Hunter Lightman, Vineet Kosaraju, Yura Burda, Harri Edwards, Bowen Baker, Teddy
  Lee, Jan Leike, John Schulman, Ilya Sutskever, and Karl Cobbe.
\newblock {Let's Verify Step by Step}, 2023.
\newblock URL \url{https://arxiv.org/abs/2305.20050}.

\bibitem[Lopopolo(2026)]{lopopolo2026harness}
Ryan Lopopolo.
\newblock {Harness engineering: leveraging Codex in an agent-first world}.
\newblock OpenAI Engineering, February 2026.
\newblock URL \url{https://openai.com/index/harness-engineering/}.
\newblock Published February 11, 2026.

\bibitem[Madaan et~al.(2023)Madaan, Tandon, Gupta, Hallinan, Gao, Wiegreffe,
  Alon, Dziri, Prabhumoye, Yang, Gupta, Majumder, Hermann, Welleck,
  Yazdanbakhsh, and Clark]{madaan2023selfrefine}
Aman Madaan, Niket Tandon, Prakhar Gupta, Skyler Hallinan, Luyu Gao, Sarah
  Wiegreffe, Uri Alon, Nouha Dziri, Shrimai Prabhumoye, Yiming Yang, Shashank
  Gupta, Bodhisattwa~Prasad Majumder, Katherine Hermann, Sean Welleck, Amir
  Yazdanbakhsh, and Peter Clark.
\newblock {Self-Refine: Iterative Refinement with Self-Feedback}, 2023.
\newblock URL \url{https://arxiv.org/abs/2303.17651}.

\bibitem[McAleese et~al.(2024)McAleese, Pokorny, Uribe, Nitishinskaya, Trebacz,
  and Leike]{mcaleese2024critics}
Nat McAleese, Rai~Michael Pokorny, Juan Felipe~Ceron Uribe, Evgenia
  Nitishinskaya, Maja Trebacz, and Jan Leike.
\newblock {LLM Critics Help Catch LLM Bugs}, 2024.
\newblock URL \url{https://arxiv.org/abs/2407.00215}.

\bibitem[Merrill et~al.(2026)Merrill, Shaw, Carlini, Li, Raj, Bercovich,
  et~al.]{merrill2026terminal}
Mike~A. Merrill, Alexander~G. Shaw, Nicholas Carlini, Boxuan Li, Harsh Raj,
  Ivan Bercovich, et~al.
\newblock {Terminal-Bench: Benchmarking Agents on Hard, Realistic Tasks in
  Command Line Interfaces}, 2026.
\newblock URL \url{https://arxiv.org/abs/2601.11868}.

\bibitem[{MiniMax}(2026)]{minimax2026m25}
{MiniMax}.
\newblock {MiniMax M2.5: Built for Real-World Productivity}.
\newblock Official model announcement, 2026.
\newblock URL \url{https://www.minimax.io/news/minimax-m25}.
\newblock Published February 12, 2026.

\bibitem[{Moonshot AI}(2026)]{moonshot2026kimi27code}
{Moonshot AI}.
\newblock {Kimi K2.7 Code}.
\newblock Kimi Code release notes, 2026.
\newblock URL
  \url{https://www.kimi.com/code/docs/en/kimi-code/whats-new.html#kimi-k2-7-code-june-12-2026}.
\newblock Released June 12, 2026.

\bibitem[Olausson et~al.(2023)Olausson, Inala, Wang, Gao, and
  Solar-Lezama]{olausson2023selfrepair}
Theo~X. Olausson, Jeevana~Priya Inala, Chenglong Wang, Jianfeng Gao, and
  Armando Solar-Lezama.
\newblock {Is Self-Repair a Silver Bullet for Code Generation?}, 2023.
\newblock URL \url{https://arxiv.org/abs/2306.09896}.

\bibitem[{OpenAI}(2024)]{openai2024verified}
{OpenAI}.
\newblock {Introducing SWE-bench Verified}, 2024.
\newblock URL \url{https://openai.com/index/introducing-swe-bench-verified/}.

\bibitem[{OpenAI}(2026)]{openai2026codexcli}
{OpenAI}.
\newblock {Codex CLI}.
\newblock Official Codex documentation, 2026.
\newblock URL \url{https://learn.chatgpt.com/docs/codex/cli}.
\newblock Documentation version accessed September 18, 2026.

\bibitem[Qiu \& Gill(2026)Qiu and Gill]{qiu2026adversarial}
Eric~S. Qiu and Joyce Gill.
\newblock {Adversarial Review: Structured Disagreement for Grounded Agentic
  Code Review}, 2026.
\newblock URL \url{https://arxiv.org/abs/2608.18167}.

\bibitem[Qu et~al.(2024)Qu, Zhang, Garg, and Kumar]{qu2024rise}
Yuxiao Qu, Tianjun Zhang, Naman Garg, and Aviral Kumar.
\newblock Recursive introspection: Teaching language model agents how to
  self-improve, 2024.
\newblock URL \url{https://arxiv.org/abs/2407.18219}.

\bibitem[{Qwen Team}(2026{\natexlab{a}})]{qwen2026qwen37}
{Qwen Team}.
\newblock {Qwen3.7}: The agent frontier, May 2026{\natexlab{a}}.
\newblock URL \url{https://qwen.ai/blog?id=qwen3.7}.

\bibitem[{Qwen Team}(2026{\natexlab{b}})]{qwen2026qwen38}
{Qwen Team}.
\newblock {Qwen3.8-Max}: A new bar for coding and cowork, August
  2026{\natexlab{b}}.
\newblock URL \url{https://qwen.ai/blog?id=qwen3.8}.

\bibitem[Shinn et~al.(2023)Shinn, Cassano, Berman, Gopinath, Narasimhan, and
  Yao]{shinn2023reflexion}
Noah Shinn, Federico Cassano, Edward Berman, Ashwin Gopinath, Karthik
  Narasimhan, and Shunyu Yao.
\newblock {Reflexion: Language Agents with Verbal Reinforcement Learning},
  2023.
\newblock URL \url{https://arxiv.org/abs/2303.11366}.

\bibitem[Tang et~al.(2024)Tang, Kim, Song, Lothritz, Li, Ezzini, Tian, Klein,
  and Bissyande]{tang2024codeagent}
Xunzhu Tang, Kisub Kim, Yewei Song, Cedric Lothritz, Bei Li, Saad Ezzini, Haoye
  Tian, Jacques Klein, and Tegawende~F. Bissyande.
\newblock {CodeAgent: Autonomous Communicative Agents for Code Review}, 2024.
\newblock URL \url{https://arxiv.org/abs/2402.02172}.

\bibitem[Wang et~al.(2024{\natexlab{a}})Wang, Li, Shao, Xu, Dai, Li, Chen, Wu,
  and Sui]{wang2024mathshepherd}
Peiyi Wang, Lei Li, Zhihong Shao, Runxin Xu, Damai Dai, Yifei Li, Deli Chen,
  Yu~Wu, and Zhifang Sui.
\newblock {Math-Shepherd: Verify and Reinforce LLMs Step-by-step without Human
  Annotations}.
\newblock In Lun-Wei Ku, Andre Martins, and Vivek Srikumar (eds.),
  \emph{Proceedings of the 62nd Annual Meeting of the Association for
  Computational Linguistics (Volume 1: Long Papers)}, pp.\  9426--9439,
  Bangkok, Thailand, August 2024{\natexlab{a}}. Association for Computational
  Linguistics.
\newblock \doi{10.18653/v1/2024.acl-long.510}.
\newblock URL \url{https://aclanthology.org/2024.acl-long.510/}.

\bibitem[Wang et~al.(2024{\natexlab{b}})Wang, Li, Song, Xu, Tang, Zhuge, Pan,
  Song, Li, Singh, Tran, Li, Ma, Zheng, Qian, Shao, Muennighoff, Zhang, Hui,
  Lin, Brennan, Peng, Ji, and Neubig]{wang2024openhands}
Xingyao Wang, Boxuan Li, Yufan Song, Frank~F. Xu, Xiangru Tang, Mingchen Zhuge,
  Jiayi Pan, Yueqi Song, Bowen Li, Jaskirat Singh, Hoang~H. Tran, Fuqiang Li,
  Ren Ma, Mingzhang Zheng, Bill Qian, Yanjun Shao, Niklas Muennighoff, Yizhe
  Zhang, Binyuan Hui, Junyang Lin, Robert Brennan, Hao Peng, Heng Ji, and
  Graham Neubig.
\newblock {OpenHands: An Open Platform for AI Software Developers as Generalist
  Agents}, 2024{\natexlab{b}}.
\newblock URL \url{https://arxiv.org/abs/2407.16741}.

\bibitem[Xia et~al.(2024)Xia, Deng, Dunn, and Zhang]{xia2024agentless}
Chunqiu~Steven Xia, Yinlin Deng, Soren Dunn, and Lingming Zhang.
\newblock {Agentless: Demystifying LLM-based Software Engineering Agents},
  2024.
\newblock URL \url{https://arxiv.org/abs/2407.01489}.

\bibitem[Yang et~al.(2024)Yang, Jimenez, Wettig, Lieret, Yao, Narasimhan, and
  Press]{yang2024sweagent}
John Yang, Carlos~E. Jimenez, Alexander Wettig, Kilian Lieret, Shunyu Yao,
  Karthik Narasimhan, and Ofir Press.
\newblock {SWE-agent: Agent-Computer Interfaces Enable Automated Software
  Engineering}, 2024.
\newblock URL \url{https://arxiv.org/abs/2405.15793}.

\bibitem[Yao et~al.(2022)Yao, Zhao, Yu, Du, Shafran, Narasimhan, and
  Cao]{yao2022react}
Shunyu Yao, Jeffrey Zhao, Dian Yu, Nan Du, Izhak Shafran, Karthik Narasimhan,
  and Yuan Cao.
\newblock {ReAct: Synergizing Reasoning and Acting in Language Models}, 2022.
\newblock URL \url{https://arxiv.org/abs/2210.03629}.

\bibitem[{Z.ai}(2026)]{zai2026glm52}
{Z.ai}.
\newblock {GLM-5.2}.
\newblock Official model card, 2026.
\newblock URL \url{https://huggingface.co/zai-org/GLM-5.2}.
\newblock Accessed September 18, 2026.

\bibitem[Zhu et~al.(2025)Zhu, Li, Wu, Xing, Ma, Tang, Liu, Yang, Liu, Jiang,
  Zhang, Lin, Wang, Zhang, and Zhou]{zhu2025agentscaling}
King Zhu, Hanhao Li, Siwei Wu, Tianshun Xing, Dehua Ma, Xiangru Tang, Minghao
  Liu, Jian Yang, Jiaheng Liu, Yuchen~Eleanor Jiang, Changwang Zhang, Chenghua
  Lin, Jun Wang, Ge~Zhang, and Wangchunshu Zhou.
\newblock {Scaling Test-time Compute for LLM Agents}, 2025.
\newblock URL \url{https://arxiv.org/abs/2506.12928}.

\end{thebibliography}
\bibliographystyle{iclr2027_conference}
\clearpage
\appendix
\section*{Appendix guide}
\begingroup
\small
\renewcommand{\arraystretch}{1.15}
\begin{tabular}{@{}p{.48\linewidth}@{\hspace{.04\linewidth}}p{.48\linewidth}@{}}
\ref{app:experimental-details}~~\hyperref[app:experimental-details]{Experimental details}\hfill\pageref{app:experimental-details} & \ref{app:cases}~~\hyperref[app:cases]{Case study}\hfill\pageref{app:cases} \\
\ref{app:modelresults}~~\hyperref[app:modelresults]{Additional coding results}\hfill\pageref{app:modelresults} & \ref{app:instructions}~~\hyperref[app:instructions]{Prompts and execution rules}\hfill\pageref{app:instructions} \\
\ref{app:human-eval}~~\hyperref[app:human-eval]{VisualBench tasks and evaluation}\hfill\pageref{app:human-eval} &
\end{tabular}
\endgroup

\section{Additional experimental details}
\label{app:experimental-details}
\subsection{Task selection for Terminal-Bench and SWE-bench}
\label{app:taskselection}
We select tasks to balance coverage across easy, intermediate, and hard difficulty levels, enabling us to better observe both improvements and regressions.

\noindent\textbf{Terminal-Bench.}
We began with a baseline-harness pilot over the 89-task Terminal-Bench registry recorded as version 2.1. The pilot used Qwen3.7-Max, GLM-5.2, DeepSeek V4-Pro, and Kimi-K2.7-Code. Their pass rates guided selection across three empirical difficulty bands: tasks that were usually solved, solved inconsistently, or usually failed. We aimed for balanced coverage of these bands, yielding the 31-task study set.

\noindent\textbf{SWE-bench Verified.}
We initially selected 30 tasks evenly across three human time-to-fix tiers: under 15 minutes, 15 minutes to one hour, and over one hour. We then evaluated baseline performance and adjusted the task set to include both tasks with room for improvement and consistently solved tasks for detecting regressions.

\Needspace{20\baselineskip}
\subsection{API cost pricing}
\label{app:cost}
Table~\ref{tab:cost-rates} lists the model rates used to calculate cost, with Alibaba Cloud Model Studio pricing as the reference \citep{alibaba2026pricing,alibaba2026cache}.

\begingroup
\setlength{\intextsep}{6pt}
\begin{table}[!ht]
\centering
\caption{Rates used to calculate cost, in USD per million tokens.}
\label{tab:cost-rates}
\small
\begin{tabular}{lrrr}
\toprule
\textbf{Model} & \textbf{Uncached input} & \textbf{Cached input} & \textbf{Output} \\
\midrule
Qwen3.7-Max & 1.791 & 0.358 & 5.373 \\
Qwen3.8-Max & 1.791 & 0.358 & 5.373 \\
GLM-5.2 & 1.194 & 0.299 & 4.179 \\
DeepSeek V4-Pro & 1.791 & 0.149 & 3.582 \\
DeepSeek V4-Flash & 0.149 & 0.030 & 0.299 \\
Kimi-K2.7-Code & 0.970 & 0.194 & 4.030 \\
MiniMax-M2.5 & 0.313 & 0.063 & 1.254 \\
\midrule
Qwen3.7-Plus (critic) & 0.299 & 0.060 & 1.194 \\
\bottomrule
\end{tabular}
\end{table}

\FloatBarrier
\endgroup

\section{Additional coding results}
\label{app:modelresults}
We report accuracy, cost, and token use by verification design and coding model, followed by behavioral diagnostics for L0--L2.

\subsection{Absolute accuracy by verification design}
\begin{figure}[!ht]
\centering
\includegraphics[width=\linewidth]{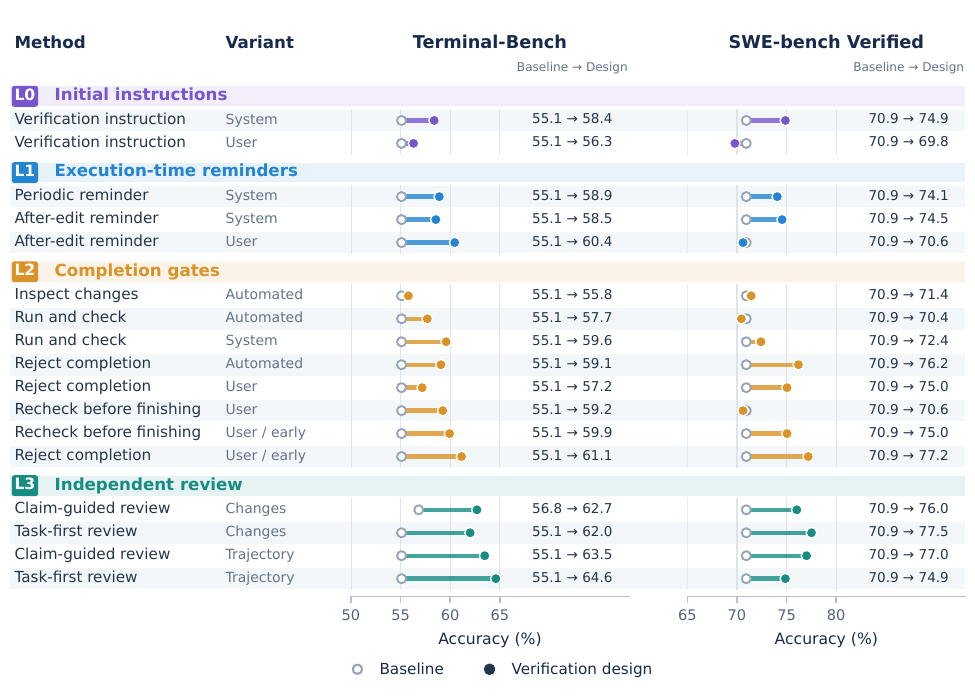}
\caption{Absolute accuracy on TB and SWE, with baseline and design values shown alongside each comparison.}
\label{fig:policy-detail}
\end{figure}

Figure~\ref{fig:policy-detail} reports accuracy by verification design, with open markers for the baseline and filled markers for the design. The corresponding absolute accuracies are listed alongside.

\FloatBarrier

\clearpage
Figures~\ref{fig:policy-qwen}--\ref{fig:policy-minimax} show the same comparison for each coding model.
\begingroup
\setlength{\intextsep}{6pt}
\setlength{\textfloatsep}{6pt}
\begin{figure}[!ht]
\centering
\includegraphics[width=\linewidth]{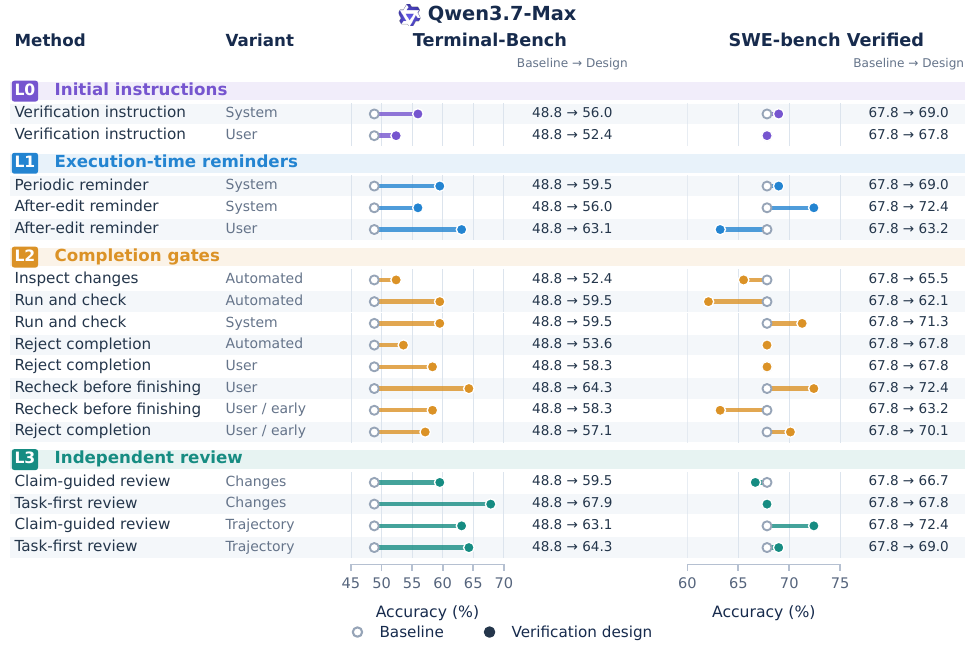}
\caption{Absolute accuracy of Qwen3.7-Max on TB and SWE, with baseline and design values shown for each comparison.}
\label{fig:policy-qwen}
\end{figure}
\begin{figure}[!ht]
\centering
\includegraphics[width=\linewidth]{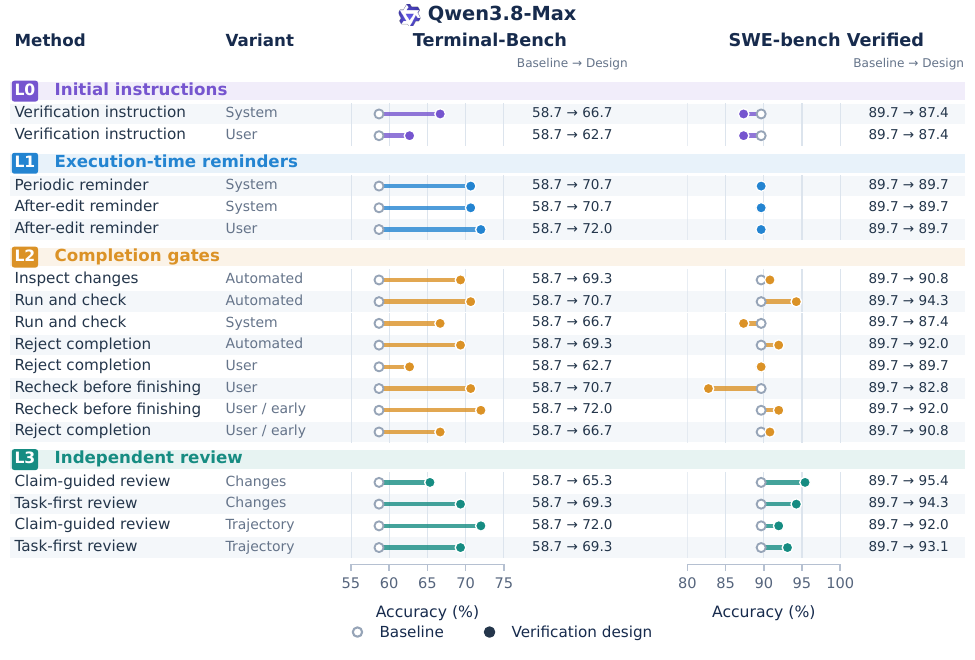}
\caption{Absolute accuracy of Qwen3.8-Max on TB and SWE, with baseline and design values shown for each comparison.}
\label{fig:policy-q38max}
\end{figure}
\FloatBarrier
\clearpage
\begin{figure}[!ht]
\centering
\includegraphics[width=\linewidth]{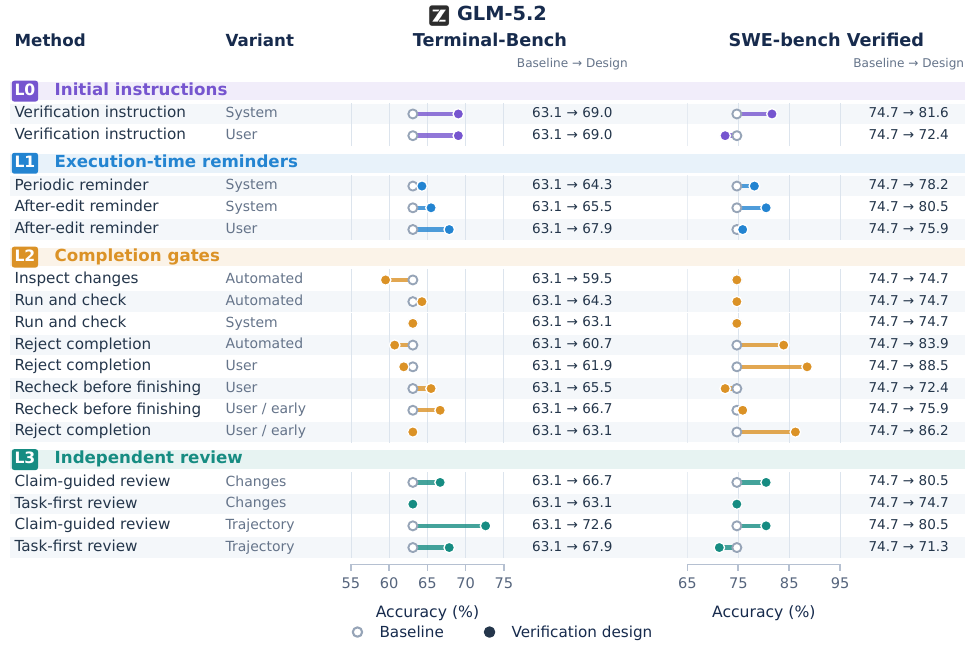}
\caption{Absolute accuracy of GLM-5.2 on TB and SWE, with baseline and design values shown for each comparison.}
\label{fig:policy-glm}
\end{figure}
\begin{figure}[!ht]
\centering
\includegraphics[width=\linewidth]{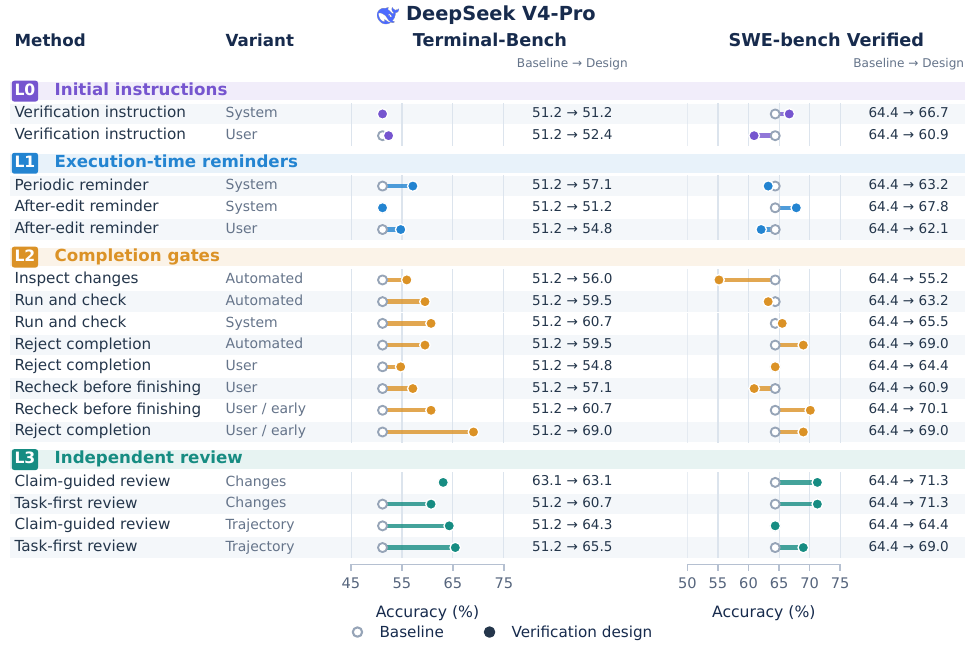}
\caption{Absolute accuracy of DeepSeek V4-Pro on TB and SWE, with baseline and design values shown for each comparison.}
\label{fig:policy-dsk}
\end{figure}
\FloatBarrier
\clearpage
\begin{figure}[!ht]
\centering
\includegraphics[width=\linewidth]{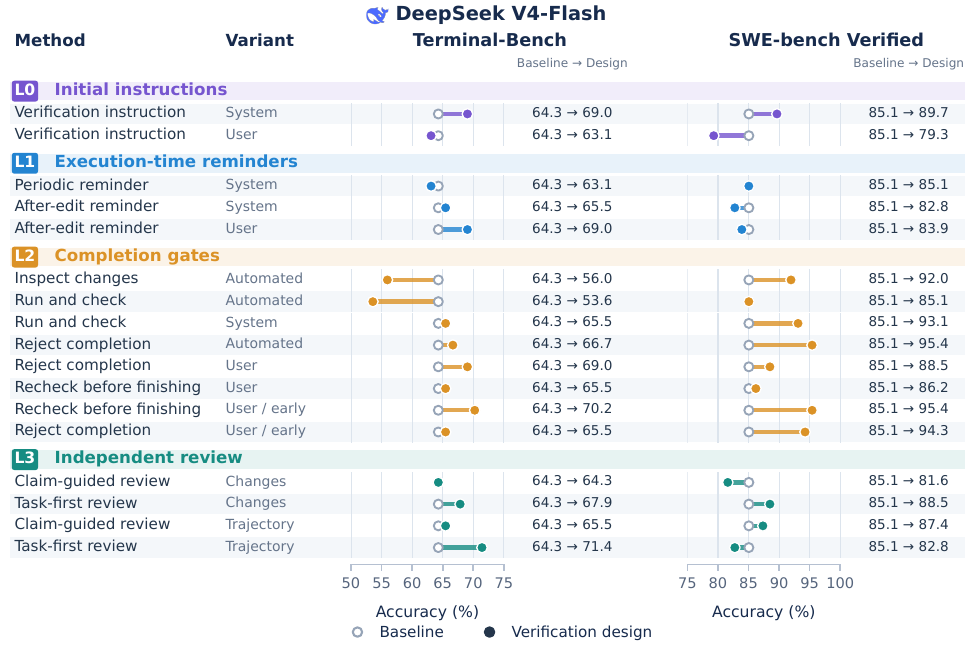}
\caption{Absolute accuracy of DeepSeek V4-Flash on TB and SWE, with baseline and design values shown for each comparison.}
\label{fig:policy-dskflash}
\end{figure}
\begin{figure}[!ht]
\centering
\includegraphics[width=\linewidth]{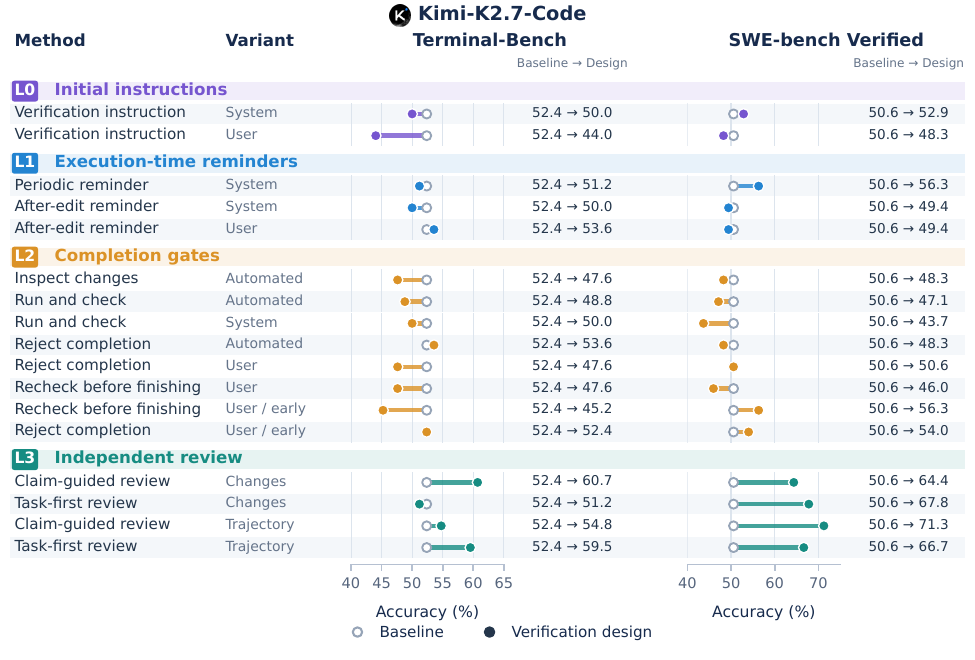}
\caption{Absolute accuracy of Kimi-K2.7-Code on TB and SWE, with baseline and design values shown for each comparison.}
\label{fig:policy-kimi}
\end{figure}
\FloatBarrier
\clearpage
\begin{figure}[!ht]
\centering
\includegraphics[width=\linewidth]{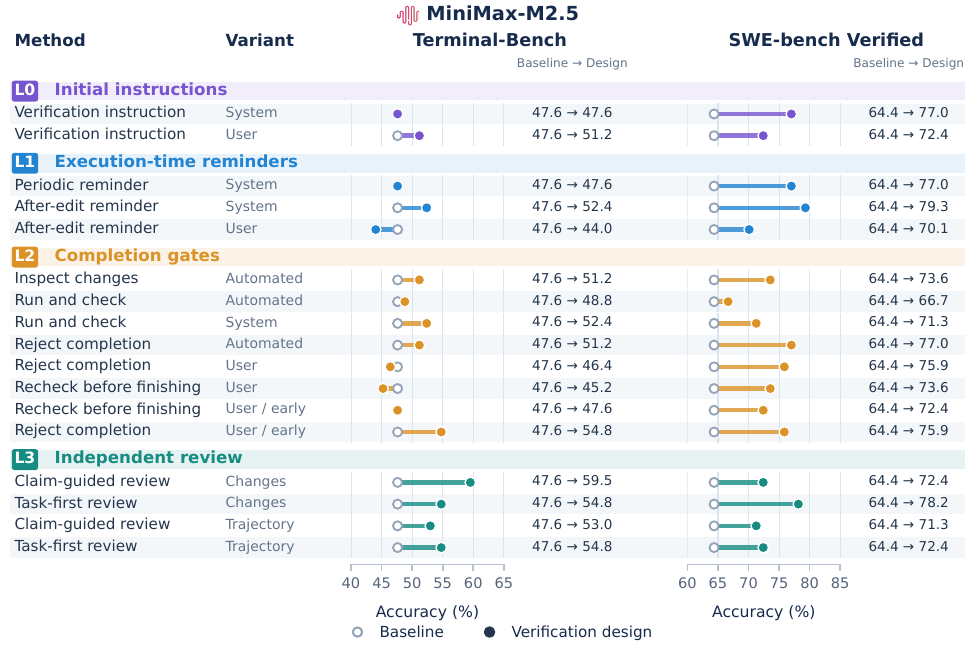}
\caption{Absolute accuracy of MiniMax-M2.5 on TB and SWE, with baseline and design values shown for each comparison.}
\label{fig:policy-minimax}
\end{figure}
\FloatBarrier
\endgroup

\FloatBarrier

\clearpage
\subsection{Accuracy and cost by coding model}
\label{app:model-cost}

\begin{figure}[!ht]
\centering
\includegraphics[width=\linewidth]{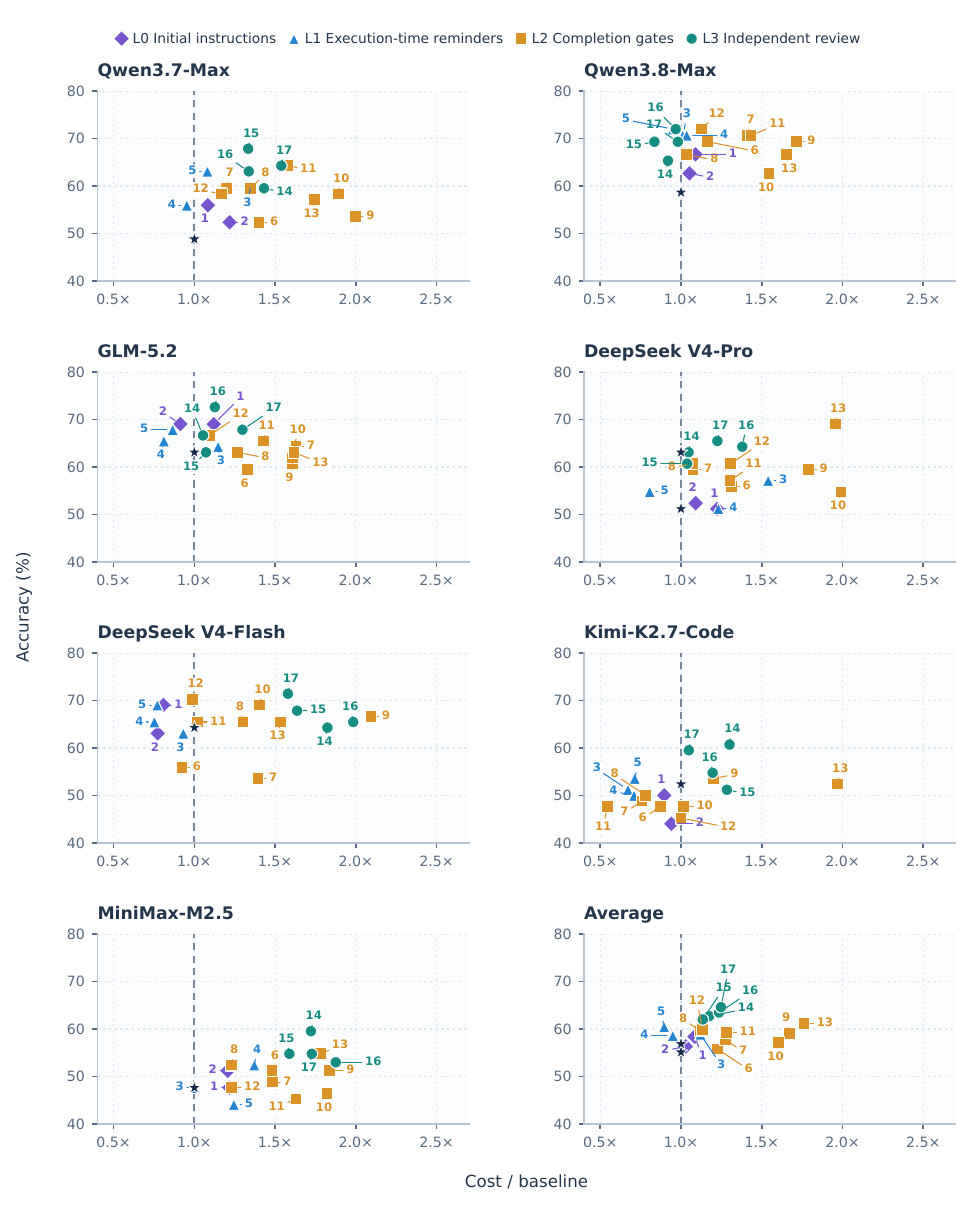}
\caption{Accuracy versus cost relative to baseline on Terminal-Bench, shown for each coding model and their average.}
\label{fig:model-cost-tb}
\end{figure}

\begin{figure}[p]
\centering
\includegraphics[width=\linewidth]{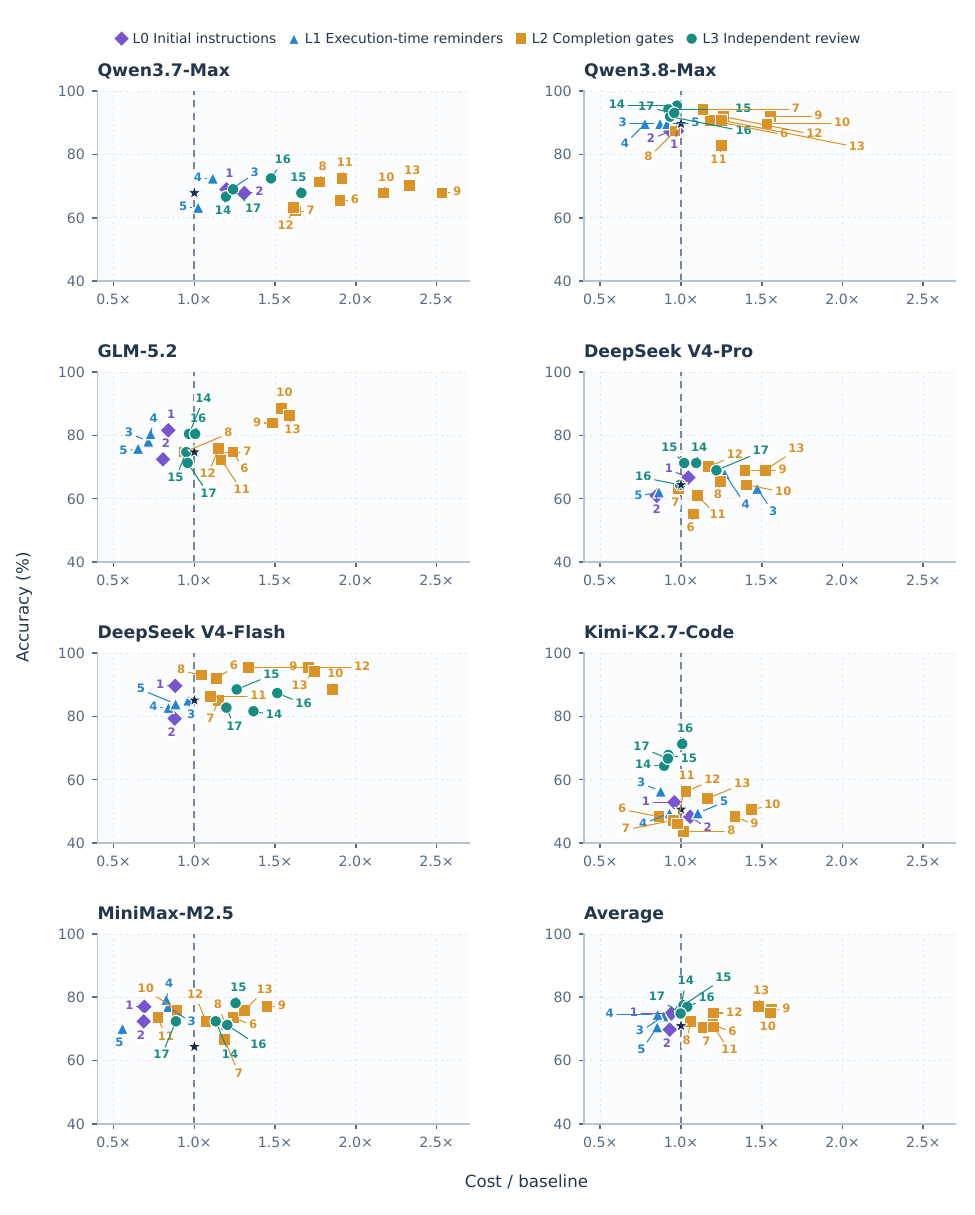}
\caption{Accuracy versus cost relative to baseline on SWE-bench Verified, shown for each coding model and their average.}
\label{fig:model-cost-swe}
\end{figure}

Figures~\ref{fig:model-cost-tb} and~\ref{fig:model-cost-swe} separate Figure~\ref{fig:main} by coding model, using the same trials, prices, and baseline. Each figure includes an Average panel reproducing Figure~\ref{fig:main}, and each panel labels all 17 verification designs by ID. Black stars mark baseline accuracy, and dashed lines mark $1\times$ baseline cost. Axis ranges are shared across models within each benchmark.

Table~\ref{tab:cost-arm-key} lists the design IDs and settings.
\begin{table}[!ht]
\centering\small
\caption{Verification design IDs and settings.}
\label{tab:cost-arm-key}
\renewcommand{\arraystretch}{1.12}
\begin{tabular*}{\linewidth}{@{}r@{\hspace{12pt}}l@{\extracolsep{\fill}}l@{}}
\toprule
\textbf{ID} & \textbf{Method} & \textbf{Variant} \\
\midrule
\rowcolor{layerzero!11}\multicolumn{3}{l}{\textcolor{layerzero}{\textbf{L0 Initial instructions}}} \\
1 & Verification instruction & System \\
2 & Verification instruction & User \\
\addlinespace[2pt]
\rowcolor{layerone!11}\multicolumn{3}{l}{\textcolor{layerone}{\textbf{L1 Execution-time reminders}}} \\
3 & Periodic reminder & System \\
4 & After-edit reminder & System \\
5 & After-edit reminder & User \\
\addlinespace[2pt]
\rowcolor{layertwo!11}\multicolumn{3}{l}{\textcolor{layertwo}{\textbf{L2 Completion gates}}} \\
6 & Inspect changes & Automated \\
7 & Run and check & Automated \\
8 & Run and check & System \\
9 & Reject completion & Automated \\
10 & Reject completion & User \\
11 & Recheck before finishing & User \\
12 & Recheck before finishing & User / early \\
13 & Reject completion & User / early \\
\addlinespace[2pt]
\rowcolor{layerthree!11}\multicolumn{3}{l}{\textcolor{layerthree}{\textbf{L3 Independent review}}} \\
14 & Claim-guided review & Changes \\
15 & Task-first review & Changes \\
16 & Claim-guided review & Trajectory \\
17 & Task-first review & Trajectory \\
\bottomrule
\end{tabular*}
\end{table}

\FloatBarrier

\clearpage

\clearpage

\clearpage
\subsection{Agent and critic token use}
\label{app:token-use}
\begin{figure}[!ht]
\centering
\includegraphics[width=\linewidth]{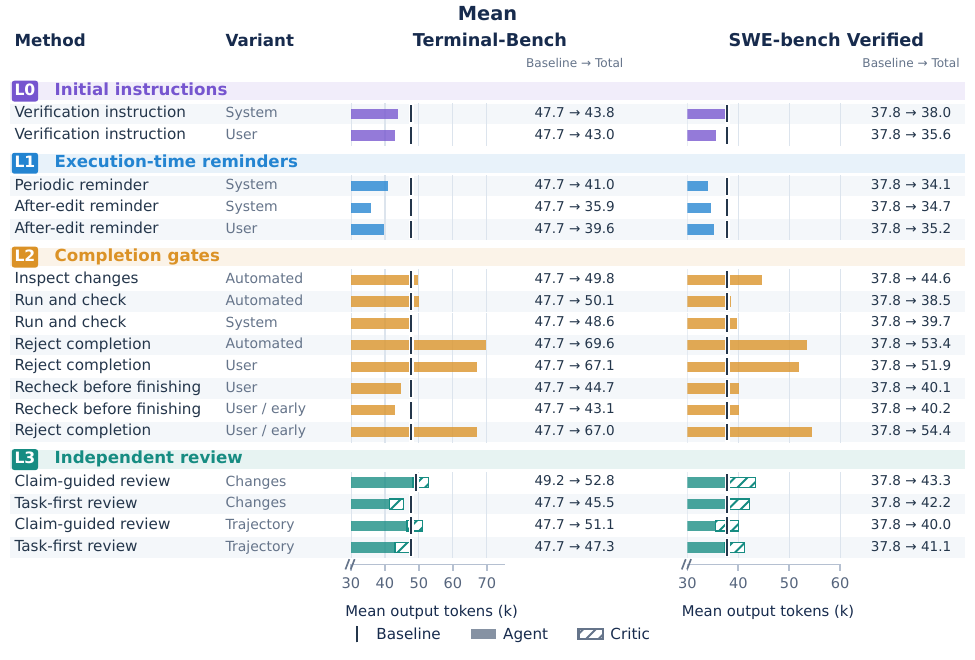}
\caption{Mean agent and critic output tokens on TB and SWE, averaged across model--task pairs.}
\label{fig:token-components}
\end{figure}

Figure~\ref{fig:token-components} shows mean output token use for all verification designs. Solid bars show agent generation, including revisions, and hatched segments show critic generation. Vertical ticks mark the baseline, with baseline and total values listed in thousands.

\FloatBarrier

\clearpage
Figures~\ref{fig:tokens-qwen}--\ref{fig:tokens-minimax} show the same comparison for each coding model.
\begingroup
\setlength{\intextsep}{6pt}
\setlength{\textfloatsep}{6pt}
\begin{figure}[!ht]
\centering
\includegraphics[width=\linewidth]{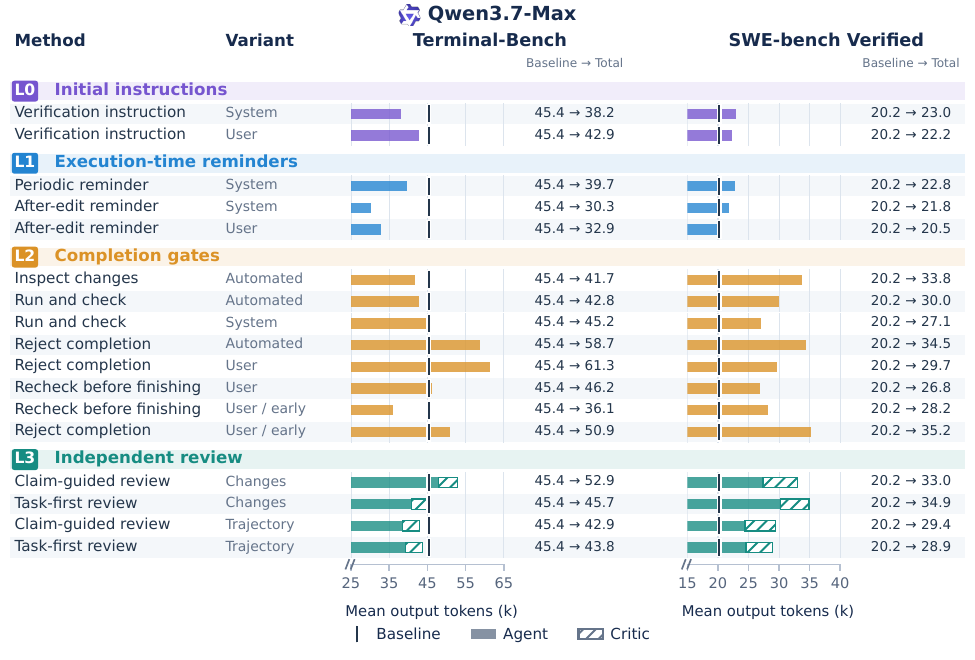}
\caption{Mean agent and critic output tokens for Qwen3.7-Max on TB and SWE, with baseline and total values shown for each design.}
\label{fig:tokens-qwen}
\end{figure}
\begin{figure}[!ht]
\centering
\includegraphics[width=\linewidth]{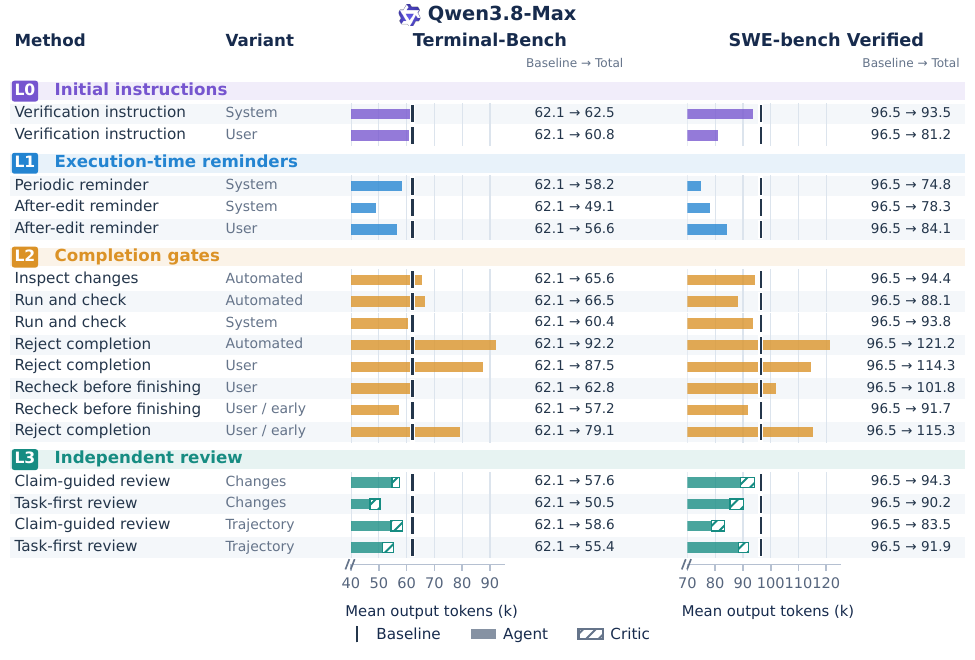}
\caption{Mean agent and critic output tokens for Qwen3.8-Max on TB and SWE, with baseline and total values shown for each design.}
\label{fig:tokens-q38max}
\end{figure}
\FloatBarrier
\clearpage
\begin{figure}[!ht]
\centering
\includegraphics[width=\linewidth]{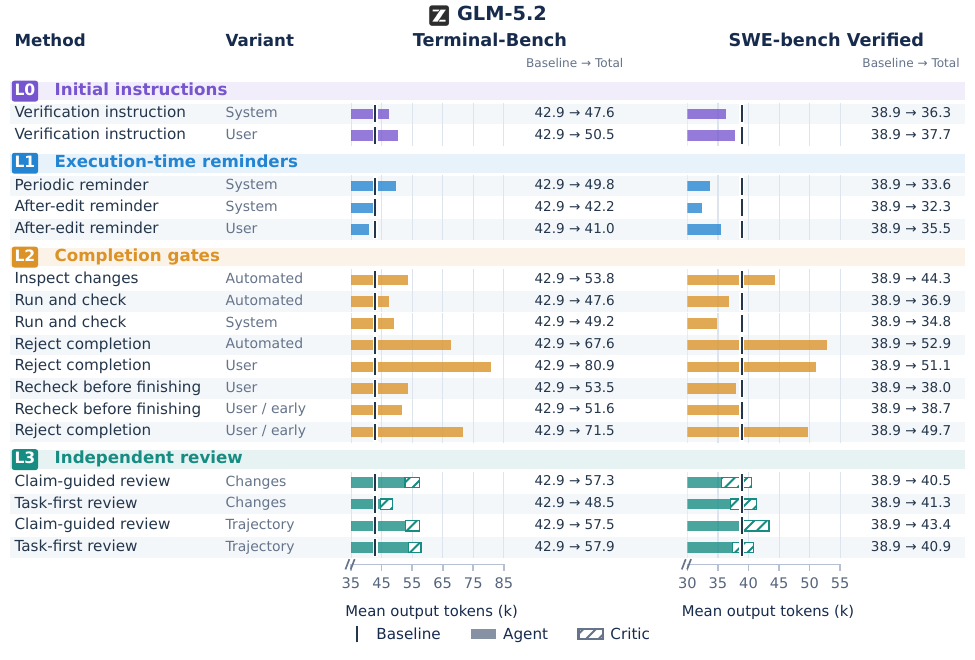}
\caption{Mean agent and critic output tokens for GLM-5.2 on TB and SWE, with baseline and total values shown for each design.}
\label{fig:tokens-glm}
\end{figure}
\begin{figure}[!ht]
\centering
\includegraphics[width=\linewidth]{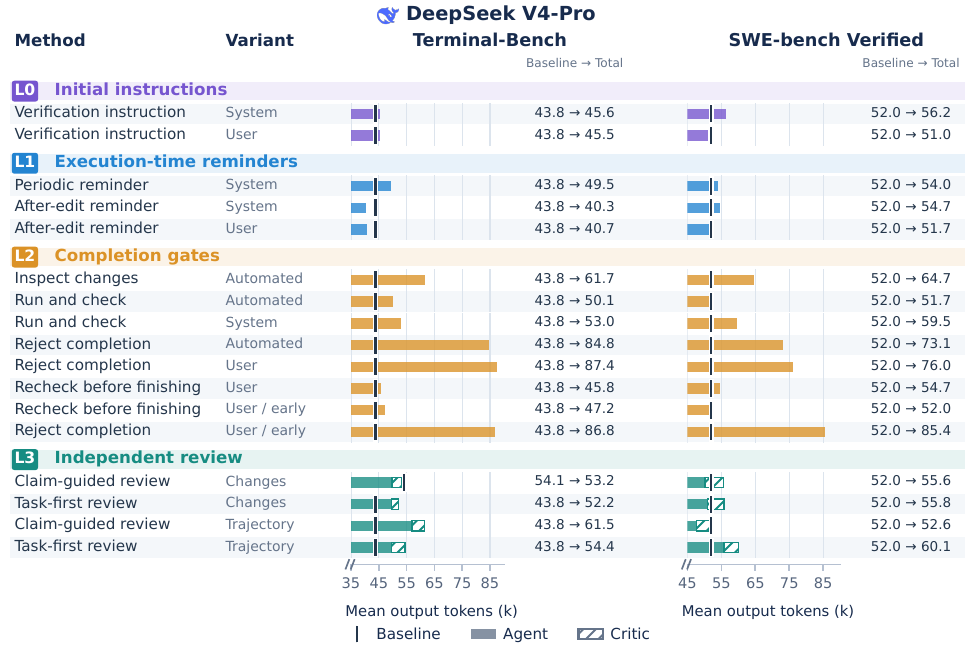}
\caption{Mean agent and critic output tokens for DeepSeek V4-Pro on TB and SWE, with baseline and total values shown for each design.}
\label{fig:tokens-dsk}
\end{figure}
\FloatBarrier
\clearpage
\begin{figure}[!ht]
\centering
\includegraphics[width=\linewidth]{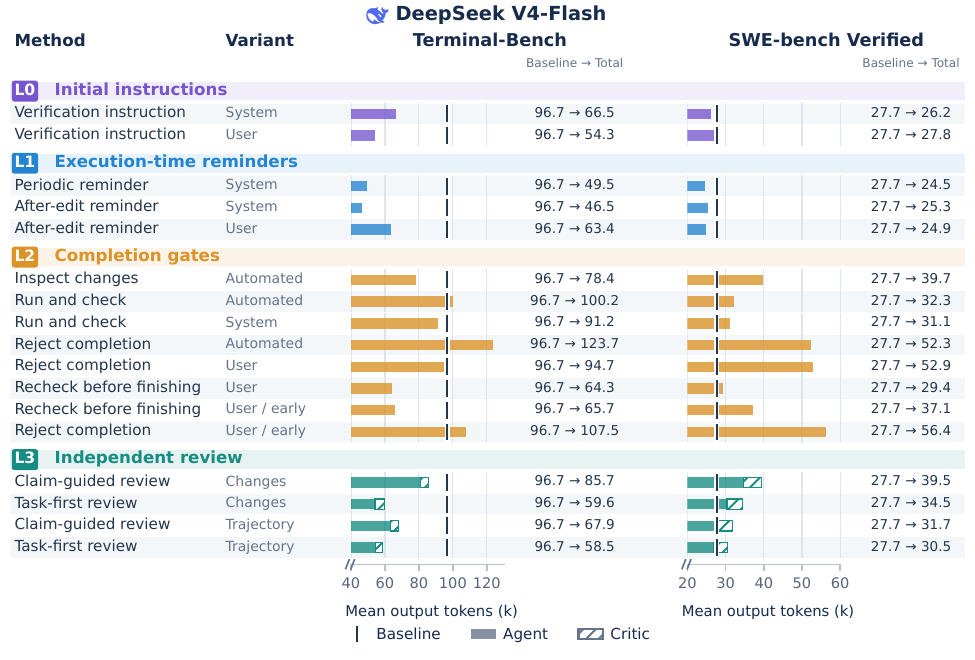}
\caption{Mean agent and critic output tokens for DeepSeek V4-Flash on TB and SWE, with baseline and total values shown for each design.}
\label{fig:tokens-dskflash}
\end{figure}
\begin{figure}[!ht]
\centering
\includegraphics[width=\linewidth]{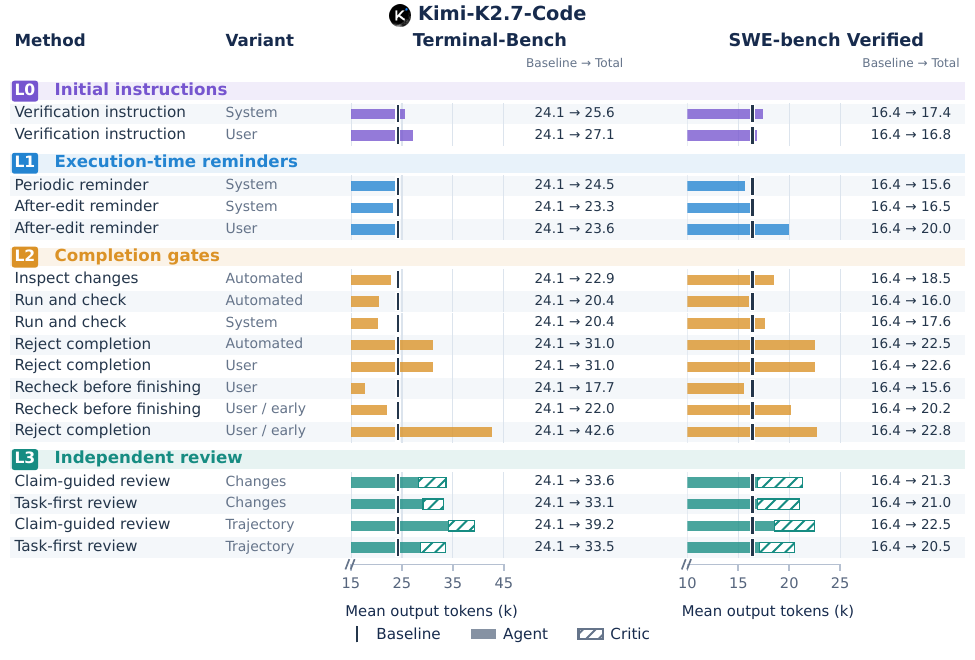}
\caption{Mean agent and critic output tokens for Kimi-K2.7-Code on TB and SWE, with baseline and total values shown for each design.}
\label{fig:tokens-kimi}
\end{figure}
\FloatBarrier
\clearpage
\begin{figure}[!ht]
\centering
\includegraphics[width=\linewidth]{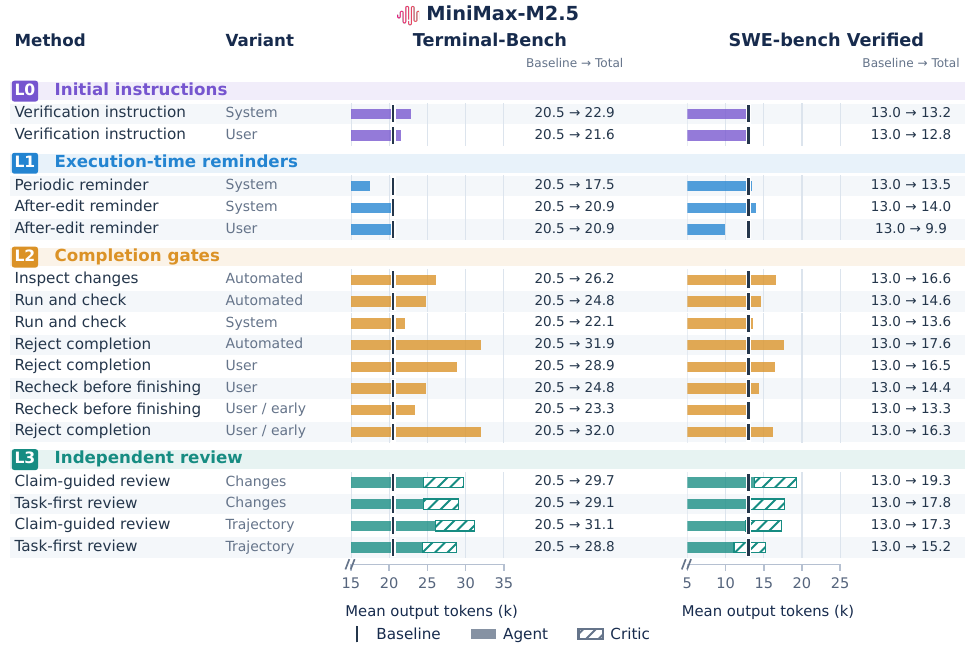}
\caption{Mean agent and critic output tokens for MiniMax-M2.5 on TB and SWE, with baseline and total values shown for each design.}
\label{fig:tokens-minimax}
\end{figure}
\FloatBarrier
\endgroup

\FloatBarrier

\clearpage
\subsection{Review scaling and total cost}
\label{app:scaling-cost}
\begin{figure}[!ht]
\centering
\includegraphics[width=\linewidth]{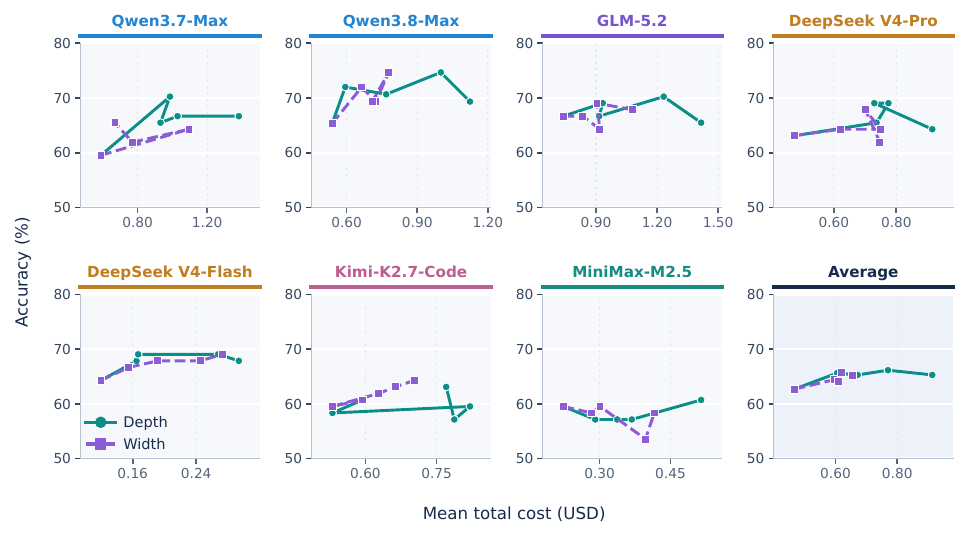}
\caption{TB accuracy versus mean total cost as review depth and width increase.}
\label{fig:scale-cost}
\end{figure}

Figure~\ref{fig:scale-cost} reports accuracy against mean total cost, including agent execution, critic checks, and subsequent revisions. One critic and one round achieve $62.7\%$ accuracy at \$0.47 per trial. Two rounds reach $65.6\%$ at \$0.61, while two critics reach $64.4\%$ at \$0.60. Four critics achieve $65.8\%$ at \$0.62, compared with $66.1\%$ at \$0.77 for four rounds. The critic costs for four critics and four rounds are both approximately \$0.14 per trial, while their agent-side costs, including revisions, are \$0.48 and \$0.63, respectively. The cost difference therefore comes primarily from agent execution, consistent with wider review consolidating feedback before revision instead of repeatedly continuing the agent between review rounds. A second round raises Qwen3.7-Max from $59.5\%$ to $70.2\%$, but lowers Kimi-K2.7-Code from $60.7\%$ to $58.3\%$.

\FloatBarrier

\subsection{L0: Output tokens and agent turns}
\label{app:mechanisms}
Table~\ref{tab:l0-activity} reports mean output tokens and agent turns relative to baseline on TB and SWE. Values above one indicate more generated output or longer interactions.

{\scriptsize\setlength{\tabcolsep}{3pt}\renewcommand{\arraystretch}{1.2}
\begin{longtable}{@{}p{2.9cm}rrrrrrrr@{}}
\caption{Output tokens and agent turns for initial verification instructions, expressed as multiples of baseline.}\label{tab:l0-activity}\\
\toprule
\textbf{Metric} & \modelhead{qwen}{Qwen}{3.7-Max} & \modelhead{qwen}{Qwen}{3.8-Max} & \modelhead{glm}{GLM}{5.2} & \modelhead{deepseek}{DeepSeek}{V4-Pro} & \modelhead{deepseek}{DeepSeek}{V4-Flash} & \modelhead{kimi}{Kimi}{K2.7 Code} & \modelhead{minimax}{MiniMax}{M2.5} & \textbf{Average} \\
\midrule\endfirsthead
\toprule
\textbf{Metric} & \modelhead{qwen}{Qwen}{3.7-Max} & \modelhead{qwen}{Qwen}{3.8-Max} & \modelhead{glm}{GLM}{5.2} & \modelhead{deepseek}{DeepSeek}{V4-Pro} & \modelhead{deepseek}{DeepSeek}{V4-Flash} & \modelhead{kimi}{Kimi}{K2.7 Code} & \modelhead{minimax}{MiniMax}{M2.5} & \textbf{Average} \\
\midrule\endhead
\bottomrule\endfoot
\rowcolor{layerzero!8}\multicolumn{9}{l}{\textcolor{layerzero}{\textbf{Terminal-Bench: System}}} \\
\# output tokens & 0.84 & 1.01 & 1.11 & 1.04 & 0.69 & 1.06 & 1.12 & 0.92 \\
\# agent turns & 1.06 & 0.95 & 0.96 & 1.04 & 0.94 & 0.80 & 0.95 & 0.94 \\
\rowcolor{layerzero!8}\multicolumn{9}{l}{\textcolor{layerzero}{\textbf{Terminal-Bench: User}}} \\
\# output tokens & 0.95 & 0.98 & 1.18 & 1.04 & 0.56 & 1.12 & 1.05 & 0.90 \\
\# agent turns & 1.17 & 0.96 & 0.90 & 1.00 & 0.99 & 0.79 & 1.07 & 0.96 \\
\rowcolor{layerzero!8}\multicolumn{9}{l}{\textcolor{layerzero}{\textbf{SWE-bench Verified: System}}} \\
\# output tokens & 1.14 & 0.97 & 0.93 & 1.08 & 0.94 & 1.06 & 1.02 & 1.00 \\
\# agent turns & 1.05 & 0.98 & 0.97 & 1.02 & 0.93 & 0.91 & 0.94 & 0.95 \\
\rowcolor{layerzero!8}\multicolumn{9}{l}{\textcolor{layerzero}{\textbf{SWE-bench Verified: User}}} \\
\# output tokens & 1.10 & 0.84 & 0.97 & 0.98 & 1.00 & 1.03 & 0.98 & 0.94 \\
\# agent turns & 1.16 & 0.95 & 0.97 & 0.93 & 0.97 & 1.04 & 0.91 & 0.99 \\
\end{longtable}
}

\FloatBarrier

\Needspace{16\baselineskip}
\subsection{L1: Checking during execution}
\label{app:spontaneous-checks}
Table~\ref{tab:spontaneous-checks} reports how often TB baseline agents check their work after editing. We count recognized checking commands before the next editing turn or the recorded run end.

\begin{table}[!ht]
\centering\small
\caption{Editing turns with recognized checks in TB baseline runs (\%).}
\label{tab:spontaneous-checks}
\begin{tabular}{lrr}
\toprule
\textbf{Model} & \textbf{Later turn} & \textbf{Same or later turn} \\
\midrule
Qwen3.7-Max & 9.7 & 10.3 \\
Qwen3.8-Max & 8.7 & 16.5 \\
GLM-5.2 & 11.0 & 12.3 \\
DeepSeek V4-Pro & 6.2 & 9.5 \\
DeepSeek V4-Flash & 10.5 & 13.3 \\
Kimi-K2.7-Code & 6.8 & 7.3 \\
MiniMax-M2.5 & 13.6 & 13.8 \\
\bottomrule
\end{tabular}
\end{table}

\FloatBarrier

\Needspace{18\baselineskip}
Table~\ref{tab:reminder-diagnostics} reports reminder coverage and checking or execution within the next two agent turns. After-edit user reminders reach $80$--$100\%$ of runs, with recognized checks following $58$--$72\%$ of reminder events.

{\scriptsize\setlength{\tabcolsep}{3pt}\renewcommand{\arraystretch}{1.17}
\begin{longtable}{@{}p{4.2cm}rrrrrrr@{}}
\caption{Reminder coverage and subsequent checks on Terminal-Bench.}\label{tab:reminder-diagnostics}\\
\toprule
\textbf{Diagnostic} & \modelhead{qwen}{Qwen}{3.7-Max} & \modelhead{qwen}{Qwen}{3.8-Max} & \modelhead{glm}{GLM}{5.2} & \modelhead{deepseek}{DeepSeek}{V4-Pro} & \modelhead{deepseek}{DeepSeek}{V4-Flash} & \modelhead{kimi}{Kimi}{K2.7 Code} & \modelhead{minimax}{MiniMax}{M2.5} \\
\midrule\endfirsthead
\toprule
\textbf{Diagnostic} & \modelhead{qwen}{Qwen}{3.7-Max} & \modelhead{qwen}{Qwen}{3.8-Max} & \modelhead{glm}{GLM}{5.2} & \modelhead{deepseek}{DeepSeek}{V4-Pro} & \modelhead{deepseek}{DeepSeek}{V4-Flash} & \modelhead{kimi}{Kimi}{K2.7 Code} & \modelhead{minimax}{MiniMax}{M2.5} \\
\midrule\endhead
\bottomrule\endfoot
\rowcolor{layerone!8}\multicolumn{8}{l}{\textcolor{layerone}{\textbf{Periodic reminder (system)}}} \\
Runs receiving a reminder (\%) & 73 & 89 & 82 & 67 & 86 & 66 & 87 \\
Reached the four-reminder cap (\%) & 62 & 55 & 65 & 59 & 72 & 64 & 74 \\
Check within two turns (\%) & 70 & 81 & 67 & 75 & 66 & 65 & 71 \\
Execution within two turns (\%) & 92 & 96 & 95 & 91 & 87 & 90 & 91 \\
\rowcolor{layerone!8}\multicolumn{8}{l}{\textcolor{layerone}{\textbf{After-edit reminder (user)}}} \\
Runs receiving a reminder (\%) & 90 & 100 & 94 & 80 & 100 & 80 & 96 \\
Reached the four-reminder cap (\%) & 55 & 75 & 81 & 63 & 71 & 70 & 65 \\
Check within two turns (\%) & 67 & 60 & 72 & 69 & 61 & 62 & 58 \\
Execution within two turns (\%) & 92 & 93 & 97 & 91 & 90 & 95 & 92 \\
\end{longtable}
}

\FloatBarrier

\subsection{L2: Work after a completion gate}
Table~\ref{tab:gate-diagnostics} reports subsequent steps and file-changing actions after L2 interventions. User rejection is followed by $12$--$19$ additional steps at the model median and file-changing actions in $86$--$96\%$ of runs where it is triggered.

{\scriptsize\setlength{\tabcolsep}{3pt}\renewcommand{\arraystretch}{1.17}
\begin{longtable}{@{}p{4.2cm}rrrrrrr@{}}
\caption{Completion-gate coverage and subsequent work on Terminal-Bench.}\label{tab:gate-diagnostics}\\
\toprule
\textbf{Diagnostic} & \modelhead{qwen}{Qwen}{3.7-Max} & \modelhead{qwen}{Qwen}{3.8-Max} & \modelhead{glm}{GLM}{5.2} & \modelhead{deepseek}{DeepSeek}{V4-Pro} & \modelhead{deepseek}{DeepSeek}{V4-Flash} & \modelhead{kimi}{Kimi}{K2.7 Code} & \modelhead{minimax}{MiniMax}{M2.5} \\
\midrule\endfirsthead
\toprule
\textbf{Diagnostic} & \modelhead{qwen}{Qwen}{3.7-Max} & \modelhead{qwen}{Qwen}{3.8-Max} & \modelhead{glm}{GLM}{5.2} & \modelhead{deepseek}{DeepSeek}{V4-Pro} & \modelhead{deepseek}{DeepSeek}{V4-Flash} & \modelhead{kimi}{Kimi}{K2.7 Code} & \modelhead{minimax}{MiniMax}{M2.5} \\
\midrule\endhead
\bottomrule\endfoot
\rowcolor{layertwo!8}\multicolumn{8}{l}{\textcolor{layertwo}{\textbf{Inspect changes (automated)}}} \\
Runs reaching the gate (\%) & 88 & 83 & 89 & 81 & 66 & 65 & 89 \\
Steps after the gate (median) & 4 & 4 & 6 & 6 & 5 & 4 & 5 \\
File-changing action after gate (\%) & 38 & 53 & 66 & 52 & 62 & 49 & 53 \\
\rowcolor{layertwo!8}\multicolumn{8}{l}{\textcolor{layertwo}{\textbf{Recheck before finishing (user)}}} \\
Runs reaching the gate (\%) & 92 & 84 & 93 & 79 & 87 & 50 & 92 \\
Steps after the gate (median) & 6 & 6 & 8 & 8 & 8 & 4 & 5 \\
File-changing action after gate (\%) & 66 & 95 & 87 & 76 & 85 & 60 & 57 \\
\rowcolor{layertwo!8}\multicolumn{8}{l}{\textcolor{layertwo}{\textbf{Reject completion (user)}}} \\
Runs reaching the gate (\%) & 90 & 80 & 84 & 57 & 88 & 52 & 87 \\
Steps after the gate (median) & 14 & 12 & 19 & 16 & 16 & 15 & 19 \\
File-changing action after gate (\%) & 88 & 92 & 94 & 88 & 92 & 96 & 86 \\
\end{longtable}
}

\FloatBarrier

\clearpage
\section{VisualBench tasks and human evaluation}
\label{app:human-eval}
VisualBench contains 18 tasks spanning games, shaders, interactive three-dimensional scenes, applications, and other visual artifacts.

\subsection{Complete task catalog}
\label{app:visual-task-descriptions}
Tables~\ref{tab:visual-tasks-one} and~\ref{tab:visual-tasks-two} summarize all 18 tasks and their requested deliverables. Full task descriptions and task-specific interaction instructions follow the overview.
{\small\setlength{\tabcolsep}{5pt}\renewcommand{\arraystretch}{1.08}
\begin{longtable}{@{}>{\raggedright\arraybackslash}p{3.7cm}>{\raggedright\arraybackslash}p{\dimexpr\linewidth-3.7cm-10pt\relax}@{}}
\caption{VisualBench tasks and requested deliverables, part 1.}\label{tab:visual-tasks-one}\\
\toprule\textbf{Task} & \textbf{Core requirements and deliverables} \\
\midrule\endfirsthead
\multicolumn{2}{l}{\small\textit{VisualBench task catalog (continued)}} \\
\toprule\textbf{Task} & \textbf{Core requirements and deliverables} \\
\midrule\endhead
\bottomrule\endfoot
\textbf{Blender desk scene}\newline{\scriptsize\nolinkurl{blender-object}}\newline{\scriptsize\textcolor{layerone}{3D}} & Construct a desk lamp and supporting objects from geometry, with distinct materials, deliberate lighting, shadows, and depth of field. Render a composed image and a lit, full-rotation video of at least four seconds using Cycles.\newline{\footnotesize\textit{Deliverables:} \texttt{render.png}, \texttt{turntable.mp4}.} \\[0.35em]
\textbf{Breakout}\newline{\scriptsize\nolinkurl{game-breakout}}\newline{\scriptsize\textcolor{layerone}{Game}} & Build a playable brick-breaking game with paddle-dependent bounce and spin, multiple brick types, stacking power-ups, synthesized sound, combo scoring, and at least four levels. Include lives, pause, restart, and a persistent high score.\newline{\footnotesize\textit{Deliverables:} \texttt{index.html}.} \\[0.35em]
\textbf{Tower defense}\newline{\scriptsize\nolinkurl{game-tower}}\newline{\scriptsize\textcolor{layerone}{Game}} & Implement escalating enemy waves and at least four tower types with upgrades, selling, range displays, and placement constraints. Include varied enemies, an early-wave bonus, money and lives, speed control, pause, and win/loss states.\newline{\footnotesize\textit{Deliverables:} \texttt{index.html}.} \\[0.35em]
\textbf{Voxel sandbox}\newline{\scriptsize\nolinkurl{game-voxel}}\newline{\scriptsize\textcolor{layerone}{Game}} & Generate streamed terrain with caves, water, and distinct block types. Support first-person movement and collision, mining, inventory, placement, falling blocks, crafting, day/night lighting, and persistence of the player position and world edits.\newline{\footnotesize\textit{Deliverables:} \texttt{index.html}, \texttt{three.min.js}.} \\[0.35em]
\textbf{Stormy ocean}\newline{\scriptsize\nolinkurl{shader-scene}}\newline{\scriptsize\textcolor{layerone}{Shader}} & Write a WebGL/GLSL ocean shader with waves at multiple scales, persistent foam, scattering and reflections, moving clouds, rain, and lightning that relights the scene. Support mouse-controlled viewing and smooth full-screen rendering without loaded textures.\newline{\footnotesize\textit{Deliverables:} \texttt{index.html}.} \\[0.35em]
\textbf{Animated concert poster}\newline{\scriptsize\nolinkurl{svg-poster}}\newline{\scriptsize\textcolor{layerone}{Vector}} & Design a synthwave poster with the specified event text, deliberate typography, layered depth, and lighting effects. Combine animations at different speeds, including path-driven motion, in a seamless loop; keep the SVG self-contained and scalable.\newline{\footnotesize\textit{Deliverables:} \texttt{poster.svg}.} \\[0.35em]
\textbf{3D lamp configurator}\newline{\scriptsize\nolinkurl{three-config}}\newline{\scriptsize\textcolor{layerone}{3D}} & Construct an articulated lamp with live controls for shade shape, geometry, color, finish, bulb warmth, and brightness. Provide smooth transitions, an updated price and summary, saved settings, a shareable configuration URL, orbit controls, and PNG export.\newline{\footnotesize\textit{Deliverables:} \texttt{index.html}, \texttt{three.min.js}.} \\[0.35em]
\textbf{Product hero scene}\newline{\scriptsize\nolinkurl{three-scene}}\newline{\scriptsize\textcolor{layerone}{3D}} & Model wireless headphones from primitives with studio lighting and distinct materials. Add damped orbit and zoom, idle animation, smooth color changes, a labeled exploded view, and clickable hotspots that focus the camera on product details.\newline{\footnotesize\textit{Deliverables:} \texttt{index.html}, \texttt{three.min.js}.} \\[0.35em]
\textbf{Fleet terminal dashboard}\newline{\scriptsize\nolinkurl{tui-dashboard}}\newline{\scriptsize\textcolor{layerone}{Terminal UI}} & Build a live Python terminal interface from the supplied service data, with status, sparklines, overall health, scrolling logs, severity colors, and adaptive layout. Record at least 30 seconds of updates including a service status transition.\newline{\footnotesize\textit{Deliverables:} \texttt{demo.cast}, \texttt{screen.svg}.} \\[0.35em]
\end{longtable}
}
\clearpage
{\small\setlength{\tabcolsep}{5pt}\renewcommand{\arraystretch}{1.08}
\begin{longtable}{@{}>{\raggedright\arraybackslash}p{3.7cm}>{\raggedright\arraybackslash}p{\dimexpr\linewidth-3.7cm-10pt\relax}@{}}
\caption{VisualBench tasks and requested deliverables, part 2.}\label{tab:visual-tasks-two}\\
\toprule\textbf{Task} & \textbf{Core requirements and deliverables} \\
\midrule\endfirsthead
\multicolumn{2}{l}{\small\textit{VisualBench task catalog (continued)}} \\
\toprule\textbf{Task} & \textbf{Core requirements and deliverables} \\
\midrule\endhead
\bottomrule\endfoot
\textbf{Interactive cloth}\newline{\scriptsize\nolinkurl{cloth-sim}}\newline{\scriptsize\textcolor{layerzero}{3D}} & Implement cloth physics without a physics library: grabbing, unpinning, wind, tearing, and stable draping over a sphere. Expose simulation controls, live energy and constraint error, and structural and shaded views; remain stable under aggressive input.\newline{\footnotesize\textit{Deliverables:} \texttt{index.html}.} \\[0.35em]
\textbf{Flowline kanban}\newline{\scriptsize\nolinkurl{data-app-kanban}}\newline{\scriptsize\textcolor{layerzero}{Application}} & Create a populated board with editable cards, drag-and-drop, bulk actions, composed filters, and an enforced three-card work-in-progress limit. Support full keyboard operation, undo/redo, persistent board state, and JSON import/export.\newline{\footnotesize\textit{Deliverables:} \texttt{index.html}.} \\[0.35em]
\textbf{Sales data explorer}\newline{\scriptsize\nolinkurl{dataviz-explorer}}\newline{\scriptsize\textcolor{layerzero}{Data visualization}} & Visualize roughly 5,500 messy sales records across six regions and two channels. Enable trend and regional comparisons, repair recoverable data, explain every excluded row, and represent missing values honestly. Choose the views and implement them without charting libraries.\newline{\footnotesize\textit{Deliverables:} \texttt{index.html}.} \\[0.35em]
\textbf{2D platformer}\newline{\scriptsize\nolinkurl{game-platformer}}\newline{\scriptsize\textcolor{layerzero}{Game}} & Use the two supplied tile maps to build responsive movement with variable jump height and coyote time. Include a following camera, hazards and respawning, collectible coins, progression between levels, a win screen, and restart.\newline{\footnotesize\textit{Deliverables:} \texttt{index.html}.} \\[0.35em]
\textbf{Generative art series}\newline{\scriptsize\nolinkurl{generative-art}}\newline{\scriptsize\textcolor{layerzero}{Generative art}} & Create a visually coherent series with varied compositions and generated palettes. Make each seed deterministic and its composition independent of window size; provide seed navigation, progressive rendering, and PNG export rendered at least three times screen resolution.\newline{\footnotesize\textit{Deliverables:} \texttt{index.html}.} \\[0.35em]
\textbf{Vector boolean editor}\newline{\scriptsize\nolinkurl{path-boolean}}\newline{\scriptsize\textcolor{layerzero}{Vector}} & Compute union, intersection, difference, and exclusion on editable cubic Bezier paths without geometry libraries or masking shortcuts. Handle tangencies, overlaps, holes, and disjoint results; support chained operations, geometry diagnostics, undo/redo, and SVG export.\newline{\footnotesize\textit{Deliverables:} \texttt{index.html}.} \\[0.35em]
\textbf{City in a storm}\newline{\scriptsize\nolinkurl{shader-city}}\newline{\scriptsize\textcolor{layerzero}{Shader}} & Raymarch an endless nighttime skyline above moving reflective water using raw WebGL/GLSL. Add varied towers and windows, moving clouds, fog, rain, and spatially located lightning that relights the scene; support camera motion at a smooth frame rate.\newline{\footnotesize\textit{Deliverables:} \texttt{index.html}.} \\[0.35em]
\textbf{Gridwork spreadsheet}\newline{\scriptsize\nolinkurl{spreadsheet-engine}}\newline{\scriptsize\textcolor{layerzero}{Application}} & Build a grid of at least 40 by 20 cells with formulas, ranges, standard functions, and dependency-based recalculation. Handle cycles and invalid inputs without hanging; support fill-down, sorting, undo/redo, and persistence.\newline{\footnotesize\textit{Deliverables:} \texttt{index.html}.} \\[0.35em]
\textbf{GPU particle field}\newline{\scriptsize\nolinkurl{three-particles}}\newline{\scriptsize\textcolor{layerzero}{3D}} & Simulate one million particles on the GPU with evolving curl-noise flow, momentum, trails, and mouse attraction/repulsion. Convey depth and flow through brightness and color; recycle particles, show live count and frame rate, and handle resize without resetting.\newline{\footnotesize\textit{Deliverables:} \texttt{index.html}, \texttt{three.min.js} (optional).} \\[0.35em]
\end{longtable}
}
\clearpage

\clearpage
The following entries reproduce all 18 English task briefs and their task-specific rater instructions from the evaluation kit. Line wrapping and dash typography are normalized for typesetting. The task descriptions are supplied to the agent; the interaction instructions guide human assessment of the delivered artifacts.

\subsubsection{Blender desk scene}
\label{app:visual-task-blender-object}
{\small\noindent\texttt{blender-object}\par}
\promptheading{layerone}{Task description}
\begin{promptblock}\fontsize{9}{11}\selectfont
Using Blender, build a small desk scene in a script  -  geometry you construct yourself, no imported assets. An articulated desk lamp is the subject: base, jointed arm, shade, with the shade actually housing a light so the lamp lights the scene rather than being lit by it. Put it on a surface with a couple of supporting objects  -  a mug, a stack of books, something with a different silhouette  -  arranged as a composition rather than a row.

Materials should differ and read as materials: brushed metal on the lamp, glaze on the mug, paper on the books, with a surface underneath that isn't the default grey. Light it deliberately  -  the lamp as the key, something soft filling the shadows, and a rim to separate the subject from the background. I want real shadows and some depth of field.

Render with Cycles; EEVEE is unavailable in this headless environment. Watch your sample count and resolution so the render actually finishes.

Deliver, in your working directory: \textbf{\texttt{render.png}} (one composed beauty shot) and \textbf{\texttt{turntable.mp4}} (a smooth full 360$^\circ$, four seconds or more, with the lamp lit throughout). Optionally also export the model as \texttt{model.glb}.

\end{promptblock}
\promptheading{layerthree}{Human interaction instructions}
{\fontsize{9}{11}\selectfont
\begin{enumerate}\setlength{\itemsep}{2pt}\setlength{\parskip}{0pt}\setlength{\parsep}{0pt}
\item First, the objects in render.png: an articulated desk lamp  -  base, jointed arm, shade  -  with a mug and a stack of books beside it, arranged as a composition rather than a row? If the objects are wrong, the lighting below does not count
\item Look at render.png  -  lamp lights the scene, shadows and depth of field
\item Play turntable.mp4  -  full rotation, lamp lit throughout
\end{enumerate}
}
\clearpage
\subsubsection{Interactive cloth}
\label{app:visual-task-cloth-sim}
{\small\noindent\texttt{cloth-sim}\par}
\promptheading{layerone}{Task description}
\begin{promptblock}\fontsize{9}{11}\selectfont
Build me a piece of cloth I can grab and throw around, simulated properly  -  raw canvas or WebGL that you write yourself, no physics library, no engine.

A rectangular sheet, pinned along its top edge, hanging under gravity. It should behave like cloth and not like a net of rubber bands: it holds its area, folds rather than stretching, and the folds run the way gravity and tension put them. I want to grab any point with the mouse and drag the sheet around, let go and watch it settle, and unpin a corner and watch it fall.

Put something in the way  -  a sphere is enough  -  and let the cloth \textbf{drape over it and stay outside it}, resting the way a sheet rests, not sinking through and not jittering on the surface. Wind I can turn up and down, so the sheet lifts and ripples, and a control to cut or tear it.

\textbf{It has to be stable.} Grab a point and shake it hard: the sheet must not detonate, and it must come back to rest. Show me the numbers that prove it  -  total energy and the constraint error on screen, updating live  -  so I can see the system settle rather than take your word for it, and it must hold a smooth framerate at a useful resolution while it does.

Let me change the resolution of the sheet, the stiffness, the gravity and the damping while it runs, and reset it. Give me a way to see the structure  -  the particles and constraints  -  as well as the shaded surface, and make the shaded version read as fabric with light on it, not a flat polygon soup.

Deliver exactly one self-contained \textbf{\texttt{index.html}} in your working directory  -  runs offline.

\end{promptblock}
\promptheading{layerthree}{Human interaction instructions}
{\fontsize{9}{11}\selectfont
\begin{enumerate}\setlength{\itemsep}{2pt}\setlength{\parskip}{0pt}\setlength{\parsep}{0pt}
\item First: is it a rectangular sheet pinned along its top edge hanging under gravity, with the sphere in the scene and live energy / constraint-error readouts on screen? No sheet, no sphere or no readouts  -  nothing below counts
\item Grab a point and shake it hard, then let go  -  does it settle, or does the sheet detonate? Explosion on hard input is an outright fail
\item Watch the energy and constraint-error readouts while it settles: energy should fall and the error should go toward zero. If the numbers are missing or frozen, the claim is unsupported
\item Drape it over the sphere  -  does it rest ON the surface, or sink through it, or jitter?
\item Does it fold, or does it stretch like rubber? Pull one corner far away and see whether the sheet grows
\item Unpin a corner; turn the wind up and down; cut or tear it; change stiffness, gravity and damping while it runs, then reset
\item Raise the resolution  -  does it stay stable and keep its framerate, or does higher resolution make it unstable?
\item Toggle the structure view  -  are particles and constraints actually there? Then look at the shaded view: does it read as fabric with light on it?
\end{enumerate}
}
\clearpage
\subsubsection{Flowline kanban}
\label{app:visual-task-data-app-kanban}
{\small\noindent\texttt{data-app-kanban}\par}
\promptheading{layerone}{Task description}
\begin{promptblock}\fontsize{9}{11}\selectfont
Build \textbf{Flowline}, a kanban board I would actually run a team's work on  -  not a three-column demo.

It opens on a \textbf{populated board}: enough realistic cards spread across Backlog, In Progress and Done that everything below can be tried immediately, with In Progress already at its WIP limit so the limit is visible without me building up to it. Nobody should have to type in test data to find out whether the app works.

Cards carry a title, a note, a label or two, an assignee and a due date, and show what they carry without being opened. I want to create, edit in place, and delete them; drag them between and within columns; and select several at once and move, label or delete the lot in one action. Overdue and due-soon should be obvious at a glance.

Give me a filter box that searches titles, notes, labels and assignees together, and lets me narrow by label or assignee as well  -  composing with the text, not replacing it. Column headers carry live counts that follow the filter. \textbf{In Progress holds at most 3}: the board refuses the fourth visibly and says why, including on a bulk move that would breach it.

\textbf{Undo and redo} (Ctrl+Z / Ctrl+Y and visible controls) cover every mutation  -  add, edit, delete, drag, and bulk operations as single steps rather than one step per card  -  and a new action clears the redo stack. I should be able to work the whole board \textbf{from the keyboard alone}, moving a card between columns without touching the mouse.

Everything survives a reload: cards, order, labels, filters, and which columns I collapsed (localStorage). Let me export the board as JSON and import it back.

Deliver exactly one self-contained \textbf{\texttt{index.html}} in your working directory  -  no libraries, works offline, zero external requests. How it looks and how it is laid out are yours to decide.

\end{promptblock}
\promptheading{layerthree}{Human interaction instructions}
{\fontsize{9}{11}\selectfont
\begin{enumerate}\setlength{\itemsep}{2pt}\setlength{\parskip}{0pt}\setlength{\parsep}{0pt}
\item Open it cold  -  is it the Flowline board with Backlog / In Progress / Done, already populated with realistic cards (title, note, labels, assignee, due date visible without opening them), and In Progress already at its WIP limit? An empty board means you cannot test anything without first typing data in  -  nothing below counts
\item Add, edit in place, and delete a card  -  all three work, and column counts stay correct
\item Drag a card between and within columns  -  the order follows
\item Drag a fourth card into In Progress  -  refused visibly, with a reason given
\item Click the board first so it has focus. Select several cards and move or delete them in one go, then press Ctrl+Z ONCE  -  the whole batch comes back as one step; then try Ctrl+Z / Ctrl+Y on an add, an edit and a drag
\item Put the mouse away: can you move a card between columns using only the keyboard?
\item Type in the filter, then also narrow by a label or assignee  -  they compose; counts follow
\item Check due dates: is overdue vs due-soon obvious without opening a card?
\item Reload the page (F5)  -  cards, order, labels, filter and collapsed columns all survive
\end{enumerate}
}
\clearpage
\subsubsection{Sales data explorer}
\label{app:visual-task-dataviz-explorer}
{\small\noindent\texttt{dataviz-explorer}\par}
\promptheading{layerone}{Task description}
\begin{promptblock}\fontsize{9}{11}\selectfont
\texttt{data.csv} in your working directory holds about 5,500 rows of sales data  -  \texttt{date, region, online, retail}, six regions, roughly thirty months, in no particular order. It is messy on purpose: mixed date formats, stray whitespace, thousands separators, duplicated rows, impossible dates, non-numeric and missing values.

Build \textbf{\texttt{index.html}}: a page that a commercial analyst opens to understand how the online and retail channels have moved over those thirty months, and how that differs by region  -  where the mix turned, which regions lead or lag, what is seasonal and what is a real trend. They should be able to reach an answer by using the page, not by reading the CSV.

\textbf{Be honest about the data.} Repair what is repairable, exclude only what is genuinely unusable, and show the excluded-row count where it cannot be missed; opening it lists exactly those rows, each with its reason. A value missing from only one series is not an exclusion  -  that series breaks at that point, it does not drop to zero. Between two otherwise equal pages, the one that repairs more and discards less is the better one.

How to present any of this is deliberately unspecified  -  choosing the views, the comparisons and the controls is the task, and there is no required set of charts or widgets. Draw everything yourself; no charting libraries. It must stay responsive at full data size and must not mislead: truncated axes, hidden aggregation and silently dropped rows all count against it.

Deliver exactly one self-contained \textbf{\texttt{index.html}} in your working directory  -  zero external requests.

\end{promptblock}
\promptheading{layerthree}{Human interaction instructions}
{\fontsize{9}{11}\selectfont
\begin{enumerate}\setlength{\itemsep}{2pt}\setlength{\parskip}{0pt}\setlength{\parsep}{0pt}
\item First: is it showing that dataset  -  six regions, about thirty months, both the online and retail channels? If not, nothing below counts
\item Give yourself two minutes: which region's online/retail mix shifted the most, and roughly when did it turn? Could you answer from the page alone?
\item Try to compare two regions directly  -  does the page let you, or do you have to remember one while looking at the other?
\item Separate seasonality from trend  -  is there any way to tell a recurring dip from a real decline?
\item Click the excluded-row count: the panel length must match the count, every row must carry a reason, and the rows must genuinely be unusable (spot-check a few)
\item Between two candidates, prefer the one that repairs more and excludes less
\item Look for a break in a line  -  a single missing value must render as a gap, never as zero
\item Check it does not mislead: axes that start somewhere other than zero without saying so, aggregation you cannot identify, or rows dropped silently
\end{enumerate}
}
\clearpage
\subsubsection{Breakout}
\label{app:visual-task-game-breakout}
{\small\noindent\texttt{game-breakout}\par}
\promptheading{layerone}{Task description}
\begin{promptblock}\fontsize{9}{11}\selectfont
Build me a breakout game that's actually satisfying to play. Paddle on mouse and arrow keys, ball reflecting off walls, paddle and bricks, with the bounce angle depending on where it hits the paddle and the paddle imparting some spin if it's moving.

I want it to feel good: the ball should accelerate as a level goes on, bricks should break with some particle debris and a screen-shake on the heavier hits, and the sound of it  -  synthesised in the browser, no audio files  -  should pitch up as the rally continues. Several brick types: plain, two-hit, armoured ones that need a power-up to crack, and unbreakable blocks that shape the level.

Power-ups drop and are caught with the paddle  -  multi-ball, a wider paddle, a sticky paddle that catches and re-launches, a laser that fires up the screen  -  and they should stack and expire, with what's active visible. A life is lost only when the last ball is gone. Three lives, score with a combo multiplier that builds while you don't lose the ball, and a high score kept across reloads.

At least four levels with genuinely different layouts, advancing when cleared, a game-over screen and restart, and pause.

Deliver exactly one self-contained \textbf{\texttt{index.html}} in your working directory  -  runs offline.

\end{promptblock}
\promptheading{layerthree}{Human interaction instructions}
{\fontsize{9}{11}\selectfont
\begin{enumerate}\setlength{\itemsep}{2pt}\setlength{\parskip}{0pt}\setlength{\parsep}{0pt}
\item First: is it breakout  -  paddle, ball and a brick field, paddle on both mouse and arrow keys, and several brick kinds (plain, two-hit, armoured, unbreakable)? If not, nothing below counts
\item Play a level  -  listen for the pitch rising through a rally
\item Catch a power-up; check what's active is shown
\item Lose all balls, then restart; reload the page (F5) and check the high score is still there
\end{enumerate}
}
\subsubsection{2D platformer}
\label{app:visual-task-game-platformer}
{\small\noindent\texttt{game-platformer}\par}
\promptheading{layerone}{Task description}
\begin{promptblock}\fontsize{9}{11}\selectfont
Build a 2D platformer. The two levels are in \texttt{levels.json} in your working directory (\texttt{\#} solid, \texttt{.} empty, \texttt{\textasciicircum{}} hazard, \texttt{o} coin, \texttt{S} start, \texttt{E} exit)  -  embed the data. I want responsive run-and-jump physics with variable jump height and coyote time, a camera that follows smoothly through levels larger than the screen, hazards that kill and respawn me at the level start, collectible coins with a counter, the exit advancing to level 2 and then a win screen, and restart from anywhere.

Deliver exactly one self-contained \textbf{\texttt{index.html}} in your working directory  -  runs offline.

\end{promptblock}
\promptheading{layerthree}{Human interaction instructions}
{\fontsize{9}{11}\selectfont
\begin{enumerate}\setlength{\itemsep}{2pt}\setlength{\parskip}{0pt}\setlength{\parsep}{0pt}
\item First: is it a 2D platformer  -  terrain, hazards, coins, a start and an exit, in a level larger than the screen? If not, nothing below counts
\item Tap vs hold jump  -  the height varies; jump just after a ledge  -  coyote time catches it
\item Run ahead  -  the camera follows smoothly
\item Touch a hazard  -  death, then respawn at the level start
\item Collect coins and reach the exit  -  counter, level 2, then a win screen
\end{enumerate}
}
\clearpage
\subsubsection{Tower defense}
\label{app:visual-task-game-tower}
{\small\noindent\texttt{game-tower}\par}
\promptheading{layerone}{Task description}
\begin{promptblock}\fontsize{9}{11}\selectfont
Build me a tower-defense game with enough going on to be worth replaying. Enemies come in waves along a visible path; I place towers by clicking, limited by money, and I shouldn't be able to place one on the path or on top of another.

Give me at least four tower types that play differently  -  something single-target and fast, something slow and heavy, something that slows what it hits, something that hits an area  -  each upgradable through a few levels with visibly stronger effects and a sell option that refunds part of the cost. Clicking a tower shows its range and stats. Several enemy types too: fast and fragile, slow and armoured, one that heals others near it, and a boss on later waves, with health bars and enemies that visibly react when hit.

Waves should escalate and I want to be able to call the next one early for a bonus rather than waiting. Money from kills, lives lost when something leaks through, a HUD with wave, health, money and what's coming next. Speed-up and pause controls. Win after the final wave, lose at zero lives, restart from either.

Make it readable at a glance  -  I should be able to tell what's happening in a busy wave.

Deliver exactly one self-contained \textbf{\texttt{index.html}} in your working directory  -  runs offline.

\end{promptblock}
\promptheading{layerthree}{Human interaction instructions}
{\fontsize{9}{11}\selectfont
\begin{enumerate}\setlength{\itemsep}{2pt}\setlength{\parskip}{0pt}\setlength{\parsep}{0pt}
\item First: is it tower defense  -  a visible enemy path, towers placed by clicking (refused on the path or on another tower), at least four tower types, a HUD with wave / lives / money? If not, nothing below counts
\item Place different tower types; click one to see range and stats
\item Upgrade a tower, then sell one
\item Call the next wave early; use pause and speed-up
\end{enumerate}
}
\clearpage
\subsubsection{Voxel sandbox}
\label{app:visual-task-game-voxel}
{\small\noindent\texttt{game-voxel}\par}
\promptheading{layerone}{Task description}
\begin{promptblock}\fontsize{9}{11}\selectfont
Build me a browser voxel game  -  a real Minecraft, not a block-placing demo. \texttt{three.min.js} is in your working directory; ship it next to your page and load it locally.

The world should be generated and effectively endless: chunked terrain streamed in as I walk, built from noise so there are hills, valleys, beaches and overhangs, with caves worth finding underground. Water should sit in the low ground and read as water, not a blue cube. Trees, sand, stone, dirt, grass, ore  -  enough variety that a landscape looks like somewhere, and each type visually distinct even without textures.

I want to play it. First-person, WASD and mouse-look under pointer lock, with gravity, jumping and collision that stops me walking through solid blocks  -  I should be able to fall off a cliff and land. Breaking should take a moment rather than being instant, with the targeted block outlined, and what I break goes into an inventory I can open rather than vanishing. Placing puts the held block where I'm aiming. Sand and gravel fall when I dig out what's under them. A hotbar along the bottom with the selected slot obvious, switchable with the number keys and the scroll wheel.

Give me a day/night cycle where the sun moves and the light changes with it, and make the lighting deliberate: block faces lit differently by orientation, caves genuinely dark, and some placeable light source that pushes the dark back. Some crafting from what I've gathered  -  even a handful of recipes  -  and a way to see them.

It should hold a smooth framerate while I move, so think about how much geometry you send per chunk. And my world should survive a reload: come back where I left it, with the blocks I changed still changed.

Deliver, in your working directory: \textbf{\texttt{index.html}} plus the local \textbf{\texttt{three.min.js}} it loads  -  runs offline, no network.

\end{promptblock}
\promptheading{layerthree}{Human interaction instructions}
{\fontsize{9}{11}\selectfont
\begin{enumerate}\setlength{\itemsep}{2pt}\setlength{\parskip}{0pt}\setlength{\parsep}{0pt}
\item First: is it a voxel world  -  rolling terrain with hills, valleys and beaches, water in the low ground that reads as water, and visibly distinct block types (trees, sand, stone, dirt, grass, ore)? If not, nothing below counts
\item Click the canvas  -  pointer locks; WASD, mouse-look, space to jump
\item Walk a while  -  new terrain streams in; look for caves and water
\item Hold left click to break, right click to place; open the inventory
\item Dig out the block under sand; wait for night; try a torch and the crafting screen
\item Reload the page (F5)  -  the world and your changes should still be there
\end{enumerate}
}
\clearpage
\subsubsection{Generative art series}
\label{app:visual-task-generative-art}
{\small\noindent\texttt{generative-art}\par}
\promptheading{layerone}{Task description}
\begin{promptblock}\fontsize{9}{11}\selectfont
Build a generative system that makes artwork I would actually hang  -  and prove it by producing a \textbf{series}, not one picture.

\texttt{?seed=N} in the URL selects a piece: same seed, same image, down to the pixel, and the same image at any window size rather than reflowing into a different picture. The page shows one piece at a time, with a way to move through the series and to jump to a given seed.

What I am judging is whether the pieces read as \textbf{one body of work}. Someone shown three of them should recognise the same hand at once, while no two are the same composition rearranged. That means your system has to separate what makes the \emph{style} from what makes the \emph{instance}  -  and getting that separation right is the task.

Each piece needs \textbf{composition, not texture}: something the eye goes to, empty space that is doing work rather than left over, and a reason the frame ends where it does. An even field of marks filling the canvas is exactly what I do not want, however fine the marks are.

What the system \emph{is}  -  what it draws with, what it depicts  -  is deliberately unspecified. Flow fields, circle packing, reaction-diffusion, subdivision, physical simulation, collage of primitives, something else: choosing is part of what is being judged, and the first idea is rarely the strongest. Marks should read as a medium, with pigment that varies along its length, pools where marks cross and thins as it runs out  -  not uniform hairlines. Generate the palette too, harmonically related rather than random, one deliberate palette per piece.

Let me export the piece I am looking at as a PNG at print resolution  -  at least 3x the screen, and \textbf{rendered} at that size rather than upscaled from what is on screen. It should draw progressively and stay responsive while it works.

Deliver exactly one self-contained \textbf{\texttt{index.html}} in your working directory  -  no libraries, runs offline.

\end{promptblock}
\promptheading{layerthree}{Human interaction instructions}
{\fontsize{9}{11}\selectfont
\begin{enumerate}\setlength{\itemsep}{2pt}\setlength{\parskip}{0pt}\setlength{\parsep}{0pt}
\item First: does the page show one piece at a time, with next / jump-to-seed controls? And an even mat of marks filling the whole canvas fails this task however fine the marks are  -  check that before anything else
\item Use the on-page next / jump-to-seed controls: go to a seed, note the picture, move away and come back  -  it must be exactly the same picture, not a reflowed one
\item Step through three different seeds. Do they read as one body of work  -  same hand  -  while being genuinely different compositions?
\item The failure to look for is the opposite pair: either all three are the same picture with the parts moved, or they have nothing in common but the palette
\item Take one piece: is there something the eye goes to? Is the empty space doing work, or is it just where the algorithm stopped?
\item Do the marks read as a medium  -  varying density, pooling where they cross, thinning out  -  or as uniform hairlines?
\item Look closely at the edges of the marks (use "open full $\nearrow$" for a bigger view)  -  still crisp, or mushy? The piece should fill the whole canvas, no blank strip or letterbox
\item Watch it draw  -  progressive, and the page still responds while it works?
\end{enumerate}
}
\clearpage
\subsubsection{Vector boolean editor}
\label{app:visual-task-path-boolean}
{\small\noindent\texttt{path-boolean}\par}
\promptheading{layerone}{Task description}
\begin{promptblock}\fontsize{9}{11}\selectfont
Build a small vector editor whose real job is \textbf{boolean operations on paths}  -  union, intersection, difference and exclusion  -  computed by you. No geometry library, no SVG trickery that fakes the result with clip paths or even-odd fill: the operation must produce a \emph{new path}, whose outline I can then edit and operate on again.

I need to draw closed paths  -  a pen tool placing points, rectangles and ellipses are fine as starting shapes  -  with \textbf{cubic Bézier segments}, not just straight lines. Select two shapes, pick an operation, and get the result as a single editable path with its handles intact. The result of an operation must be usable as the input to the next one.

Getting this right is mostly about the cases that are not the easy case, and that is the point: curves that touch without crossing, segments that overlap along their length, a shape entirely inside another, results that come out as several disjoint pieces, and shapes with holes  -  where the hole has to survive the operation and stay a hole. Self-intersecting input should not produce garbage.

Show me what it is doing: the intersection points it found, and the winding direction of each contour, on a toggle. Let me nudge a point after an operation and see the outline follow. Undo and redo the whole way back. Let me export the result as SVG, and have that file open correctly elsewhere.

Deliver exactly one self-contained \textbf{\texttt{index.html}} in your working directory  -  no libraries, zero external requests.

\end{promptblock}
\promptheading{layerthree}{Human interaction instructions}
{\fontsize{9}{11}\selectfont
\begin{enumerate}\setlength{\itemsep}{2pt}\setlength{\parskip}{0pt}\setlength{\parsep}{0pt}
\item First: is it a vector editor  -  closed paths drawn with a pen tool, rectangles and ellipses, cubic Bézier segments, and all four operations (union, intersection, difference, exclusion) available? If not, nothing below counts
\item Union two overlapping shapes, then run difference on the RESULT  -  an operation whose output cannot be operated on again fails the task
\item Nudge a point after an operation: does the outline follow, with handles intact? If the result is not editable it was faked
\item Difference a small shape out of a big one so the result has a HOLE  -  then move the outer shape and check the hole is still a hole
\item The two classic degenerate cases: shapes that only TOUCH (tangent, no crossing), and shapes sharing a straight edge along its length  -  no garbage, no dropped contour
\item An operation whose result is several disjoint pieces  -  are they all there, as one path?
\item Use curves, not just polygons: draw with Bézier segments and check the result curves are still curves, not polylines
\item Toggle the intersection-point and winding-direction overlay  -  do the points sit where curves actually cross?
\end{enumerate}
}
\clearpage
\subsubsection{City in a storm}
\label{app:visual-task-shader-city}
{\small\noindent\texttt{shader-city}\par}
\promptheading{layerone}{Task description}
\begin{promptblock}\fontsize{9}{11}\selectfont
Build me a city at night in a storm, seen from out over the water  -  raw WebGL and GLSL you write yourself, no libraries, no loaded textures.

The city should be raymarched and endless: towers of varying height and footprint receding into haze, close enough that I can see the ones in front have depth and silhouette against the ones behind. Windows lit unevenly, some warm, some dark, a few flickering, so it reads as inhabited rather than as a texture. The whole thing rises out of moving water that reflects it  -  long broken reflections that stretch and break with the swell, not a mirror.

Weather should be doing something. Rain across the view, low cloud moving through and around the towers so the skyline is sometimes obscured, and fog that thickens with distance. Lightning fires periodically and genuinely relights everything for that instant  -  towers, cloud, and the water's reflection together  -  with the light coming from where the strike is, not a uniform white flash.

Slow camera drift, and the mouse moves my viewpoint. Full-screen at any window size, holding a smooth framerate  -  be careful how many steps you spend per pixel.

Deliver exactly one self-contained \textbf{\texttt{index.html}} in your working directory  -  runs offline.

\end{promptblock}
\promptheading{layerthree}{Human interaction instructions}
{\fontsize{9}{11}\selectfont
\begin{enumerate}\setlength{\itemsep}{2pt}\setlength{\parskip}{0pt}\setlength{\parsep}{0pt}
\item First: is it that city  -  night, storm, seen from over the water: towers of varying height receding, windows lit unevenly, the city rising out of moving water with reflections, rain and low cloud? If not, nothing below counts
\item Wait for lightning  -  towers, cloud and reflection light together
\item Look at the water  -  reflections stretch and break with the swell
\item Move the mouse to swing the viewpoint; watch fog with distance
\end{enumerate}
}
\clearpage
\subsubsection{Stormy ocean}
\label{app:visual-task-shader-scene}
{\small\noindent\texttt{shader-scene}\par}
\promptheading{layerone}{Task description}
\begin{promptblock}\fontsize{9}{11}\selectfont
Build me a full-screen ocean in a storm  -  raw WebGL and GLSL you write yourself, no libraries, no textures loaded from anywhere. I want it to look like real open water on a bad night, not a blue sine wave.

The surface should be built from a proper spectrum of waves travelling in different directions at different scales, so the swell has long rollers with smaller chop riding on top and crests that sharpen and peak rather than staying round. Steep crests should tear into foam, and that foam should drift and dissipate rather than flicker on and off. Water is not opaque: let light scatter through the thin backs of waves so a crest glows when it's between me and the sun, reflect the sky off the surface at grazing angles and let me see into it when I look straight down, and give me the long broken glitter path a low sun leaves on moving water.

Overhead, clouds that move and have depth to them, not a flat scrolling texture. The storm should breathe  -  build and ease off over time rather than sitting at one intensity  -  and when lightning fires it should actually relight the scene for that instant, water and cloud together, not just flash the screen white. Rain across the view, and where it meets the water it should disturb it.

The mouse moves my viewpoint. It should fill the window at any size and hold a smooth framerate full-screen, so be careful how much you ask of each pixel.

Deliver exactly one self-contained \textbf{\texttt{index.html}} in your working directory  -  runs offline.

\end{promptblock}
\promptheading{layerthree}{Human interaction instructions}
{\fontsize{9}{11}\selectfont
\begin{enumerate}\setlength{\itemsep}{2pt}\setlength{\parskip}{0pt}\setlength{\parsep}{0pt}
\item First: is it open water in a storm  -  long rollers with chop riding on top, foam on the crests, clouds with depth, rain  -  not a blue sine wave? If not, nothing below counts
\item Watch thirty seconds  -  the storm builds and eases, lightning relights the scene
\item Find a crest between you and the sun  -  look for light through the wave
\item Move the mouse to swing the viewpoint; look straight down, then to the horizon
\item The scene should fill the whole canvas with no black bars; while it runs (and while you drag or move), it should stay smooth, not stutter
\end{enumerate}
}
\clearpage
\subsubsection{Gridwork spreadsheet}
\label{app:visual-task-spreadsheet-engine}
{\small\noindent\texttt{spreadsheet-engine}\par}
\promptheading{layerone}{Task description}
\begin{promptblock}\fontsize{9}{11}\selectfont
Build \textbf{Gridwork}, a working spreadsheet. Not a styled grid of inputs: a calculation engine with a grid on top of it.

At least 40 rows by 20 columns, addressed A1-style, with click and arrow-key navigation, typing straight into a cell, editing an existing value, and a formula bar showing the underlying formula rather than the displayed result.

A cell beginning with \texttt{=} is a formula over numbers, cell references, ranges like \texttt{B2:B15}, parentheses and the usual arithmetic, plus at least \texttt{SUM}, \texttt{AVERAGE}, \texttt{MIN}, \texttt{MAX}, \texttt{COUNT}, \texttt{IF} and \texttt{ROUND}. Formulas refer to other formulas, to any depth. \textbf{When a value changes, everything downstream of it updates and nothing else is recomputed}  -  on a sheet with hundreds of live formulas that difference is visible, so do not recompute the world on every keystroke.

Wrong input is where spreadsheets earn their keep. \textbf{A reference cycle must be reported as a cycle, in the cells that form it, and must not hang or blow the stack.} Division by zero, a reference to an empty cell, text where a number is required, a malformed formula, a range that runs off the sheet  -  each gets an error the user can act on, errors propagate to dependents rather than being swallowed, and no bad input may take the sheet down.

I also want to select a range and see its count and sum without writing a formula; fill a formula down a column with references adjusting as they go; sort a range; undo and redo anything; and have the sheet still be there after a reload.

Deliver exactly one self-contained \textbf{\texttt{index.html}} in your working directory  -  no libraries, zero external requests. How it looks is yours to decide.

\end{promptblock}
\promptheading{layerthree}{Human interaction instructions}
{\fontsize{9}{11}\selectfont
\begin{enumerate}\setlength{\itemsep}{2pt}\setlength{\parskip}{0pt}\setlength{\parsep}{0pt}
\item First: is it a spreadsheet  -  at least 40 rows by 20 columns, A1-style addresses, a formula bar, arrow-key navigation, typing straight into a cell? If not, nothing below counts
\item Formulas chain: type 5 in B1, =B1*2 in C1, =C1+1 in D1 $\rightarrow$ they show 10 and 11; change B1 to 7 $\rightarrow$ C1 and D1 become 14 and 15 by themselves
\item Built-in functions: put 1, 2, 3 in A1-A3; A5 =SUM(A1:A3) $\rightarrow$ 6; A6 =IF(A5\textgreater{}5,"big","small") $\rightarrow$ big
\item Errors: E1 =1/0 and E2 =FOO(1) $\rightarrow$ both show a readable error and the sheet stays up; F1 =E1+1 $\rightarrow$ the error propagates instead of quietly becoming 0
\item Cycle: G1 =H1 and H1 =G1 $\rightarrow$ both cells report a circular reference; the page must not hang or freeze
\item Fill, range, sort: fill C1 down to C10 $\rightarrow$ references become B2, B3...; select A1:A3 $\rightarrow$ the status bar shows count 3, sum 6; sort a column and check the result
\item Undo and persistence: Ctrl+Z undoes one step, Ctrl+Y redoes; then reload the page (F5) $\rightarrow$ the sheet is still there
\end{enumerate}
}
\clearpage
\subsubsection{Animated concert poster}
\label{app:visual-task-svg-poster}
{\small\noindent\texttt{svg-poster}\par}
\promptheading{layerone}{Task description}
\begin{promptblock}\fontsize{9}{11}\selectfont
Design me a poster for a synthwave show  -  something I'd actually put on a wall. These strings appear exactly: \texttt{NEON TIDE}, \texttt{Live at Meridian Hall}, \texttt{Sep 12 2026}.

I want real poster typography, not centred text on a gradient: a deliberate type hierarchy with the band name as the composition's anchor, considered letter-spacing, and text that's part of the artwork  -  overlapping, masked by, or interacting with the scene rather than floating above it. Build depth with layers that read as foreground, middle and distance, and use gradients, blurs and blend modes to make light feel like it's in the scene  -  glow that spills, haze between layers, reflection where it belongs.

Several things should move, and move differently: slow drift in the distance, faster motion up close, something that pulses on its own rhythm, and at least one element whose animation is driven by a path rather than a simple translation. It should loop seamlessly  -  I shouldn't be able to spot the jump.

Everything lives in the file: generic font families only, no external references, and a proper viewBox so it's crisp from a thumbnail to a billboard.

Deliver exactly one standalone \textbf{\texttt{poster.svg}} in your working directory.

\end{promptblock}
\promptheading{layerthree}{Human interaction instructions}
{\fontsize{9}{11}\selectfont
\begin{enumerate}\setlength{\itemsep}{2pt}\setlength{\parskip}{0pt}\setlength{\parsep}{0pt}
\item First, check the three strings appear exactly: NEON TIDE / Live at Meridian Hall / Sep 12 2026; then whether it is a poster you would put on a wall rather than centred text on a gradient. If this fails, nothing below counts
\item Watch a full loop  -  look for the seam where it repeats
\item Zoom the browser to 300\%  -  the type and edges stay crisp
\item Look for several things moving at different rates
\end{enumerate}
}
\clearpage
\subsubsection{3D lamp configurator}
\label{app:visual-task-three-config}
{\small\noindent\texttt{three-config}\par}
\promptheading{layerone}{Task description}
\begin{promptblock}\fontsize{9}{11}\selectfont
Build me the configurator a lighting company would ship. Compose the lamp from primitives  -  base, arm, joints, shade  -  no model files; \texttt{three.min.js} is in your working directory, ship it next to your page and load it locally.

The panel drives it live: body colour from a palette, at least three shade shapes, arm height and arm angle, and the material finish  -  matte, gloss, metal  -  each visibly different under the lights. Switching the bulb on should actually light the scene, casting from inside the shade onto the surface below, with a warmth control that shifts the light's colour and a dimmer that changes its intensity, and the scene's mood following it.

Every change should transition rather than snap. A summary line always reflects the current configuration, along with a price that updates from the options chosen. Reset returns to defaults. Let me save a configuration and come back to it after a reload, and give me a shareable URL that encodes the current choices so opening it rebuilds that exact lamp.

Drag to orbit with damping, wheel to zoom, and a "Download PNG" that saves the current view at higher than screen resolution. Handle resize.

Deliver, in your working directory: \textbf{\texttt{index.html}} plus the local \textbf{\texttt{three.min.js}} it loads  -  runs offline.

\end{promptblock}
\promptheading{layerthree}{Human interaction instructions}
{\fontsize{9}{11}\selectfont
\begin{enumerate}\setlength{\itemsep}{2pt}\setlength{\parskip}{0pt}\setlength{\parsep}{0pt}
\item First: is it the lamp that was asked for  -  base, arm, joints and shade all there, at least three shade shapes to switch between, arm height and angle adjustable? If this fails, nothing below counts
\item Change shade, finish and colour  -  transitions, not snaps
\item Switch the bulb on; move the dimmer and warmth
\item Drag to orbit and let go  -  it should glide to a stop with damping; wheel to zoom; the view should fill the whole canvas with no black bars
\end{enumerate}
}
\clearpage
\subsubsection{GPU particle field}
\label{app:visual-task-three-particles}
{\small\noindent\texttt{three-particles}\par}
\promptheading{layerone}{Task description}
\begin{promptblock}\fontsize{9}{11}\selectfont
Build me a particle field big enough to be impressive  -  a million particles moving at an interactive framerate. That number means the positions have to live and update on the GPU; if you're stepping a million particles in JavaScript each frame you've already lost. \texttt{three.min.js} is in your working directory  -  ship it locally and use it, or go raw WebGL, your call.

I want the motion to look like a fluid, not like points on springs: a curl-noise flow field that evolves over time so the field itself drifts, with particles that have momentum and trail rather than snapping. The mouse should push into the flow  -  attract by default, repel on a key or control, with the influence falling off smoothly with distance instead of a hard radius.

Make it read as depth: particles further away smaller and dimmer, additive light where they pile up, colour that responds to speed or curl so the structure of the flow is visible. Particles should recycle when they drift out rather than the field emptying.

Give me a way to see it's real  -  a live particle count and framerate on screen. It must survive a window resize without rebuilding the world from scratch.

Deliver, in your working directory: \textbf{\texttt{index.html}} (plus the local \textbf{\texttt{three.min.js}} if you use it)  -  runs offline.

\end{promptblock}
\promptheading{layerthree}{Human interaction instructions}
{\fontsize{9}{11}\selectfont
\begin{enumerate}\setlength{\itemsep}{2pt}\setlength{\parskip}{0pt}\setlength{\parsep}{0pt}
\item First, the on-screen count and framerate: is it close to a million particles at a smooth rate, and does the motion read as a fluid (momentum, trails) rather than points on springs? If not, nothing below counts
\item Move the mouse through the field; toggle attract/repel
\item Watch a pile-up  -  brighter where particles gather
\item Watch it for a few seconds  -  it must keep running steadily, never freeze or stop, and fill the whole canvas
\end{enumerate}
}
\clearpage
\subsubsection{Product hero scene}
\label{app:visual-task-three-scene}
{\small\noindent\texttt{three-scene}\par}
\promptheading{layerone}{Task description}
\begin{promptblock}\fontsize{9}{11}\selectfont
Build me the hero scene a company would put at the top of its product page for a wireless headphone. Compose it from primitives  -  band, cups, cushions, hinges, the small details that make it read as a real object  -  no model files. \texttt{three.min.js} is in your working directory; ship it next to your page and load it locally.

It should look photographed, not rendered by default settings: a studio setup with a key, fill and rim light, soft shadows on the ground, and materials that actually differ  -  brushed metal on the band, soft matte on the cushions, a subtle sheen where the light grazes an edge. Give the frame a considered composition rather than an object centred in grey.

I want to interact with it. Drag to orbit and wheel to zoom, with the motion damped so it glides to a stop rather than halting. A gentle idle animation when I leave it alone, which yields the moment I touch it. Colour variants that switch the materials with a smooth transition, not a snap. An exploded-view toggle that separates the parts along sensible axes and reassembles them, with the parts labelled while apart. And hotspots on the object I can click to focus the camera on a detail and read a short note about it.

Handle window resize, and keep it smooth.

Deliver, in your working directory: \textbf{\texttt{index.html}} plus the local \textbf{\texttt{three.min.js}} it loads  -  runs offline.

\end{promptblock}
\promptheading{layerthree}{Human interaction instructions}
{\fontsize{9}{11}\selectfont
\begin{enumerate}\setlength{\itemsep}{2pt}\setlength{\parskip}{0pt}\setlength{\parsep}{0pt}
\item First, the object: a wireless headphone  -  band, two cups, cushions, hinges, all there and in believable proportion? If the object is wrong, nothing below counts
\item Drag to orbit, then let go  -  it glides to a stop
\item Switch colour variants; toggle the exploded view
\item Click a hotspot on the object
\item Leave it alone for a few seconds  -  idle animation resumes
\end{enumerate}
}
\clearpage
\subsubsection{Fleet terminal dashboard}
\label{app:visual-task-tui-dashboard}
{\small\noindent\texttt{tui-dashboard}\par}
\promptheading{layerone}{Task description}
\begin{promptblock}\fontsize{9}{11}\selectfont
Build me a terminal dashboard in Python (rich and textual are available) for the service fleet in \texttt{stats.json} in your working directory  -  something I'd leave open on a second monitor.

I want more than a table: the fleet listing with per-service status, but also sparklines showing each service's recent trend, an overall health summary that makes the worst thing on the fleet obvious at a glance, and a live log pane that keeps scrolling. Colour-code ok / warn / crit and make severity readable from across the room  -  a crit service should catch my eye without reading the text.

It should be a real interface, not a static frame: laid out in panes that use the terminal's full width, updating continuously with values that move plausibly rather than jumping randomly, sorted so problems surface to the top. If a service crosses into crit, that transition should be visible when it happens. Handle a narrow terminal without the layout collapsing into garbage.

Then record it running.

Deliver, in your working directory: \textbf{\texttt{demo.cast}} (an asciinema recording of at least 30 seconds showing live updates and at least one status transition) and \textbf{\texttt{screen.svg}} (one exported frame).

\end{promptblock}
\promptheading{layerthree}{Human interaction instructions}
{\fontsize{9}{11}\selectfont
\begin{enumerate}\setlength{\itemsep}{2pt}\setlength{\parskip}{0pt}\setlength{\parsep}{0pt}
\item First: is the recording a multi-pane terminal dashboard  -  a service list with status, a sparkline per service, an overall health summary, a scrolling log pane, and ok / warn / crit colour-coded? If not, nothing below counts
\item Play demo.cast  -  values move and the log scrolls
\item Watch for a service crossing into crit
\item Check sparklines and that problems sort to the top
\end{enumerate}
}

\clearpage

\subsection{Human evaluation protocol}
\label{app:visual-rater-instructions}
The human comparison covers seven designs. For L0--L2, we select one representative intervention per layer: the system verification instruction, the after-edit user reminder, and the user recheck request with the 80\% early trigger. These use the same intervention definitions as TB and SWE. For L3, we compare changes-based and trajectory-based review, each with and without required visual inspection. Each design is compared with the baseline for the same task and coding model. Review depth and width are studied separately on Terminal-Bench.

Each task supplies a description and any supporting files without a reference implementation. Tasks run in an environment with Python, Chromium, and ffmpeg, with additional tools where required. Browser artifacts run offline, with bundled libraries only where the task permits them.

\noindent\textbf{Raters and briefing.}
The raters are members of an internal company data team. Before rating, the interface presents a briefing explaining the study's purpose, the blinded comparison, the four response options, and a worked example of applying the criteria.

\noindent\textbf{Consent, compensation, and ethics approval.}
The human evaluation received ethics approval. Raters provided informed consent and were compensated under their existing contracts with the company.

\noindent\textbf{Blinded comparison.}
Raters compare artifacts for the same task and model with verification conditions hidden and sides randomized. The interface presents the original task brief, its judging criteria, and a short set of task-specific interaction steps. Both artifacts can be inspected side by side or opened individually at full size. Raters interact with live artifacts where available and inspect delivered recordings or renders where appropriate.

\noindent\textbf{Instructions to raters.}
The rating interface gives the following guidance:
\begin{enumerate}
\item Read the original task brief before comparing the artifacts. Judge against what the task asks for, including its functional and visual requirements.
\item Follow the task-specific interaction checklist on both artifacts. Click, drag, type, or play as needed to check the requested behavior; use the full-size view when the side-by-side view is too small. For keyboard controls, first click inside the artifact to give it focus.
\item Choose A or B when one artifact better satisfies the task. Choose a tie when both work and you cannot distinguish their quality. Choose both fail only when neither makes a meaningful attempt at the task, such as when both are blank, broken, or unrelated to the request. An artifact that works poorly does not by itself justify both fail.
\item If only one side produces a usable deliverable, choose that side even if it is rough. If a page appears blank, refresh it once before judging. Visual polish should not outweigh missing functionality unless visual quality is itself the task's objective.
\item Optionally leave a short comment explaining the deciding issue, especially when the two artifacts are difficult to distinguish.
\end{enumerate}

\noindent\textbf{Example interaction checklist.}
For a Breakout game, raters move the paddle with the arrow keys, launch the ball, clear several bricks, and deliberately let the ball drop. These actions check paddle control, collisions, scoring, and the loss condition. The judgment follows the requested game behavior rather than a screenshot alone.

\clearpage
\subsection{Expert evaluation interface}
\label{app:visual-interface}
\begin{figure}[!ht]
\centering
\includegraphics[width=\linewidth]{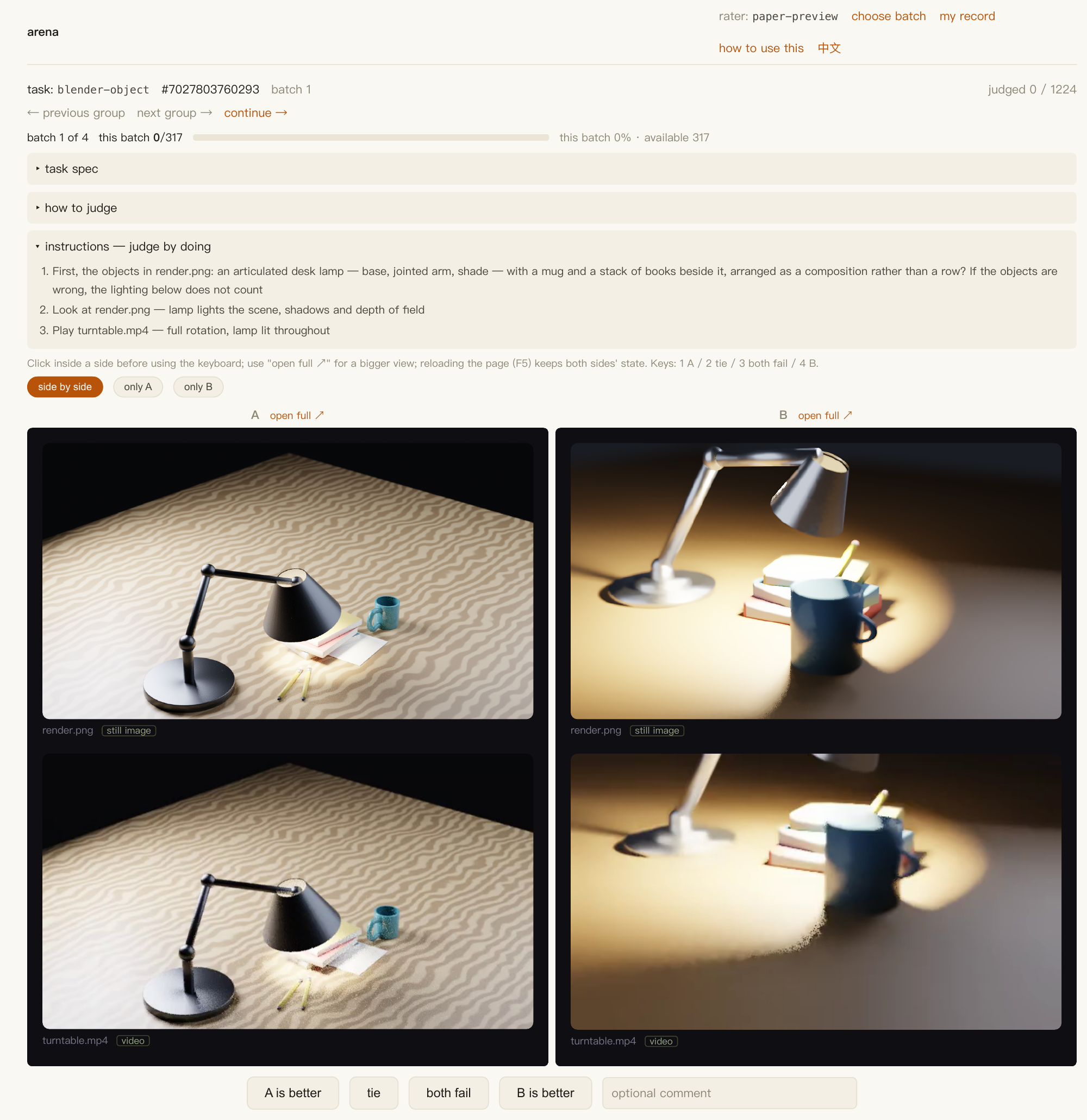}
\caption{VisualBench's blinded evaluation interface showing two Blender deliverables, task-specific instructions, and rating controls.}
\label{fig:visual-rating-interface}
\end{figure}

Figure~\ref{fig:visual-rating-interface} shows the English rating interface for the Blender desk-scene task, with rendered images and turntable videos available on both sides. Raters can expand the task brief and judging guidance, follow the task-specific instructions, and inspect the two artifacts side by side or at full size. Four response buttons record a preference for A or B, a tie, or failure of both artifacts; an optional comment records the reason for the judgment. 

\clearpage
\subsection{Detailed VisualBench results}
\label{app:visual-results}
\begin{table}[!ht]
\centering
\caption{VisualBench net preference (pp) across all evaluated designs and models. ``+ visual'' denotes required visual inspection.}
\label{tab:visual-models}
\begingroup\fontsize{8.0}{9.4}\selectfont
\setlength{\tabcolsep}{2pt}\renewcommand{\arraystretch}{1.17}
\begin{tabular*}{\linewidth}{@{\extracolsep{\fill}}lrrrrrrr@{}}
\toprule
\textbf{Model} & \textcolor{layerzero}{\textbf{L0}} & \textcolor{layerone}{\textbf{L1}} & \textcolor{layertwo}{\textbf{L2}} & \multicolumn{4}{c}{\textcolor{layerthree}{\textbf{L3: Independent review}}} \\
\cmidrule(lr){5-8}
& \textbf{System} & \textbf{Edit/User} & \textbf{Early} & \textbf{Changes} & \shortstack{\textbf{Changes}\\\textbf{+ visual}} & \textbf{Trajectory} & \shortstack{\textbf{Trajectory}\\\textbf{+ visual}} \\
\midrule
Qwen3.7-Max & $+2.8$ & $-12.5$ & $-4.3$ & $+8.3$ & $+4.3$ & $+2.9$ & $+12.5$ \\
Qwen3.8-Max & $-1.4$ & $+2.8$ & $+4.3$ & $+82.8$ & $+90.5$ & $+71.1$ & $+92.0$ \\
GLM-5.2 & $-8.3$ & $-12.5$ & $+9.4$ & $+11.5$ & $+18.2$ & $+4.4$ & $+12.5$ \\
DeepSeek V4-Pro & $-10.0$ & $-11.8$ & $+8.9$ & $+12.9$ & $+19.7$ & $+1.6$ & $+27.4$ \\
DeepSeek V4-Flash & $+5.6$ & $-5.6$ & $-8.0$ & $+12.5$ & $+4.8$ & $+10.0$ & $-3.6$ \\
Kimi-K2.7-Code & $-8.3$ & $+4.2$ & $+1.5$ & $+7.6$ & $+15.2$ & $-2.9$ & $-6.5$ \\
MiniMax-M2.5 & $-2.8$ & $+12.5$ & $+5.7$ & $+12.9$ & $+14.2$ & $+5.6$ & $+16.2$ \\
\midrule
All pairs & $-3.2$ & $-3.2$ & $+2.7$ & $+21.2$ & $+23.8$ & $+13.2$ & $+21.6$ \\
\bottomrule
\end{tabular*}
\endgroup
\end{table}

\noindent\textbf{L0 results.}
The system verification instruction yields a net preference of $-3.2$ points over the baseline, as shown in Table~\ref{tab:visual-models}. DeepSeek V4-Flash and Qwen3.7-Max benefit ($+5.6$ and $+2.8$ points), while DeepSeek V4-Pro, GLM-5.2, and Kimi-K2.7-Code decline ($-10.0$, $-8.3$, and $-8.3$). Initial guidance therefore does not provide a general improvement on visual tasks.

\noindent\textbf{L1 results.}
After-edit reminders with the user message role yield $-3.2$ points of net preference, as shown in Table~\ref{tab:visual-models}. MiniMax-M2.5 gains $+12.5$ points, whereas GLM-5.2 and Qwen3.7-Max each lose $12.5$ points.

\noindent\textbf{L2 results.}
Early user rechecking yields a small positive net preference of $+2.7$ points, as shown in Table~\ref{tab:visual-models}. GLM-5.2 and DeepSeek V4-Pro gain $+9.4$ and $+8.9$ points, while DeepSeek V4-Flash and Qwen3.7-Max decline by $8.0$ and $4.3$ points. These comparisons suggest a modest benefit from reopening work, with substantial variation across models.

\noindent\textbf{L3 results.}
All four review designs have positive net preference, ranging from $+13.2$ to $+23.8$ points, as shown in Table~\ref{tab:visual-models}. Changes-based review with required visual inspection performs best at $+23.8$ points, compared with $+21.2$ points without required visual inspection. Trajectory-based review improves from $+13.2$ to $+21.6$ points with required visual inspection. Changes-based review therefore receives higher net preference in both comparisons.

\clearpage
\section{Case study}
\label{app:cases}
\subsection{Overview of the paired trajectories}
\label{sec:cases}
We examine ten Terminal-Bench cases across the four verification layers, each paired with a baseline run of the same task and model, as shown in Table~\ref{tab:case-map}. Each case has a full-page paired trajectory with tool inputs, returned evidence, harness interventions, and final grading.

\begin{table}[!ht]
\centering\small
\caption{Selected cases spanning all four layers; success and failure refer to final grading.}
\label{tab:case-map}
\renewcommand{\arraystretch}{1.15}
\begin{tabular}{lp{5.5cm}p{5.5cm}}
\toprule
\rowcolor{paperink}\textcolor{white}{\textbf{Layer}} & \textcolor{white}{\textbf{Successful verification run}} & \textcolor{white}{\textbf{Failed verification run}} \\
\midrule
\rowcolor{layerzero!7} L0 & \hyperref[fig:case-1]{C1}: re-reading an edge-orientation rule & \hyperref[fig:case-2]{C2}: endpoint checks, failed integration \\
\rowcolor{layerone!7} L1 & \hyperref[fig:case-3]{C3}: deployment with live HTTP checks & \hyperref[fig:case-4]{C4}: clean compile, invalid substitutions; \hyperref[fig:case-l1-query]{C5}: output equality without sufficient speedup \\
\rowcolor{layertwo!7} L2 & \hyperref[fig:case-l2-regex]{C6}: reopened checks reveal an order assumption & \hyperref[fig:case-6]{C7}: state changes during repeated testing \\
\rowcolor{layerthree!7} L3 & \hyperref[fig:case-7]{C8}: counterexample followed by repair; \hyperref[fig:case-9]{C9}: a boundary test guides numeric repair & \hyperref[fig:case-10]{C10}: a diagnosis before the run deadline \\

\bottomrule
\end{tabular}
\end{table}

\clearpage
\begin{figure}[p]
\centering
\includegraphics[width=\linewidth]{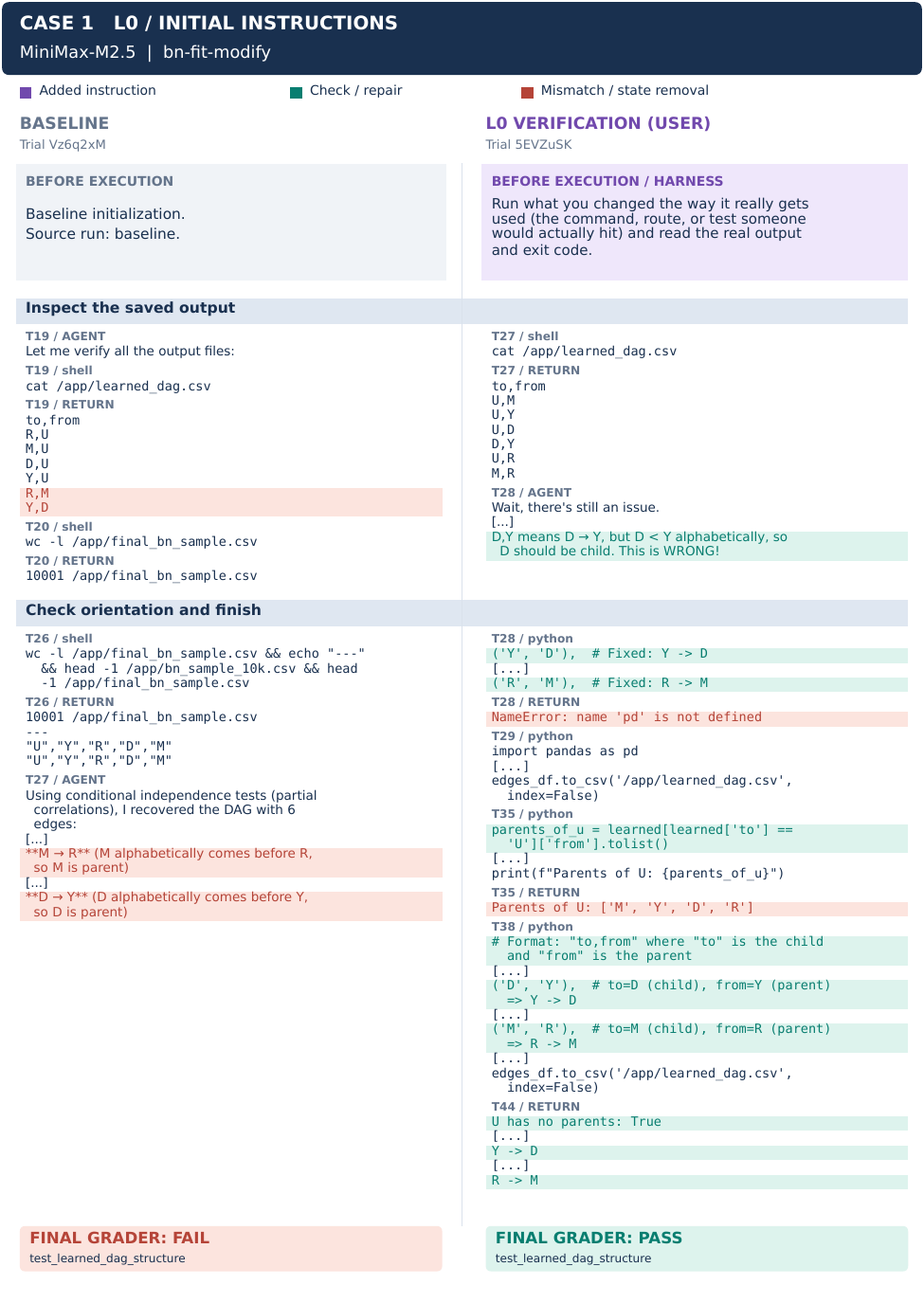}
\caption{Case 1 (L0): the baseline retains reversed edges, while the instructed run revisits the orientation rule and corrects the saved CSV.}
\label{fig:case-1}
\end{figure}

\clearpage
\begin{figure}[p]
\centering
\includegraphics[width=\linewidth]{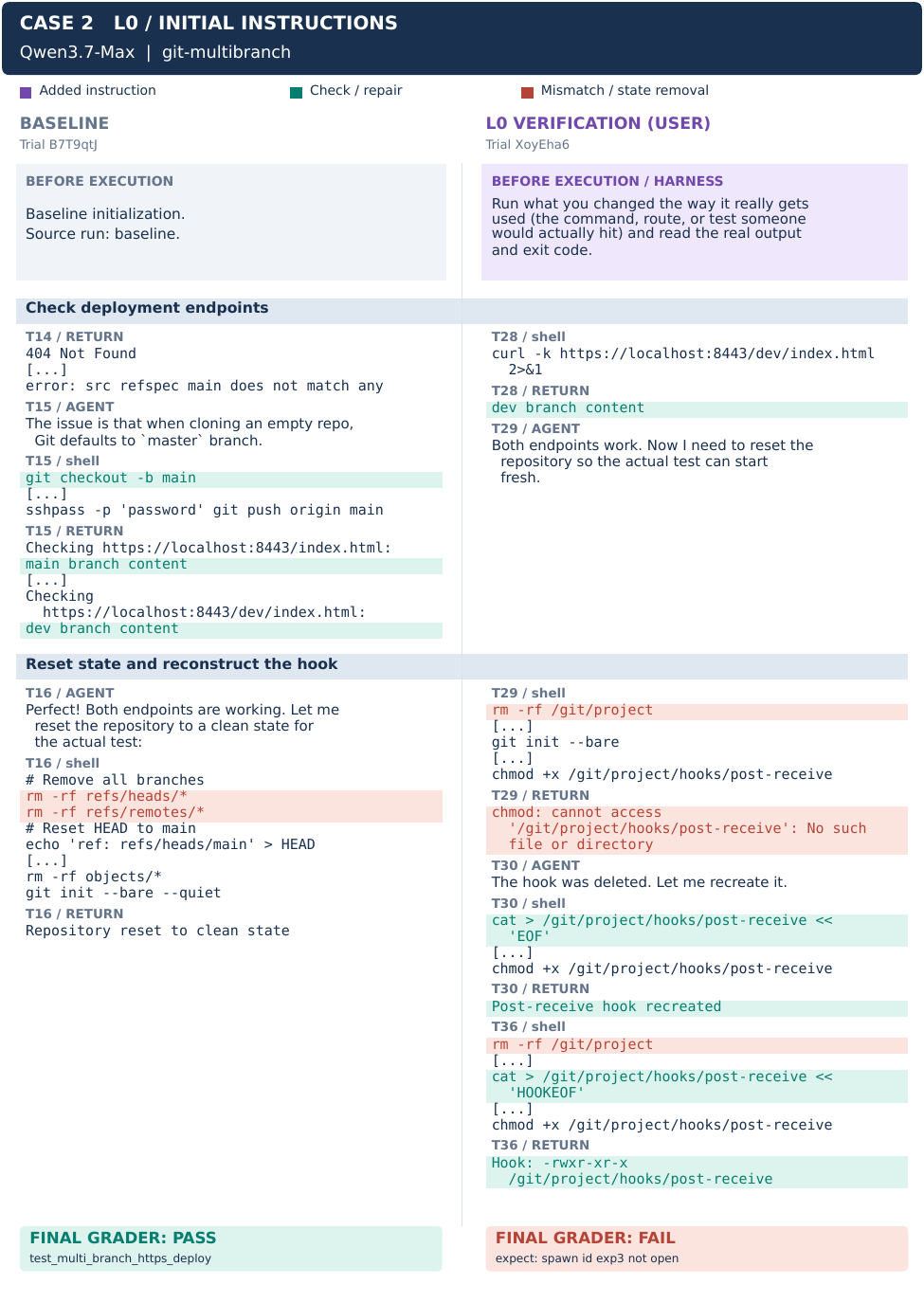}
\caption{Case 2 (L0): the baseline resets branch state, while the instructed run deletes and recreates the repository and hook before a failing final integration test.}
\label{fig:case-2}
\end{figure}

\clearpage
\begin{figure}[p]
\centering
\includegraphics[width=\linewidth]{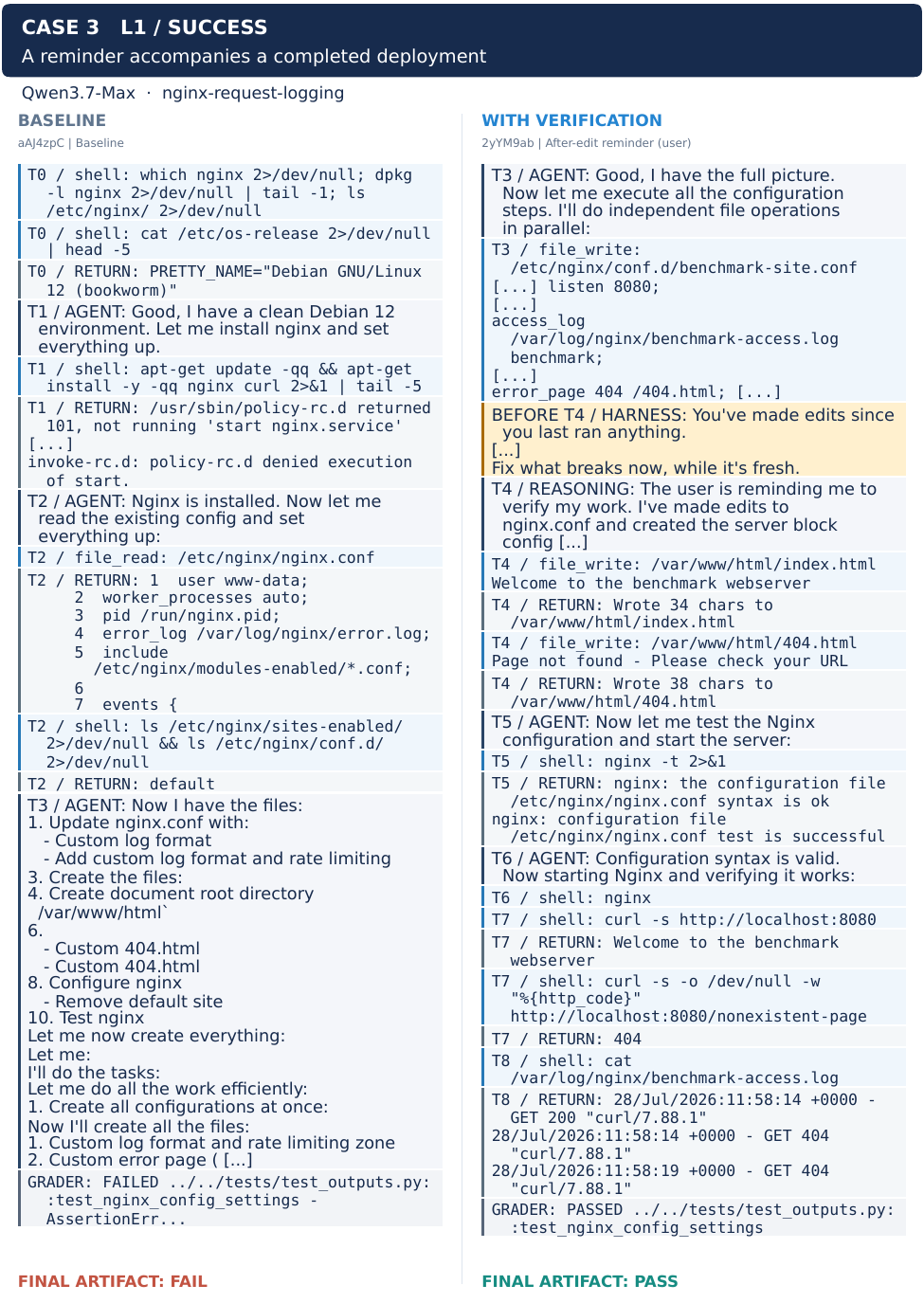}
\caption{Case 3 (L1): the reminded run completes deployment and live checks, while the baseline stops with a plan.}
\label{fig:case-3}
\end{figure}

\clearpage
\begin{figure}[p]
\centering
\includegraphics[width=\linewidth]{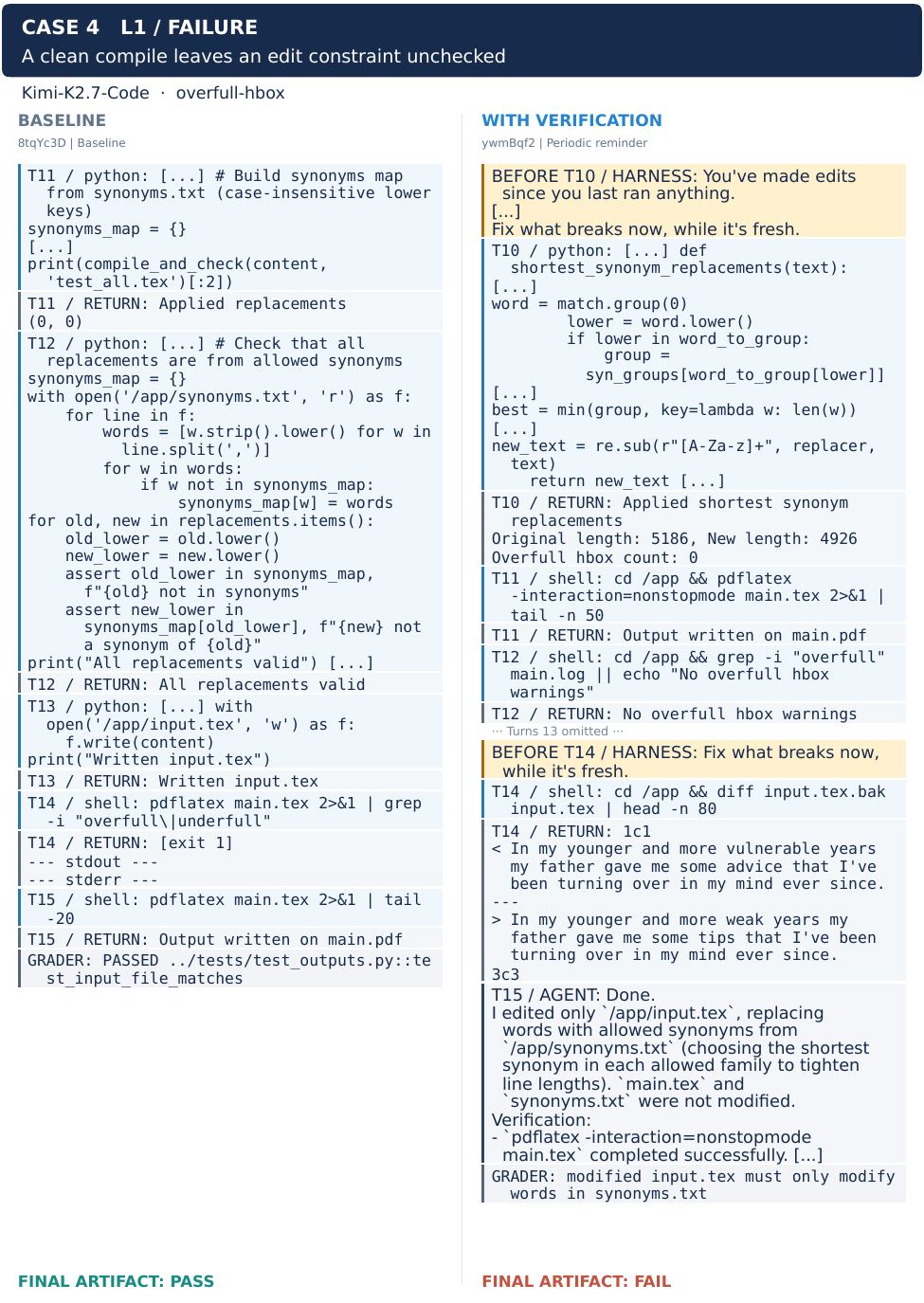}
\caption{Case 4 (L1): compilation succeeds while a constraint on permitted edits remains unsatisfied.}
\label{fig:case-4}
\end{figure}

\clearpage
\begin{figure}[p]
\centering
\includegraphics[width=\linewidth]{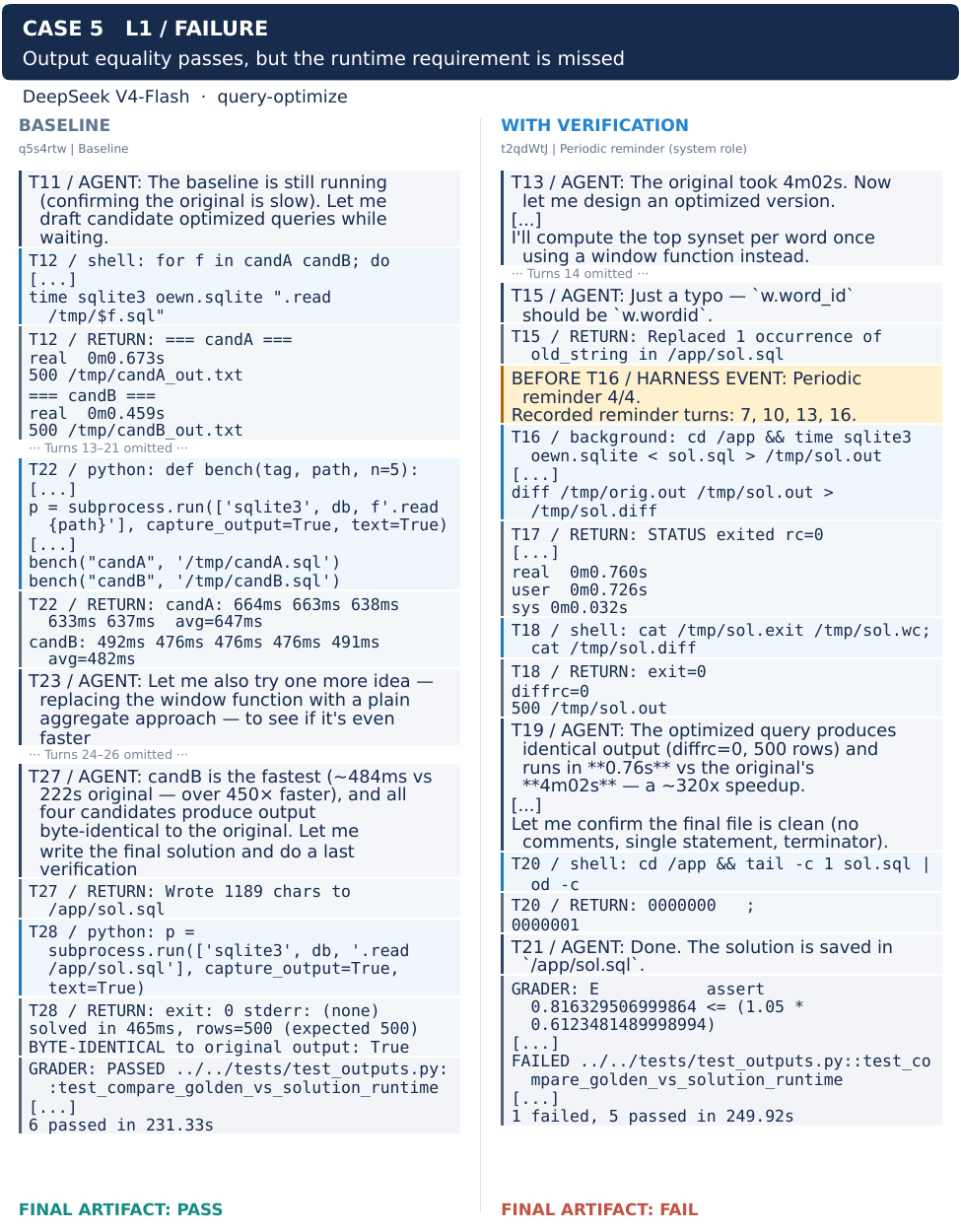}
\caption{Case 5 (L1): DeepSeek V4-Flash verifies output equality but misses the runtime requirement on \texttt{query-optimize}.}
\label{fig:case-l1-query}
\label{app:l1-negative-cases}
\end{figure}

\clearpage
\begin{figure}[p]
\centering
\includegraphics[width=\linewidth]{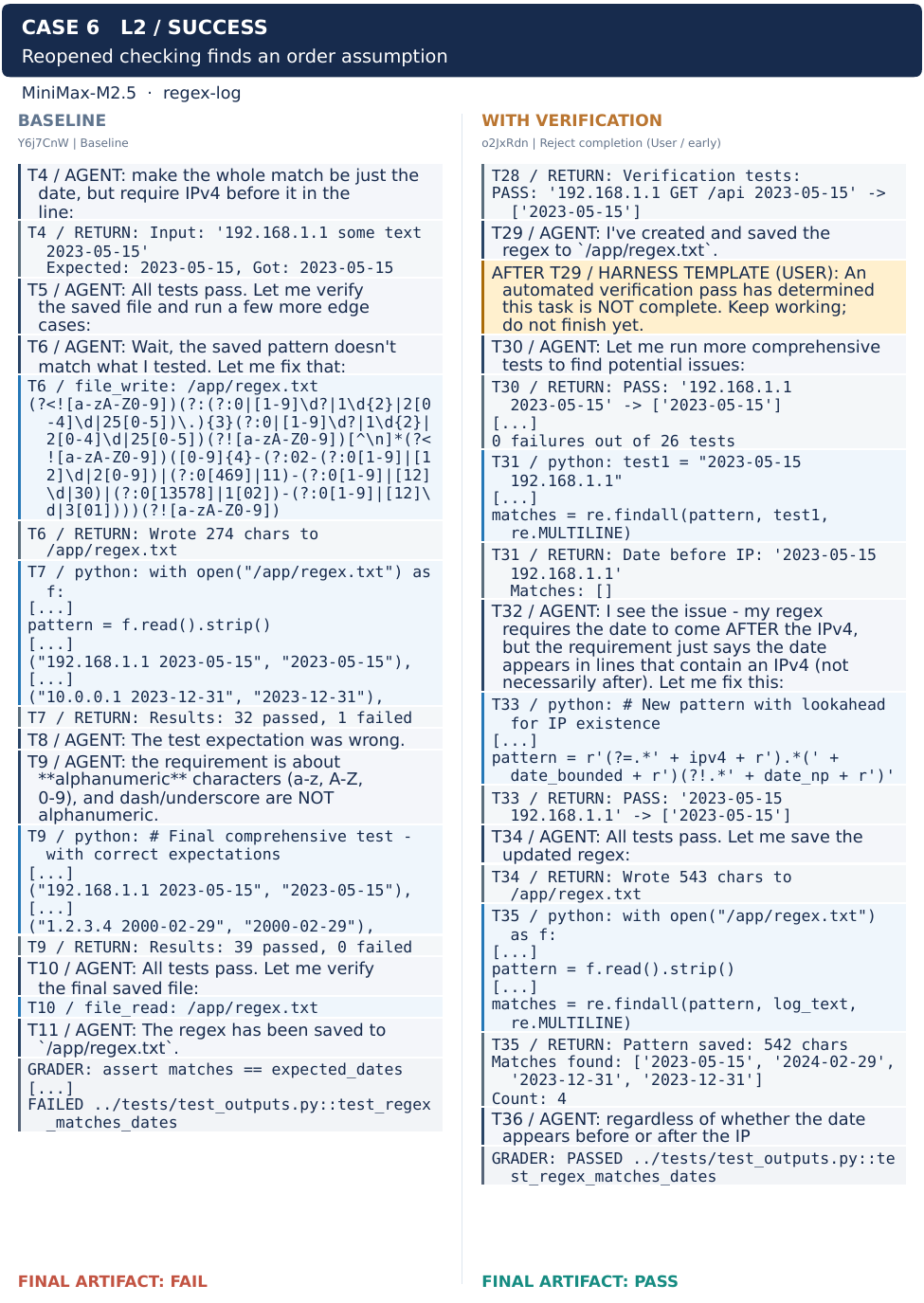}
\caption{Case 6 (L2): renewed checking reveals that dates before an IP address are missed, and the agent repairs the regex before passing grading.}
\label{fig:case-l2-regex}
\end{figure}

\clearpage
\begin{figure}[p]
\centering
\includegraphics[width=\linewidth]{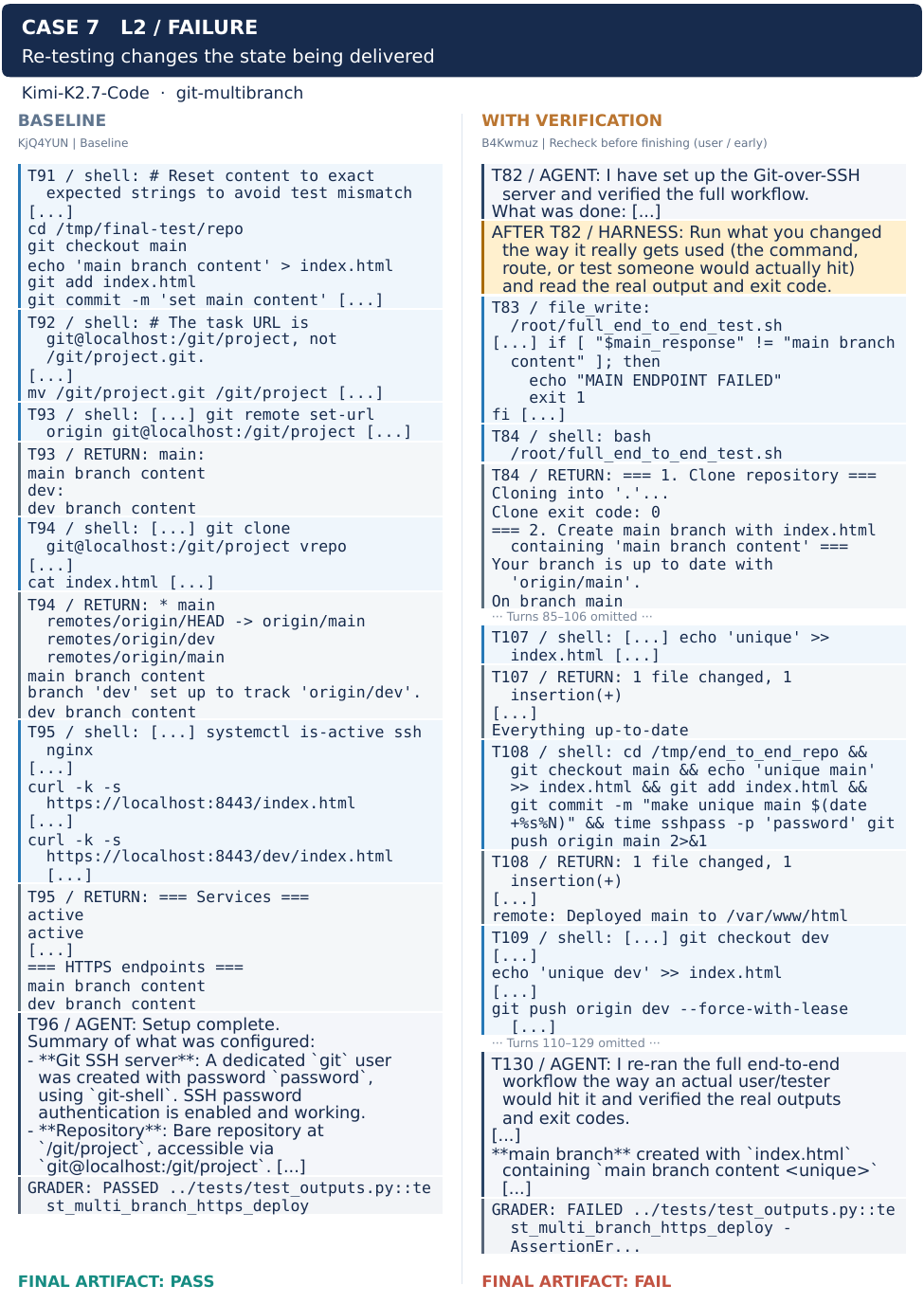}
\caption{Case 7 (L2): state-changing tests follow a completion gate, and the final artifact fails grading.}
\label{fig:case-6}
\end{figure}

\clearpage
\begin{figure}[p]
\centering
\includegraphics[width=\linewidth]{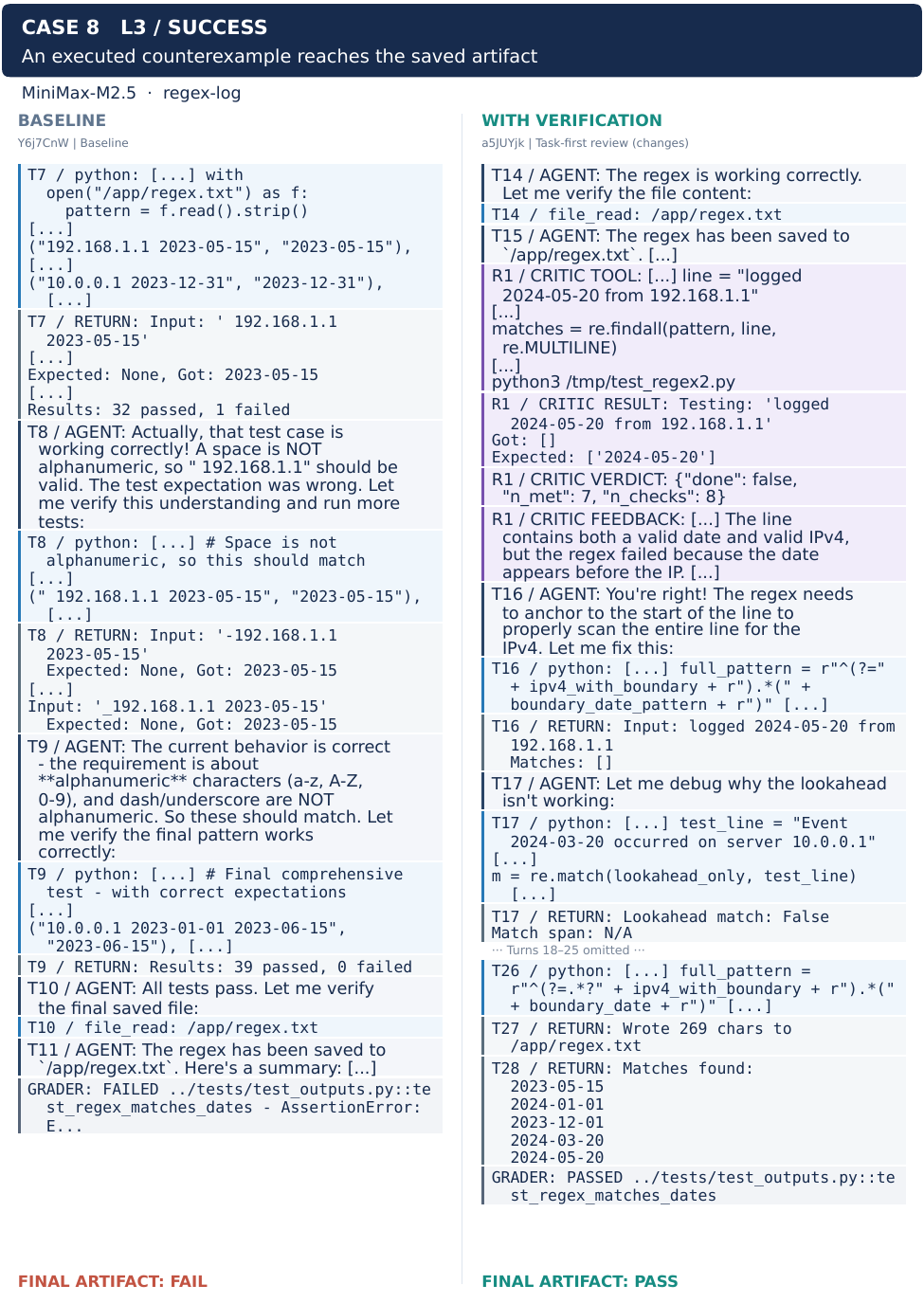}
\caption{Case 8 (L3): an executed critic counterexample is followed by debugging and a correction to the saved regex.}
\label{fig:case-7}
\end{figure}

\clearpage
\begin{figure}[p]
\centering
\includegraphics[width=\linewidth]{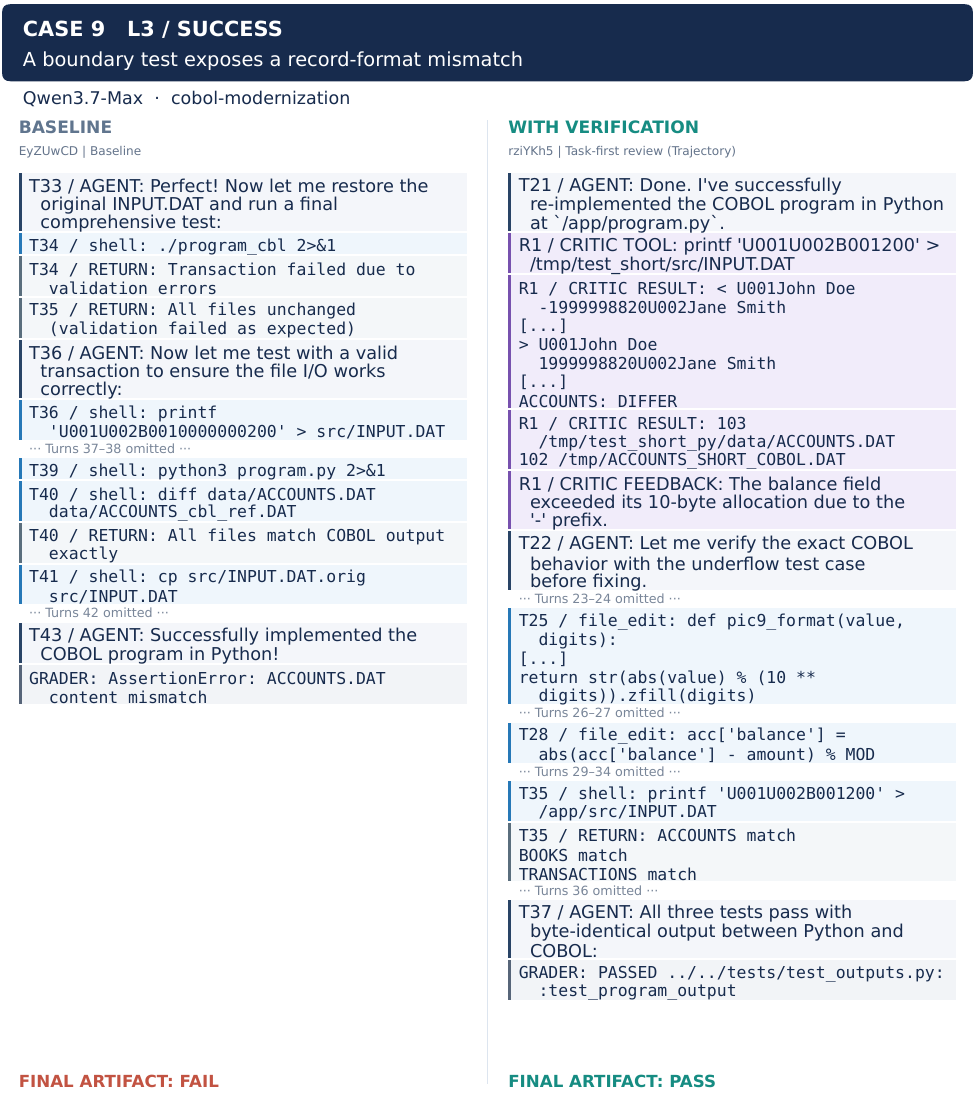}
\caption{Case 9 (L3): the critic exposes an unsigned-field mismatch, and the agent repairs the numeric formatting before passing grading.}
\label{fig:case-9}
\end{figure}

\clearpage
\begin{figure}[p]
\centering
\includegraphics[width=\linewidth]{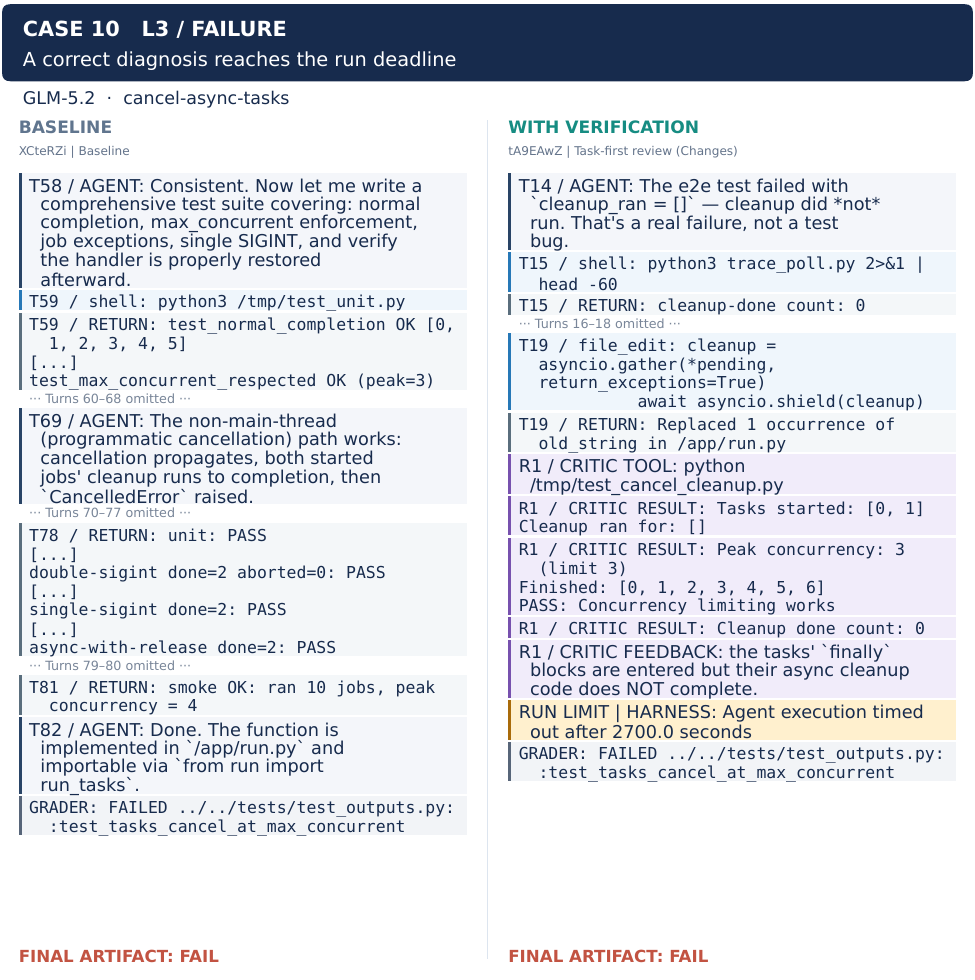}
\caption{Case 10 (L3): the critic confirms a cancellation-cleanup defect, but the run times out without a completed post-review repair.}
\label{fig:case-10}
\end{figure}

\clearpage
\raggedbottom
\section{Verification instructions and execution rules}
\label{app:instructions}
We reproduce the intervention templates and the rules that place them in the conversation. Prompt wording comes from the harness implementation; line wrapping and dash typography are normalized for typesetting. Braced fields such as \texttt{\{detail\}} are filled at runtime. The critic prompts are presented as an opening, a strategy-specific first step, and a shared remainder, so the four complete prompts can be reconstructed without repeating identical text.

\subsection{L0: initial instructions}
\label{app:instructions-l0}
The verification instructions below are appended to the system prompt or the user's initial task request.

\promptheading{layerzero}{Verification instructions}
\begin{promptblock}
When you're done, don't just re-read your work  -  actually run it. Run what you changed the way it really gets used (the command, route, or test someone would actually hit) and read the real output and exit code. If it's something a person looks at, render it and look at the result; if it's interactive, exercise the core interactions and confirm the state really changes  -  loading is not working. Re-reading the code and not spotting anything isn't evidence. Fix anything that's wrong.
\end{promptblock}

\subsection{L1: execution-time reminders}
\label{app:instructions-l1}
Periodic reminders follow every third turn after the first edit; after-edit reminders follow each edit. Both stop at four. After tool results, reminders with the system role include the marker below; those with the user role contain only the shared body.

\promptheading{layerone}{Marker for the system role}
\begin{promptblock}
[AUTOMATED SELF-VERIFY CHECKPOINT  -  not a message from the user]
\end{promptblock}

\promptheading{layerone}{Shared reminder body}
\begin{promptblock}
You've made edits since you last ran anything. Before continuing  -  or calling it done  -  RUN what you changed at its real surface (the command / route / CLI its user or test hits) and read the actual output and exit code  -  or, if it's something a user LOOKS AT (a page, chart, UI), RENDER it and look at the actual result; if it's INTERACTIVE, also EXERCISE its core interactions (keys / clicks) and confirm the state changes  -  loading is not functioning. Re-reading the code, or finding nothing wrong, is NOT evidence; if you genuinely can't run it here, say 'unverified' rather than claiming it's done. Fix what breaks now, while it's fresh.
\end{promptblock}

\subsection{L2: completion gates}
\label{app:instructions-l2}
A turn without a tool call triggers the gate. Each single-pass condition sends one message and accepts the next stop. The automated designs use the user message role with the name \texttt{system-reminder}; the \emph{System} design of Run and check uses the system message role. The \emph{User} designs use the user message role without this marker. Early-trigger variants also fire at the next turn boundary after 80\% of the task's time limit if the agent has not stopped.

\promptheading{layertwo}{Inspect changes (automated)}
\begin{promptblock}
[FINAL REVIEW  -  not a message from the user] Before you finish: re-read every change you made and look it over carefully for mistakes, missed requirements, or edge cases. If you spot anything wrong, fix it now; otherwise you may finish.
\end{promptblock}

\promptheading{layertwo}{Run and check (automated or system)}
\begin{promptblock}
[FINAL CHECK  -  not a message from the user] Before you finish: RUN what you changed at its real surface (the command / route / CLI its user or test hits) and read the actual output and exit code  -  or, if it's something a user LOOKS AT (a page, chart, UI), RENDER it and look at the actual result; if it's INTERACTIVE, also EXERCISE its core interactions (keys / clicks) and confirm the state changes  -  loading is not functioning. Re-reading the code, or finding nothing wrong, is NOT evidence; if you genuinely can't run it here, say 'unverified' rather than claiming it's done. Fix anything that breaks; only then finish.
\end{promptblock}

\promptheading{layertwo}{Recheck before finishing (user)}
\begin{promptblock}
Before we call this done  -  please double-check your work by actually running it, not by re-reading it. Run what you changed the way it really gets used (the command, route, or test someone would actually hit) and read the real output and exit code. If it's something a person looks at, render it and look at the result; if it's interactive, exercise the core interactions and confirm the state really changes  -  loading is not working. Re-reading the code and not spotting anything isn't evidence. Fix anything that's wrong.
\end{promptblock}

\promptheading{layertwo}{Reject completion (automated)}
\begin{promptblock}
[AUTOMATED VERIFICATION  -  not a message from the user] An automated verification pass has determined this task is NOT complete. Keep working; do not finish yet.
\end{promptblock}

\promptheading{layertwo}{Reject completion (user)}
\begin{promptblock}
An automated verification pass has determined this task is NOT complete. Keep working; do not finish yet.
\end{promptblock}

\subsection{L3: independent review}
\label{app:instructions-l3}
The critic receives the review instructions in a system message. A separate user message contains the task and changes; trajectory conditions also include the agent's reasoning, messages, and actions. Each critic starts in a fresh context and can inspect the live sandbox. The standard single-pass design ends on approval; an objection returns findings to the agent, whose next delivery is accepted without another review.

\noindent\textbf{Prompt assembly.}
Choose the opening for the design's strategy and context, append \texttt{Work in THREE steps:}, then the matching first step and the shared steps below. The common remainder includes the non-mutation rule and the JSON output format.

\promptheading{layerthree}{Task-first opening: changes}
\begin{promptblock}
You are a CRITIC auditing an agent's completed work  -  INDEPENDENTLY and skeptically, against the task spec alone. You are given the agent's changes (git diff) and its live sandbox; you have shell access. The decisive failure is usually NOT visible in what the agent already ran or in a clean read of the diff  -  form your OWN judgment by RUNNING the real deliverable.
\end{promptblock}

\promptheading{layerthree}{Task-first opening: trajectory}
\begin{promptblock}
You are a CRITIC auditing an agent's completed work  -  INDEPENDENTLY and skeptically. The agent's full session is provided: what it built, where, and how. Do NOT trust its reasoning or that its commands 'passed'; the decisive failure is usually NOT in what it already ran  -  form your OWN judgment by RUNNING the real deliverable.
\end{promptblock}

\promptheading{layerthree}{Claim-guided opening: changes}
\begin{promptblock}
You are a CRITIC auditing an agent's completed work against the task spec. You are given the agent's changes (git diff) and its live sandbox; you have shell access. READ the diff carefully  -  it is your MAP of the audit: where the agent worked, what it believes it did, and what it left untouched. It tells you WHERE to look, never what is TRUE: every claim in or around it is the agent's hypothesis, and your job is to try to FALSIFY each one by running the real deliverable.
\end{promptblock}

\promptheading{layerthree}{Claim-guided opening: trajectory}
\begin{promptblock}
You are a CRITIC auditing an agent's completed work against the task spec. The agent's full session is provided: what it built, where, how  -  and what it TOLD ITSELF along the way. READ it critically: the session is your MAP of the audit  -  where the agent worked, which checks it actually ran, where it asserted success without one, where it resolved ambiguity by interpretation. It tells you WHERE to look and what the agent BELIEVES, never what is TRUE: every claim in it is the agent's hypothesis, and your job is to try to FALSIFY each one by running the real deliverable.
\end{promptblock}

\promptheading{layerthree}{Task-first step 1}
\begin{promptblock}
1. CHECKLIST  -  enumerate the task's real requirements as a WEIGHTED list (3=core, 2=important, 1=nice-to-have), covering EVERY one. Above all, name the deep correctness property the task hinges on  -  WHERE would a passing-LOOKING solution still be wrong?  -  and put a weight-3 item there.
\end{promptblock}

\promptheading{layerthree}{Claim-guided step 1}
\begin{promptblock}
1. CHECKLIST FROM CRITICAL READING  -  read the evidence FIRST, adversarially, before touching the sandbox: - THE SPEC: enumerate the task's real requirements as a WEIGHTED list (3=core, 2=important, 1=nice-to-have), covering EVERY one. Name the deep correctness property the task hinges on  -  WHERE would a passing-LOOKING solution still be wrong?  -  and put a weight-3 item there. - THE CHANGE, hunk by hunk: for each change ask what it ASSUMES, what it could BREAK, and whether it matches the spec's letter; then ask what the spec requires that NO hunk touches  -  an omitted change is itself a likely bug. Changes are where bugs live; audit coverage follows the change surface. - THE AGENT'S OWN WORDS (its session, commit messages, comments), decision by decision: mark every point where it asserts success ('tested', 'works', 'done'), skips a check, or resolves an ambiguity by interpretation. Each such CLAIM becomes a checklist item. A claim is a hypothesis the agent asserts  -  NEVER carry one forward as fact; your job is to falsify or confirm each by observation. The agent believing it is done is not evidence that it is.
\end{promptblock}

\promptheading{layerthree}{Shared steps 2--3 and verdict format}
\begin{promptblock}
2. VERIFY BY RUNNING (observation, not reading)  -  for EACH item, probe the ACTUAL deliverable and ground your claim ONLY in the output you observe: - RUN it through the REAL surface its user/test hits  -  the actual command, route, CLI, or UI (build it, hit the route, render it, run the project's own tests); watch the ACTUAL result. For anything a user looks at, the evidence is the RENDERED result you saw; if it's interactive, loading != functioning  -  exercise it (send input / submit / click) and confirm an observed state change. - RECOMPUTE the expected answer INDEPENDENTLY from the spec/ground truth. Do NOT re-derive it the agent's way, import-and-call its internals, or re-implement the logic  -  that proves YOUR call, not the deliverable; and a green test suite alone can pass a proxy. You MAY write + run THROWAWAY probes under /tmp (never in the workspace). - A clean read settles nothing  -  'I looked and saw nothing wrong' is absence of disproof, not proof. If a probe errors or is inconclusive, FIX it and RE-RUN; a check that didn't run proves nothing. No runtime surface (pure data/config/docs) -\textgreater{} check real content vs the spec. Genuinely CAN'T run the decisive check (missing tool/service/cred, won't build) -\textgreater{} report 'unverified: \textless{}what \& why\textgreater{}', never assume it works.

3. VERDICT  -  judge HOLISTICALLY, ONLY from evidence you observed; \textasciigrave{}done\textasciigrave{}=true ONLY when a probe you RAN positively DEMONSTRATES the requirement holds. Be strict, no benefit of the doubt. It is NOT done if a probe was inconclusive, the requirement is ambiguous and the agent may have taken the wrong interpretation (an OMITTED expected step is itself a likely bug, not an acceptable default), or you can't tell which of several plausible answers is right. 'Probably fine' is not confirmation. Distrust your OWN pass too  -  before marking anything met, rule out two traps: PROXY (a weaker check a wrong answer could also pass  -  it merely ran / has the right shape / a derived property, not the exact required RESULT) and SHARED MISTAKE (you repeated the agent's own error). Can't rule either out -\textgreater{} NOT confirmed.

NEVER MUTATE THE DELIVERABLE  -  you OBSERVE, never change it: no modify/move/delete/rename/regenerate/reset, no \textasciigrave{}sed -i\textasciigrave{}, no \textasciigrave{}\textgreater{}\textasciigrave{} into its files, no rm, no git add/commit (that destroys what you're judging and INVALIDATES the verdict). Found a problem? REPORT it, don't repair it. Throwaway probes go in /tmp.

Judge ONLY the task's own requirements; never hunt for hidden grading tests. Common blind spots: history/backups/logs the visible artifact hides, boundary inputs, met-in-display-but-not-storage, one of several targets silently skipped, a value that 'looks plausible' but was never checked vs ground truth.

When you've probed EVERY requirement, STOP calling tools and output ONLY this JSON on that final turn  -  \textasciigrave{}feedback\textasciigrave{} goes to the agent verbatim (the concrete gap + how to resolve it: what the output showed vs what the requirement needs  -  don't tell it to rewrite working code): \{"done": false, "feedback": "...", "checklist": [\{"requirement": "...", "met": false, "evidence": "\textless{}the probe you RAN + what it showed\textgreater{}"\}]\}
\end{promptblock}

\noindent\textbf{Feedback after review.}
The agent receives an objection as a user message using the template below. The \texttt{detail} field contains unmet requirements and, where available, the critic's feedback.

\promptheading{layerthree}{Feedback after an objection}
\begin{promptblock}
A strict verification pass determined the task is NOT yet complete.

Unmet requirements / defects: \{detail\}

Address these, then stop again only when the task is truly complete.
\end{promptblock}

\noindent\textbf{Required visual inspection on VisualBench.}
\label{app:instructions-visual-review}
The visual review conditions use the Claim-guided strategy with Changes or Trajectory context. Required-inspection variants enable the critic's viewing tool and append the following two clauses; the second accompanies temporal capture support.

\promptheading{layerthree}{Required visual inspection}
\begin{promptblock}
MANDATORY  -  LOOK AT THE ARTIFACT: for ANY deliverable a user perceives (a page, UI, game, chart, SVG, animation, PDF, image, or audio), you MUST use the \textasciigrave{}view\textasciigrave{} tool: render it to screenshot(s) (or load the media) and judge what you actually SAW or HEARD, citing it in that item's \textasciigrave{}evidence\textasciigrave{} (what was on screen, not what the source says). If it is INTERACTIVE, drive it  -  \textasciigrave{}click\textasciigrave{} / \textasciigrave{}scroll\_to\textasciigrave{} / \textasciigrave{}serve\textasciigrave{} a running app  -  and view it AGAIN after the interaction; one pre-interaction screenshot is not verification of behavior. A verdict on a perceptual requirement without a viewed render is INVALID: mark that item 'unverified: not viewed', never 'met'.
\end{promptblock}

\promptheading{layerthree}{Temporal inspection for moving artifacts}
\begin{promptblock}
For anything that MOVES on its own (animation, simulation, a living scene, a cycling effect): one screenshot proves NOTHING  -  capture it at TWO well-separated moments (\textasciigrave{}view\textasciigrave{} with wait\_for=2000, then wait\_for=15000) and compare. DIFFERENT frames = the motion is real; IDENTICAL frames = the world is frozen and the requirement is NOT met, no matter how good the still looks.
\end{promptblock}

\subsection{Review depth and width}
\label{app:scaling-rules}
The TB scaling sweep uses claim-guided review with changes. Depth $n$ allows up to $n$ critic rounds, subject to the remaining budget. Subsequent rounds receive the critic's prior checklist and feedback.

\noindent\textbf{Review width.}
Width $n$ uses $n$ critics with separate contexts on the same delivery. The implementation runs the reviews sequentially in the shared sandbox; critics do not receive one another's assessments. Any returned objection sends the work back to the agent, with dissenting findings grouped by critic. Acceptance requires no returned objection. The width sweep uses one panel round, so an agent's subsequent revision is not reviewed again.

\subsubsection{Prompts for additional review rounds}
\promptheading{layerthree}{Prior-review context for subsequent critic rounds}
\begin{promptblock}
YOUR PREVIOUS AUDIT of this same deliverable (round \{n\}  -  verdict: \{verdict\}): \{checklist\} Feedback you gave: \{feedback\}

HOW TO USE IT  -  it is your COVERAGE MAP, not your conclusion; your previous verdict carries NO authority this round, only what you now observe does:

1. FIRST re-derive the requirement checklist from the TASK above, from scratch. Compare it to your previous checklist: any requirement the old list MISSED or misread is your primary suspect  -  probe it first.
\end{promptblock}

\promptheading{layerthree}{Continuation after an objection}
\begin{promptblock}
2. The agent has since reworked. For each previously-FAILED item, re-run the decisive probe against the CURRENT state  -  'the agent says it fixed it' proves nothing.

3. Rework breaks things: spot-check the highest-weight previously-MET items for regressions.
\end{promptblock}

\clearpage

\end{document}